\documentstyle[ustockholm_thesis,11pt]{book}

\newcommand{\var}[1]{$\it \langle #1 \rangle$}

{\end{list}\end{small}}

{\end{list}}

{\begin{list}{$\bullet$}{\addtolength{\parsep}{-.25\parsep}
\setlength{\itemsep}{0in}
\setlength{\topsep}{0in}}}%
{\medskip \end{list}}

\begin{document}

\pagestyle{fancy}

\bibliographystyle{plain}

\pagenumbering{roman}


\begin{titlepage}

\title{\vspace{-3cm} \hspace{-2cm} Abductive Equivalential Translation \\
\hspace{-2cm} and its application to  \\
\hspace{-2cm} Natural Language Database Interfacing\\
\vspace{2cm}
{\footnotesize \hspace{-2cm} Dissertation submitted in partial fulfilment of
the
requirements for the degree of} \\ \vspace{1.5cm}
{\large \hspace{-2cm} {\bf Doctor of Philosophy \\ \hspace{-2cm}
Royal Institute of Technology, Stockholm}}
}

\author{\hspace{-2cm} Manny Rayner \\ \hspace{-2cm} {\normalsize SRI
International, Cambridge}}

\date{\hspace{-2cm} September 1993}

\maketitle

\end{titlepage}


\chapter*{Abstract}

The thesis describes a logical formalization of natural-language
database interfacing. We assume the existence of a ``natural language
engine'' capable of mediating between surface linguistic string and
their representations as ``literal'' logical forms: the focus of
interest will be the question of relating ``literal'' logical forms to
representations in terms of primitives meaningful to the underlying
database engine. We begin by describing the nature of the problem, and
show how a variety of interface functionalities can be considered as
instances of a type of formal inference task which we call ``Abductive
Equivalential Translation'' (AET); functionalities which can be
reduced to this form include answering questions, responding to
commands, reasoning about the completeness of answers, answering
meta-questions of type ``Do you know...'', and generating assertions
and questions. In each case, a ``linguistic domain theory'' (LDT)
$\Gamma$ and an input formula $F$ are given, and the goal is to
construct a formula with certain properties which is equivalent to
$F$, given $\Gamma$ and a set of permitted assumptions. If the LDT is
of a certain specified type, whose formulas are either conditional
equivalences or Horn-clauses, we show that the AET problem can be
reduced to a goal-directed inference method. We present an abstract
description of this method, and sketch its realization in Prolog. The
relationship between AET and several problems previously discussed in
the literature is discussed. In particular, we show how AET can
provide a simple and elegant solution to the so-called ``Doctor on
Board'' problem, and in effect allows a ``relativization'' of the
Closed World Assumption.  The ideas in the thesis have all been
implemented concretely within the SRI CLARE project, using a real
projects and payments database. The LDT for the example database is
described in detail, and examples of the types of functionality that
can be achieved within the example domain are presented.

{\bf KEYWORDS}: Natural language processing, natural language
interfaces, databases, logic programming, equivalences, Closed World
Assumption, ``Doctor on Board'' problem

\chapter*{Acknowledgements}

Nearly every PhD thesis contains a lengthy acknowledgements section,
and this one is no exception. I would like to start by admitting my
debt to the other members of the CLARE project: Hiyan Alshawi, Dave
Carter, Dick Crouch, Steve Pulman and Arnold Smith. Without the unique
contributions they each made to the project, the work I report here
could quite simply not have been carried out. I would particularly
like to thank Hiyan Alshawi. Many of the key ideas grew out of
discussions with Hiyan, and I am sure that some of them are really
more his than mine. I would also like to thank the CLARE sponsors: BP
Research, British Aerospace, British Telecom, SRI International, the UK
Defence Research Agency, and the UK Department of Trade and Industry.

I would never have gotten around to writing the thesis if I hadn't
been patiently persuaded to do so by Carl-Gustaf Jansson, my
supervisor, Elisabeth Kron, my wife, and Lotta Almgren and Alison
Golding, our sometime {\it au pairs}. I hope they consider their
efforts rewarded. For similar reasons, I must also thank my own thesis
students, Bj\"orn Gamb\"ack, Barney Pell and Christer Samuelsson, for
mercilessly teasing me during periods when I thought I wanted to be
doing quite different things. Jerry Hobbs, \"Osten Dahl, Seif Haridi
and Sture H\"agglund agreed to act as my opponent and examiners
respectively despite the heavy demands made on their time. Many people
encouraged me with kind remarks about previous drafts of this work,
and among them I would particularly like to mention Barbara Grosz and
Karen Sparck-Jones.

Finally I would like to thank you, the reader. Please read on: there
are a few good bits, I think, in the middle of all the boring stuff.
And I need your discerning mind to tell me which are which.

\begin{flushright}
Manny Rayner\\
Cambridge, September 1993
\end{flushright}

\tableofcontents

\cleardoublepage

\pagenumbering{arabic}
\chapter{Introduction}

\section{What this thesis is about}\label{Overview}

The basic aim of this thesis is to describe a logical formalization of
the topic usually described as ``Natural Language Database
Interfacing'', presented in a way which is consistent with the methods
of present-day computational linguistics and artificial intelligence.
We will interpret the phrase ``Natural Language Database Interfacing''
in a broad sense, to encompass any kind of use of language in
connection with a database; this is of course somewhat too broad as it
stands, but we want to stress that we are not limiting ourselves to
the procedural goal of constructing a system that can use the database
to answer questions posed in natural language. Our starting point is
rather that we are building a theory which describes the relationship
between a database and the natural-language utterances which in some
way or another make reference it. These utterances can certainly be
questions explicitly about the content of the database (``Which
payments were made on May 23rd?''), but they can also be commands to
supply information (``Tell me about payments made on May 23rd''),
statements supplying new information for the database (``A payment to
British Telecom was made on May 23rd''), or meta-questions (``Do you
know whether any payment was made on May 23rd?''). We will not assume
either that use of language is exclusively by human participants in
the dialogue: we want the theory to support generation of language,
for example statements describing facts already in the database, or
questions asking for fillers of incomplete database records. Since
this thesis is within the field of computer science (rather than, for
example, those of linguistics or philosophy), we also make another
demand, namely that the theory should be computationally tractable:
that is to say, it should in principle (and, preferably, in practice),
be capable of efficient realization as a functioning program.  In
accordance with a now well-established paradigm, such a program will
operate in a way that can be conceptualized as performing inference on
the theory to obtain conclusions of a particular type.

The goals set out in the paragraph above are so vague that they could
potentially encompass an almost limitless variety of approaches,
(although it is also worth noting that they in fact exclude most
previously published work on the subject: see
section~\ref{Related-work}). We will now narrow them to focus more
closely on the actual research that has been accomplished here.
First, we will drastically simplify the problem by not dealing
directly with natural language, but only with logical formulas
representing natural language utterances. It is of course by no means
clear that such a move is really possible (or, some would claim, even
desirable); on the other hand, a great deal of work has been done
during the last century on the representation of language in logic,
which has recently found expression in the creation of {\it natural
language engines}, large implemented systems practically capable of
fairly robust automatic translation between utterances and their
logical representations. This thesis makes extensive, often implicit,
use of one such system, the SRI Core Language Engine (CLE), which is
described in more detail in section~\ref{CLE-overview}.  By availing
oneself of this accumulated body of knowledge, it is possible to
abstract away the linguistic phenomena which intuitively belong to the
surface form of language (morphology, and much of the simpler aspects
of syntax), and instead concentrate on what hopefully are the deeper
semantic issues: how words refer to database records, and how the
context in which a word is used can affect the way in which it refers.
In order for there still to be a problem to solve, it is necessary to
limit the scope of operation of the language engine; we demand,
roughly, that the logical representations which it assigns to
utterances should be in some sense ``literal'', that is to say that
there should exist a close correspondence between the component words
of the utterance and the component symbols of its logical
representation. Although this principle is not applied with complete
rigidity in the CLE, it is given sufficiently full expression that we
hope the results will be of fairly general relevance. In the main body
of the thesis, we will later see numerous examples of these
``literal'' logical forms.

Informally, the key idea in our treatment of the subject is that of
``translation'': we wish to translate vague natural language
utterances into precisely defined expressions in a formal language
that can be used to manipulate and reason about the database.  By
replacing natural language utterances with their logical forms, we can
make this goal more precise: given a logical form $F$, whose constants
correspond to word-senses, we might wish to find an equivalent
expression, $F^\prime$, whose constants correspond to database objects
and relations, perhaps together with elementary arithmetic operations
and other uncontroversial relations; this, roughly, will be our
characterization of the {\it analysis} type of task, the class of
tasks where natural language has be turned into database commands.
Conversely, given a logical expression $F$ whose constants refer to
the database, we might wish to find an equivalent expression
$F^\prime$ whose constants are word-senses; this will be the form of
{\it generation} tasks, tasks where the database is given, and natural
language is to be produced.

The ideas in the last paragraph constrain the picture considerably.
If we are to talk about logical forms of natural language utterances
being ``equivalent'' with formulas directly representing database
concepts, then the equivalence must be with respect to a background
theory. Calling the theory $\Gamma$, we will be able to define a
formula $F_{db}$ expressed in the ``database language'' to be
a translation of a logical form $F_{ling}$ iff
\begin{eqnarray}
\Gamma \Rightarrow (F_{ling} \equiv F_{db})\label{Basic-equiv-0}
\end{eqnarray}
Thus one of our major goals will be the construction of suitable
theories $\Gamma$, and we will have a good deal to say about this in a
moment. Before doing so, however, we will make an important
modification to (\ref{Basic-equiv-0}). It turns out in practice that
if $\Gamma$ is a fairly general theory (that is to say, if it is
applicable in a wide range of situations), then it will
often be the case that equivalences like those in
(\ref{Basic-equiv-0}) can only be proved to follow from it with the
addition of extra assumptions $A$; the reason why these assumptions
have to be treated specially is that they are defeasible,
that is to say that additional knowledge may force them to be
retracted. So the modified form of our specification of what
it means for a logical form $F_{ling}$ and a database expression
$F_{db}$ to be equivalent is
\begin{eqnarray}
\Gamma\cup A \Rightarrow (F_{ling} \equiv F_{db})\label{Basic-equiv-1}
\end{eqnarray}
where $\Gamma$ is a background theory and the $A$ belong to a
set of permissible assumptions. In this case, we will say that
$F_{db}$ is an {\it abductive equivalential translation} of
$F_{ling}$, the translation being from the logical vocabulary of
word-senses to the logical vocabulary of database concepts.
Since equivalence is a symmetric relation, it will be equally
correct to say that $F_{ling}$ is an abductive equivalential
translation of $F_{db}$, reversing the roles of the source and
target languages.
The greater part of the sequel will be concerned
with exploring the ramifications of the concept of
abductive equivalential translation (hereafter usually shortened to
AET), and how it can be used to provide
a practically useful framework within which various natural language
interfacing tasks can be formalized and implemented. Illustrative
examples will be taken from the SRI CLARE system, an advanced
natural language and reasoning system built on top of the CLE,
which makes extensive use of AET.

The rest of the thesis is organized as follows. Section~\ref{CLE-overview}
gives an overview of CLARE, focussing on the role that AET plays
in the system as a whole, and Section~\ref{CLARE-demo} shows
an example session with CLARE.
The thesis proper starts with Chapter~\ref{translation}, which
explains in more detail the various interface functionalities that
we will consider, and how they can be realized in terms of AET;
it also discusses {\it en route} the rough form of the ``background
theory'' $\Gamma$ and the permissible assumptions $A$
of (\ref{Basic-equiv-1}). Chapter~\ref{AET} then considers
computational mechanisms that can be used to solve the problem
of finding an abductive equivalential translation of a formula
given a background theory: it is shown that this can be reduced to
a goal-directed inference task if $\Gamma$ is constructed in a
certain way. Theories of this form will be referred to as
{\it equivalential linguistic domain theories}, and are the subject
of Chapter~\ref{dmi}. A goal-directed reasoning engine used to support
the translation functionalities of chapter~\ref{AET} is described
in chapter~\ref{reasoning}.

Up to this point, we will mostly have
considered only the functionalities which involve mapping queries
and assertions into a form that can be used by the database.
Chapter~\ref{genass} shows how we can use the framework in the reverse
direction to formalize generation of statements describing the
contents of database records, and questions asking about the fillers
of unfilled database fields.  Finally, chapter~\ref{conclusions} sums
up and concludes.  There are three appendices. Appendix~\ref{qsem}
discusses and motivates the theory of question semantics used
throughout the thesis; appendix~\ref{trlsem} defines the syntax and
semantics of TRL, the extended first-order logic used throughout the
thesis; and appendix~\ref{SQL-convert} outlines the methods used in
CLARE to translate TRL logical expressions into SQL query language to
connect to actual SQL databases.

\section{The SRI CLARE system}\label{CLE-overview}

In this section we will briefly describe the SRI CLARE system.  CLARE
is a sophisticated natural language and reasoning system built on top
of the SRI Core Language Engine (CLE).  Although it would certainly
have been impossible to carry out the research reported here without the
benefit of an implemented environment like that provided by the CLE,
much of the CLE's internal structure is at best of only tangential
interest in this context; since the CLE is well-described elsewhere
(Alshawi 1992), we will confine ourselves to providing sufficient
detail that the reader will be able to see where the translation and
reasoning components fit into the total scheme.

CLARE is an integrated system, intended to provide the core
language processing capabilities needed for a wide range of
tasks such as knowledge-based access, machine translation,
text scanning and speech synthesis, which in one way or another
involve manipulating human language. The design of CLARE is
characterized by giving high priority to {\it modularity} and
{\it declarative representation of information}: the system
is composed of a dozen or so cleanly separated subsystems,
which pass information, coded in well-defined intermediate
representations, over well-defined interfaces. Different
types of linguistic knowledge are available in various declaratively
specified forms; the processing modules will usually {\it compile}
the declarative knowledge they need into a form suitable for
the uses to which they intend to put it.

At the coarse level of granularity which will suffice for our
purposes, CLARE can be separated into three main pieces. The first of
these is the CLE, which can be viewed as a module that converts
between two levels of representation, {\it surface form} and {\it
quasi logical form} (QLF). Conversion can take place in either
direction; thus the CLE's {\it analysis} module can use the
declaratively specified linguistic information to derive QLF
representations from surface forms, while the {\it generation} module
uses the same information, compiled in a different way, to derive
surface forms from their QLF representations. QLF can be characterized
as a logical representation which attempts to encode only the
linguistic information in the utterance; this implies that QLF
representations of vague expressions like pronouns and ellipsis are
themselves vague (Alshawi and Crouch 1992). We will actually have
little to say about QLF in this thesis, since it is a level of
representation that is difficult to use directly to support
inferential reasoning.

The second major piece of CLARE is the {\it resolution} component,
which converts between QLF and logical form representations; logical
forms are represented in a conservatively extended first-order logic
called Target Reasoning Language (TRL), defined in
appendix~\ref{trlsem}. The resolution component converts QLF into TRL
by ``resolving'' referentially vague sub-expressions; this can involve
operations corresponding to finding referents for pronouns and
definite noun-phrases, expanding elliptic phrases, and determining the
relative scope of quantifiers. Typically, all these operations involve
inference in some way. In the reverse direction, CLARE has a limited
ability to synthesize QLFs corresponding to TRL expressions. We will
say a little more about the resolution component (primarily in
Chapter~\ref{genass}), but nearly the whole of the thesis will focus
on CLARE's third main subsystem, the {\it inference} component.  This
is the part that manipulates TRL expressions, and which mediates
between TRL and the expressions used for direct communication with the
underlying database; as we have said above, the inference component
also provides an important functionality utilised by the resolution
subsystem.  In the annotated example session with CLARE shown in the
next section, the reader will be able to see more clearly how the
three top-level sub-systems in CLARE interact.

\section{An example session with CLARE}\label{CLARE-demo}

In this section, we present an example session with CLARE to
illustrate some of its capabilities; we will focus on the inference
sub-system. Some of the less interesting output will be edited
away. Comments will be in normal font, and input from the
user prefixed by {\tt user>>}.

In the session, CLARE is acting as an interface to a database of
projects and payments. The database records are based on real ones for
SRI Cambridge, but for confidentiality reasons names have been
replaced by fanciful variants\footnote{Credit for the more amusing
entries is due to Arnold Smith.}, and the larger sums of money
randomly reduced in size.

We omit the start-up sequence, and jump to the first sentence,
a simple question. In this first example, we present the entire
CLARE output; in subsequent sentences, we will mostly focus on the
inference stage.
\begin{verbatim}
user>> List all payments made during 1990.

Segmentation. Morphology. Phrases. Syntactic analysis.
Semantic analysis.
6 well-sorted semantic analyses.
Preference ranking.
   Complete sentence with bracketing:

     "{list {all payments {made {during {1990}}}}}."

   Word senses (unordered):

     list: enumerate
     payment: paying event
     make: perform (rather than "present with")
\end{verbatim}
CLARE has analyzed the sentence as far as the QLF level of
representation, and prints it out with bracketing to indicate major
constituents and paraphrases of the content words. It now goes on to
the resolution stage.
\begin{verbatim}
Initial Resolution.

   "list all payments made during 1990 (the year)!"

   Entities:   you ;
               all payments made during 1990 (the year) ;
               1990 (the year) ;

Scoping.
\end{verbatim}
 CLARE has now completed the resolution stage as well (including
quantifier scoping), and produced a logical form. The only non-trivial
action taken has been to resolve ``1990'' to a year (rather than for
example to a project whose project ID is 1990). The resulting LF is
now passed to inference.
\begin{verbatim}
Assumptions used in query translation:

  all_transactions_referred_to_are_from_SRI
\end{verbatim}
The command could be translated into an executable form, if it was
permissible to assume that all the payments are from SRI.  The result,
slightly simplified, is the following formula:
\begin{verbatim}
forall([Id,Amt,Payee,Date]
       impl(sql_select([Id,Amt,Payee,Date]
                       [SELECT DISTINCT t_1.trn_id,
                                        t_1.amount,
                                        t_1.payee,
                                        t_1.cheque_date,
                        FROM TRANS t_1,
                        WHERE t_1.amount > '0',
                        AND '1-JAN-90' <= t_1.cheque_date,
                        AND t_1.cheque_date <= '31-DEC-90']),
             execute_in_future(display([Id,Date,Payee,Amt]))))
\end{verbatim}
This states that the command will be executed if at some point in
the future each tuple describing an appropriate payment is
displayed.

In order to perform the translation, CLARE has made use of a number of
facts about the relationship between language and database, which
are stored as logical axioms in its {\it linguistic domain theory}
for the projects and payments domain.
\begin{enumerate}
\item Axioms determining the relationship between the word {\it
payment} and the records in the {\tt TRANS} relationship; these say,
roughly, that payments made by SRI between August 17th, 1989 and March
31st, 1991 stand in a one-to-one relationship with records in the
{\tt TRANS} relationship for which the {\tt amount} field is positive.
\item Axioms defining the meaning of the verb {\it make} in the case
when its object is a payment: these say in effect that a payment, and
the making of a payment, are one and the same thing.
\item Axioms defining how the word {\it during} is used; roughly,
that an event is ``during'' an interval if the time-point associated
with it follows the time-point associated with the start of the interval,
and precedes the one associated with its end.
\item An axiom relating the time-point associated with a payment to
the {\tt cheque\_date} field in the {\tt TRANS} record corresponding
to the payment.
\item Axioms defining an appropriate interpretation of the verb {\it
show} when applied to a payment: roughly, that ``showing'' a payment
is the same as finding the payment's associated {\tt TRANS} record,
and printing a tuple consisting of its {\tt trn\_id}, {\tt
cheque\_date}, {\tt payee} and {\tt amount} fields.
\end{enumerate}
CLARE executes the command, resulting in a number of tuples
being printed:
\begin{verbatim}
Executing command:

Answer:

        [100321,09-JAN-90,Martian Systems plc,780]
        [100322,09-JAN-90,UKU,150]
        [100323,09-JAN-90,PonyRides International Ltd,24]
        [100324,09-JAN-90,PonyRides International Ltd,36]
        [100325,09-JAN-90,B Baggins and Sons Ltd,79.09]
        ...
        [100755,17-DEC-90,Maxwells,492.57]
        [100756,17-DEC-90,Maxwells,270.96]
        [100758,21-DEC-90,Point Counterpoint,6.8]
        [100849,29-NOV-90,Ziek Fender,4788.73]
        [100850,29-NOV-90,Reg Donaldson,2853.84]
\end{verbatim}
Finally, a ``key'' is displayed: CLARE uses its knowledge about the
relationships between words and database fields to find one or more
common nouns to describe each field in the tuples it has printed.
\begin{verbatim}
Key for answer fields:

	[A,B,C,D]

    A: payment, transaction
    B: date, time
    C: payee
    D: amount, money

Command executed.
\end{verbatim}
\begin{verbatim}
user>> Show payments from 12/90 to 2/91.

...

   "show all payments from 12/1990 (the month) to
    2/1991 (the month)!"
\end{verbatim}
We skip the initial parts of processing, except for the paraphrase
summary produced by the resolution stage: this confirms that CLARE has
correctly interpreted {\tt 12/90} and {\tt 2/91} as months.
\begin{verbatim}
Assumptions used in query translation:

  all_transactions_referred_to_are_from_SRI
\end{verbatim}
Once again, CLARE has had to assume in order to carry out the
translation that the payments referred to are ones made by SRI.  In
addition to the facts from its linguistic domain theory used in
translating the previous sentence, it has had to apply knowledge
concerning the use of the prepositions {\it to} and {\it from}:
specifically, it has used axioms which state that these prepositions
refer to relations of temporal precedence when used together with
arguments that can be interpreted as time points or intervals. Further
axioms state that months are objects of such a type. The final
translated formula is similar to the previous one.
\begin{verbatim}
Executing command:

Answer:

        [100721,04-DEC-90,Mohawk Taxi Services,172.55]
        [100722,04-DEC-90,Maybe's Couriers Inc,27]
        [100723,04-DEC-90,World Travel,253.35]
        [100724,04-DEC-90,World Travel,226]
        [100725,04-DEC-90,World Travel,226]
        ...
        [100846,21-FEB-91,Streetside Newsagent,12.22]
        [100847,21-FEB-91,Mohawk Taxi Services,151.7]
        [100851,01-FEB-91,Reg Donaldson,1759.13]
        [100853,27-FEB-91,Zippy Express,153.74]
        [100854,28-FEB-91,Camreality,45.2]

Key for answer fields:
                ...
Command executed.
\end{verbatim}
The next question involves translating into a form that involves
a counting operation: we suppress most of the details.
\begin{verbatim}
user>>  How many payments were made during 1990?
        ...
       "how many payments were made during 1990 (the year)?"
        ...
Assumptions used in query translation:

  all_transactions_referred_to_are_from_SRI
\end{verbatim}
Using axioms already described, the final result of translation is as
follows:
\begin{verbatim}
forall([Count],
       impl(sql_select([Count],
                       [SELECT COUNT ( DISTINCT t_1.trn_id ),
                        FROM TRANS t_1,
                        WHERE t_1.cheque_date <= '31-DEC-90',
                        AND '1-JAN-90' <= t_1.cheque_date,
                        AND t_1.amount > '0']),
            execute_in_future(display([Count]))))
\end{verbatim}
and the result of executing it is
\begin{verbatim}
Answer:

        426

Key for answer fields:

	A

    A: number
\end{verbatim}
The next example, though superficially very similar, reveals a couple
of novel points.
\begin{verbatim}
user>>  How much did Cow's Milk receive from SRI?
        ...
        "how much money did Cow's Milk Dairies (the payee)
         receive from SRI (the payee)?"
        ...
Assumptions used in query translation:

  money_is_measured_in_pounds_sterling
  transactions_referred_to_made_between_1989/8/17_and_1991/4/1
\end{verbatim}
In this question, the payee is explicitly specified as SRI, but the
range of payment dates is not mentioned. CLARE consequently no longer
needs to print a warning that the payee is assumed to be SRI, but
instead warns the user about the range for which payment records
exist. The final result of translation is
\begin{verbatim}
forall([Total],
       impl(sql_select([Total],
                       [SELECT SUM ( DISTINCT t_1.amount ),
                        FROM TRANS t_1,
                        WHERE t_1.payee = 'Cow''s Milk Dairies',
                        AND t_1.amount > '0',
                        AND t_1.cheque_date < '15-JUL-93']),
            execute_in_future(display([Total]))))
\end{verbatim}
which when executed yields the answer
\begin{verbatim}
Executing command:

Answer:

        454.22

Key for answer fields:

	A

    A: quantity

Command executed.

(Warning: the response is based on possibly incomplete
information).
\end{verbatim}
CLARE has been told that it is allowed to assume that the payments
referred to have all been made during the period for which payment
records exist.  The assumption has however also been flagged as one
which can potentially result in incomplete answers, so the warning on
the last line is issued.

In the next example, this theme is carried a stage further.
\begin{verbatim}
user>>  Which companies received cheques from SRI last
        February?
        ...
        "which companies received some cheques from SRI
         (the payee) during 2/1992 (the month)?"
        ...
The command could not be translated without violating
the following:

  transactions_referred_to_made_between_1989/8/17_and_1991/4/1
\end{verbatim}
CLARE was able to complete translation of the LF into executable form
with the usual assumption that all the transactions referred to were
made in the period for which records exist. This time, however, it is
able to prove that the assumption is contradicted by the context in
which it is made, since February 1992 is after March 1991. It
consequently informs the user of this and does not attempt to execute the
translated query.

The next few examples illustrate CLARE's handling of the
well-known ``Doctor on Board'' problem, which arises when database
fields are filled by Boolean (yes/no) values.
\begin{verbatim}
user>>  Which employees have a car?
        ...
        (resolution does nothing interesting)
        ...
Assumptions used in query translation:

  all_cars_referred_to_are_company_cars
  all_employees_referred_to_are_at_SRI
\end{verbatim}
CLARE needs to access a new set of axioms in order to translate this
query:
\begin{enumerate}
\item Axioms defining the usage of the expression ``SRI employee'':
roughly, that objects which can be referred to as ``SRI employees''
stand in one-to-one correspondence with records in the {\tt EMPLOYEE}
relation.
\item Axioms defining the usage of the words ``have'' and ``company
car'' when applied to SRI employees: roughly, that there is an object
which can be referred to as a ``company car'', and which a given SRI
employee can be referred to as ``having'' if and only if the {\tt
has\_car} field in the corresponding {\tt EMPLOYEE} record is filled
by a {\tt y}.
\item ``Employee'' may be assumed to be equivalent to ``SRI
employee'', and ``car'' to ``company car''.
\end{enumerate}
Using these facts, CLARE is able to translate the query into the
executable form
\begin{verbatim}
forall([Empl],
       impl(sql_select([Empl],
                       [SELECT DISTINCT t_1.name,
                        FROM EMPLOYEE t_1,
                        WHERE t_1.has_car = 'y']),
            execute_in_future(display([Empl]))))
\end{verbatim}
yielding the response
\begin{verbatim}
Answer:

        Peter Piper
        Zorba Greek

Key for answer fields:

	A

    A: employee, person
\end{verbatim}
By making use of the additional fact that the {\tt has\_car}
field must be filled with either a {\tt y} or a {\tt n}, it is also
possible to deal successfully with the following query:
\begin{verbatim}
user>>  Which employees do not have a car?
\end{verbatim}
finally producing the query
\begin{verbatim}
forall([Empl],
       impl(sql_select([Empl],
                       [SELECT DISTINCT t_1.name,
                        FROM EMPLOYEE t_1,
                        WHERE t_1.has_car = 'n']),
            execute_in_future(display([Empl]))))
\end{verbatim}
and the answer
\begin{verbatim}
Answer:

        Darth Vader
        Geoff Steiner
        Gordon Bennett
        James Cagney
        Mac the Knife
        Maid Marion
        Naomi Brett-Adams
        Phoebe Waters
        R B N Fielder
        Sammy Davis Junior
\end{verbatim}
One of the things that make ``Doctor on Board'' sentences difficult is
that the correct response is sometimes to report that the sentence
cannot be translated: the next example illustrates this.
\begin{verbatim}
user>>  Which car does Peter Piper have?
        ...
\end{verbatim}
CLARE succeeds in partially translating the query, producing the
intermediate formula
\begin{verbatim}
forall([Event,Car],
       impl(employee_has_car(Event,employee1#'Peter Piper',Car),
            execute_in_future(display([Car]))))
\end{verbatim}
with the accompanying informative message
\begin{verbatim}
Assumptions used in query translation:

  all_cars_referred_to_are_company_cars
\end{verbatim}
It has, however, no way to translate this further into a formula
that uses an SQL call to access the {\tt EMPLOYEE} relation; the
only rules available for translating the {\tt employee\_\-has\_\-car}
predicate require the {\tt Car} variable to be existentially
quantified. In other words, they can only apply to formulas referring
to the {\it existence} of cars. The response is now to print
an informative failure message:
\begin{verbatim}
Warning:

My definitions of the following predicates may be inadequate
in this context: [employee_has_car/3]

Query could not be translated further:
not attempting an answer.
\end{verbatim}
The same method can be used to deal with many ``figurative''
or ``idiomatic'' constructions. For example, the original logical
form for the query
\begin{verbatim}
user>> Which companies have a stake in WHIZ?
       ...
\end{verbatim}
is a faithful encoding of the apparent surface form of the sentence:
it asks for objects called ``companies'', which are such that they
``have'' an object called a ``stake'', which is ``in'' WHIZ project.
To answer the query, it must be translated into a form which
accesses the {\tt PROJECT\_\-SPONSOR} relation: the facts used
say in effect that each record in the relation corresponds to a
suitable instance of a ``having'' and a ``stake''. The final
executable query and result are
\begin{verbatim}
forall([Sponsor],
       impl(sql_select([Sponsor],
                       [SELECT DISTINCT t_1.sponsor_name,
                        FROM PROJECT_SPONSOR t_1,
                        WHERE t_1.proj_name = 'WHIZ']),
            display_tuple_in_future([Sponsor])))

Answer:

        Triton Interstellar Electronics

Key for answer fields:

	A

    A: company, sponsor
\end{verbatim}
The next two examples illustrate how CLARE can apply knowledge about
the way in which it is embedded in its real-world ``discourse situation''.
The first one shows an elementary capability in this respect: to be
able to translate the word {\it current} into meaningful database
concepts, CLARE must know what the date is.
\begin{verbatim}
user>>  Show all current projects.
        ...
\end{verbatim}
CLARE's definition of {\it current} amounts to saying that an object
associated with an interval is ``current'' if that interval includes
the present moment. For projects, the relevant interval is the one
bounded by their start and end dates, which for SRI projects are
listed in two fields of the {\tt PROJECT} relation. After informing
the user that it is assuming that all projects referred to are SRI
projects, CLARE ends up with the query
\begin{verbatim}
forall([Name,AccNum],
       impl(sql_select([Name,AccNum],
                       [SELECT DISTINCT t_1.name , t_1.proj_no,
                        FROM PROJECT t_1,
                        WHERE t_1.start_date <= '15-JUL-93',
                        AND '15-JUL-93' <= t_1.end_date]),
            execute_in_future(display([Name,AccNum]))))
\end{verbatim}
which can be executed to yield the answer
\begin{verbatim}
Key for answer fields:

        [SLT,3775]
        [WHIZ,1392]

	[A,B]

    A: project
    B: account
\end{verbatim}
The next sentence is a more interesting example of CLARE's
connection to its utterance situation: it illustrates how a verb like
{\it show} can be treated by the framework as an ordinary content
word.
\begin{verbatim}
user>> Have you shown me the SLT project?
       ...
\end{verbatim}
The axioms for {\it show}, used in translating many of the previous
queries, define an equivalence between the activity referred to as
``showing database objects to the user'' on one hand, and successfully
executed calls to the {\tt print\_\-tuple\_\-in\_\-future} on the
other. By providing a log of such calls (implemented as a set of
clauses for the predicate {\tt tuple\_\-printed\_\-in\_\-past}), CLARE
is able to use the same mechanism to answer questions about
``showing'' actions which have already been carried out. In the present
case, CLARE performs the translation to find out what kind of action
would correspond to a ``showing'' of the SLT project. The translated
query, which is essentially a pattern to match against the query log,
is
\begin{verbatim}
tuple_printed_in_past([SLT,3775])
\end{verbatim}
Since the previous sentence has left a record of this type, the
response is
\begin{verbatim}
Answer:

	Yes
\end{verbatim}
The next sentence illustrates an interesting functionality which
follows as a simple consequence of the translation method: the ability
to answer meta-questions about the system's knowledge.
\begin{verbatim}
user>> Do you know what the sex of each employee is?
       ...
\end{verbatim}
Questions of the above type are inherently ambiguous between the reading
which treats them literally as Y-N questions, and the indirect
question reading which makes them politely phrased WH-questions.
CLARE's strategy is to attempt to translate the ``inner'' or embedded
question, here {\it ``What is the sex of each employee?''}. If this
can be done, it informs the user and asks whether the answer to the
embedded question is also required.

In the present case, the translation could be carried out
successfully, giving the following result.
\begin{verbatim}
Assumptions used in query translation:

  all_employees_referred_to_are_at_SRI

kw(forall([Name,Sex,SexCode]
          impl(and(sql_select([SexCode,Name],
                              [SELECT DISTINCT t_1.sex,
                                               t_1.name,
                               FROM EMPLOYEE t_1]),
                   sex_code(Sex,SexCode))
               display_tuple_in_future([Name,SexCode]))))
\end{verbatim}
The outer-level {\tt kw} operator is to be read ``knows what'' or
``knows whether'', and holds if the argument has been translated
into an executable form. Note also that the final result of translation
includes a call to the predicate {\tt sex\_\-code}, which relates the
internal code used to indicate sex in the {\tt EMPLOYEE} record, to
the externally meaningful values {\tt Male} and {\tt Female}.
The result of executing the query is
\begin{verbatim}
Answer:

	Yes

Do you want me to to tell you? user>> y.

Answer:

        Darth Vader                     Male
        Geoff Steiner                   Male
        Gordon Bennett                  Male
        James Cagney                    Male
        Mac the Knife                   Male
        Maid Marion                     Female
        Naomi Brett-Adams               Female
        Peter Piper                     Male
        Phoebe Waters                   Female
        R B N Fielder                   Male
        Sammy Davis Junior              Male
        Zorba Greek                     Male

Key for answer fields:

	A	B

    A: employee, person
    B: sex
\end{verbatim}
Our final examples show CLARE being used to enter information into the
database by constructing new records, and then describing these records
back to the user. Since it is often impossible to describe a record in
a single sentence, CLARE represents the content of the sentences
previously processed in an intermediate form where fields so far
unfilled by specified values hold existentially bound variables.
Database relations are represented as predicates, with one argument
place for each column in the real relation.

We first show how a {\tt PROJECT} record can be constructed from a
series of sentences, each of which desribes the filler of one or more
field.
\begin{verbatim}
user>> FOO is a project.
       ...
Assumptions used in query translation:

  all_projects_referred_to_are_SRI_projects
\end{verbatim}
After the first sentence has been fully translated, the intermediate
form is
\begin{verbatim}
x([X,Y,Z],
  PROJECT('FOO',X,Y,Z))
\end{verbatim}
The field indicating the project name has been filled by the atom
{\tt FOO}; those for the account number and start and end dates have
not been specified, and thus contain only existentially bound
variables.

The second sentence specifies the project's account, and fills another
field:
\begin{verbatim}
user>> FOO's account is 1234.
       ...
      "the account of FOO (the project) is account 1234."
\end{verbatim}
Resolution paraphrasing make it clear that {\it ``1234''} is being
interpreted as {\it ``account 1234''}. After translation (we omit the
details), the intermediate form is
\begin{verbatim}
x([Y,Z],
  PROJECT('FOO',1234,Y,Z))
\end{verbatim}
The third sentence describes the fillers of both the start and end
date fields at once.
\begin{verbatim}
user>> FOO started on 1/1/91 and finished yesterday.
       ...
       "FOO (the project) started on 1/1/1991 (the day)
        and finished during 14/7/1993 (the day)."
\end{verbatim}
The final result is the completed database record
\begin{verbatim}
PROJECT('FOO',1234,1-JAN-91,14-JUL-93))
\end{verbatim}
It is now possible to ask the system to describe the new record it has
built:
\begin{verbatim}
user>>  Talk about FOO.
        ...
        "talk about FOO (the project)!"
\end{verbatim}
CLARE translates the logical form of the request into a call to the
predicate {\tt talk\_\-about\_\-in\_\-future}, analogous to
{\tt print\_\-template\_\-in\_\-future}.
\begin{verbatim}
talk_about_in_future(project1#'FOO')
\end{verbatim}
CLARE describes the record by filling in predicates in a set of QLF
templates which have been acquired by generalization from training
examples. It avoids unnecessary repetition by checking each new
sentence before printing to make sure it does not logically follow
from those previously printed. It terminates when the set of
descriptive sentences printed is large enough that their conjunction
is logically equivalent with the record being described.
\begin{verbatim}
Trying to describe a database record:

        "FOO (the project)'s end date is 14/7/1993 (the day)."
        "FOO (the project)'s number is 1234."
        "FOO (the project)'s start date is 1/1/1991 (the day)."

(The record was fully described)
\end{verbatim}
Note that the linguistic forms used in the description differ in
almost every way from those used to enter the record.

The final example illustrates CLARE's ability to generate questions.
We repeat the previous example, but this time allowing CLARE to
take the initiative. The first sentence is the same.
\begin{verbatim}
clare>> FOO is a project.
        ...
\end{verbatim}
After this, control is passed to CLARE:
\begin{verbatim}
clare>> .ask
\end{verbatim}
CLARE now attempts to use its domain theory to synthesize questions
that ask for the values of the unfilled fields in the record.
CLARE asks the following question, and waits for the user's reply.
\begin{verbatim}
        "what is FOO (the project)'s account?"

answer_please>> user>> 1234.
\end{verbatim}
The answer is processed in the same way as any other utterance:
resolution interprets in in the context of the question just asked
by the system, producing the following paraphrase:
\begin{verbatim}
         ...
         "account 1234 is FOO (the project)'s account."
\end{verbatim}
Translation then uses this to fill in the ``account'' field in
the record, following which a second question is asked.
Once again, processing proceeds as normally to fill in the
``start date'' field.
\begin{verbatim}
         "what is FOO (the project)'s start date?"

answer_please>> user>> 1/1/91.
         ...
         "1/1/1991 (the day) is FOO (the project)'s
          start date."
\end{verbatim}
The final question demonstrates the flexibility of the interaction
strategy: CLARE has no problems in coping with an answer
which is a complete clause, rather than an elliptical fragment.
\begin{verbatim}
         "what is FOO (the project)'s end date?"

answer_please>> user>> FOO ended on 14/7/93.
         ...
         "FOO (the project) ended on 14/7/1993 (the day)."
\end{verbatim}
When the last answer has been processed, the record is complete:
the fields have the same values as before,
\begin{verbatim}
PROJECT(FOO,1234,1-JAN-91,14-JUL-93)
\end{verbatim}

\chapter{Interfacing as Translation}
\label{translation}

\section{Introduction}
\label{AETIntro}

In this chapter, we will describe at a high level of abstraction the
basic functionalities provided by CLARE that allow it to be used as a
natural-language interface to a knowledge base, and in particular to a
relational database. Before going any further, we will recapitulate at
greater length the arguments from Section~\ref{Overview}, and
establish more precisely what we mean by ``natural-language
interfacing''. Normally, the problem is thought of as that of
producing answers to natural-language questions posed to a knowledge
base (cf. Perrault and Grosz, 1988).  Here, we will adopt a more
general view; we will regard it as the problem of formalizing the
relationship between 1) the contents of a database, and 2) the way in
which people use language to talk about it. Asking questions about the
contents of databases is far from being the only way in which language
and database can be related; thus apart from question-answering, we
wish the model to support reasoning that will allow at least the
following.
\begin{itemize}
\item Responding to natural-language commands.
\item Accepting statements describing new records for the database.
\item Reasoning about whether the database contains the information
necessary to answer an NL question.
\item Indicating whether answers to WH-questions are complete
or not, and distinguishing between ``No'' and ``Don't know'' answers to
Y-N questions.
\item Describing existing records in the database using NL.
\item When records are incomplete, formulating NL questions to
elucidate the information needed to complete them.
\end{itemize}
We want the information needed to support these functionalities to be
expressed declaratively, using some sort of predicate logic; the
result will be a {\it linguistic domain theory} for the database, a
theory of how language and database are related. Languages are
normally based on vague, common-sense concepts, databases on precise,
formal ones.  Relating these two disparate systems raises the usual
problem of common-sense reasoning: interpretation of vague NL concepts
in terms of precise database concepts is often only possible if
additional assumptions, not implied by the explicit linguistic
content, are made.  These assumptions are in general defeasible. In
some cases, it may be desirable to relay them back to human users. If
the scheme is to be practically useful, we also want to be able to
specify a methodology for constructing linguistic domain theories. For
the usual software engineering reasons, they should also, as far as
possible, be modular and re-usable. These, in brief, are our top-level
goals. We will now make them more specific, and relate them to the
concrete architecture of the Core Language Engine.

The CLE, being a general natural language system (rather than one
tailored to a specific application), constructs representations of
utterances that essentially mirror their linguistic content.
Normally, every content word in the utterance will correspond to an
occurrence of a word-sense predicate in that utterance's TRL
representation. Thus for example in the CLARE Project Resource
Management domain the sentence
\begin{description}
\item[(S1)] Show all payments made to BT during 1990.
\end{description}
gets a TRL representation approximately of the form
\begin{eqnarray}
&\forall X.((\exists E\exists A.payment^\prime(X)\wedge
make^\prime(E,A,X,bt^\prime)\wedge during^\prime(E,1990)) \rightarrow \nonumber
\\
&\exists E^\prime.show^\prime(clare^\prime,X)\wedge in\_future(E))
\end{eqnarray}
in which the predicates $show^\prime$, $payment^\prime$, $make^\prime$
and $during^\prime$ correspond directly to the words {\it show}, {\it
payment}, {\it make} and {\it during}. In order to carry out the
command, however, CLARE needs to construct an SQL {\tt SELECT}
statement which searches for {\tt TRANS} tuples where the {\tt payee}
field is filled by {\tt bt}, and the {\tt cheque\_\-date} field by a
date constrained to be between 1st January and 31st December, 1990.
The gap between the two representations can lead to several possible
kinds of difficulties.  Firstly, the ``linguistic'' and ``database''
representations are quite simply very different. Obvious problems
arise when we frame the task in terms of converting the original TRL
representation of {\bf (S1)} into a suitable SQL query. In particular,
the rules that link specific language-related representation elements
to specific SQL constructs will in general need to be
context-sensitive: in {\bf (S1)}, for example, mapping the
representation of the word {\it during} into a restriction on the {\tt
cheque\_date} field will involve taking account of the way {\it
during} is used together with the words {\it make} and {\it payment}
at least.

Alongside of this there is another, more subtle problem. It may be
impossible to derive an SQL query that corresponds to the original
natural-language utterance; or if it is possible, the SQL may not
correspond exactly to the English. There are three particularly
important ways in which this can happen, which will motivate much of
the following discussion:
\begin{enumerate}
\item A query can be {\it conceptually} outside the database's domain.
For example, if ``payments'' in {\bf (S1)} is replaced by
``phone-calls'', the interface should be able to indicate to the user
that it is unable to relate the query to the information contained in
the database.

\item A query can be {\it contingently} outside the database's domain.
Thus if ``1990'' is replaced by ``1985'', it may be possible to derive
a query; however, if the database only contains records going back to
1989, the result will be an empty list.  Presenting this to the user
without explanation is seriously misleading.

\item A query may need additional implicit assumptions to be translatable
into database form. Asking {\bf (S1)} in the context of our example
Project Resource Management domain, it is implicitly understood that
all payments referred to have been made by SRI. If the user receives
no feedback describing the assumptions that have been made to perform
the translation, it is again possible for misunderstandings to arise.
\end{enumerate}
Keeping in mind the problems we have just discussed, we will next
consider methodologies previous used to attack these aspects of the
natural language interfacing task.

\section{Related work}\label{Related-work}

This section examines previous work on the problem of converting
between ``literal'' representations of natural-language utterances and
representations in terms of database primitives. We will in particular
examine three approaches, all of which have to some extent influenced
the work reported here. We list them as follows, with a brief
description of each:
\begin{enumerate}
\item Horn-clause methods: the database is encoded as a set of
unit clauses, and the truth-conditions for the word-sense predicates
are defined by Horn-clauses which eventually bottom out in the
database.
\item Non-inferential equivalential translation: this approach, first
implemented in the PHLIQUA system, is like ours in deriving database
queries equivalent to the original logical forms; unlike ours, it
does so without using general inferential methods.
\item ``Interpretation as abduction'': this approach, most widely
known from the SRI TACITUS system, allows general inference and
also assumption of unproven hypotheses under suitable circumstances.
Unlike our approach, it attempts however to prove implications rather
than equivalences.
\end{enumerate}
In the following three sub-sections, we will
now describe each approach more fully. In the final sub-section,
we summarize the way in which they have contributed to the
scheme described in this thesis.

\subsection{Horn-clause approches}

The currently most popular approach is perhaps to attempt to effect
the connection between LF and database query by encoding the database
as a set of unit clauses, and then building an interpreter for the
logical forms which captures the relations between linguistic and
database predicates as ``rules'' or ``meaning postulates'' written in
Horn-clause form (cf. e.g. McCord 1987, Pereira and Shieber 85).  The
strength of the Horn-clause method is that it allows arbitrary
inferences to be used; moreover, since Horn-clause deduction can be
implemented as a goal-directed backward-chaining algorithm, the
inference process can be made reasonably efficient.  The drawback is
that Horn-clauses are ``if'' rules; they give conditions for the LF's
being true, but (as pointed out in Konolige 1981), they lack the
``only if'' half that says when they are false.  This means that they
can normally only be used to justify positive answers to questions;
there is no easy way to distinguish between ``no'' and ``don't know''
in a pure Horn-clause theory.  It is of course possible to invoke the
Closed World Assumption (CWA); in this interpretation, finite failure
is regarded as equivalent to negation.  Unfortunately, experience also
shows that it is extremely difficult to write meaning postulates for
non-trivial domains that are valid under this strict interpretation.

For these reasons, Scha (1982) concludes that Horn-clause methods are
insufficient; this is regretable, in view of their many obvious
advantages. One way of viewing the present piece of work is as an
attempt to redeem the Horn-clause approach from Scha's objections.

\subsection{Non-inferential equivalential translation}

The second line of attack we will consider could be called
``non-inferential equivalential translation'': the basic idea is to
derive a query which is logically equivalent with the orginal logical
form, but to do so without using general inferential methods, on the
grounds that these cannot be implemented in a sufficiently efficient
way. The best-known systems to embody variants of this approach are
PHLIQA (Bronnenberg et al 1980) and the systems developed at BBN
during the mid-80's (Stallard 1986) where the method is referred to as
`recursive terminological simplification'.  In both of these there is
restricted inference amounting to a form of type-checking. The
advantage of equivalential translation methods in general is that they
justify both positive and negative answers. However, the lack of
general inference results in a loss of expressiveness which can most
obviously be illustrated by variants of the so-called ``Doctor on
Board'' problems (Perrault and Grosz, 1986) caused by the presence of
existential quantifiers. We will return to this point in
section~\ref{Doctor-on-board}.

We should mention at this point a little-known paper of Konolige
from 1981, which also formulates the problem in terms of equivalence
between logical form and database query; particularly with regard
to treatment of proper names we have borrowed from Konolige's idea
to some extent. The paper, however, is written entirely at the
level of denotation, and makes no mention of inference methods
suitable for concretely solving the problem as posed.

\subsection{``Interpretation as abduction''}

The third strand of research we will examine attacks a problem ignored
by both of the first two: as already pointed out, it is not in general
possible to justify a sound inferential connection between logical
form and database without making additional unproven assumptions,
which later information may potentially invalidate. The most obvious
way to extend an inferential framework to account for this objection
is to allow leaf-nodes in proofs to include not only axioms from the
background theory but also formulas which it can be regarded as
reasonable to assume; in order to prevent explosion of the
search-space, it is normally necessary to place constraints on the
formulas which can potentially be assumed. The best-known variant of
this idea is that pioneered by Hobbs and his colleagues under the name
of ``Interpretation as Abduction'' (Hobbs et al 1988) and embodied in
the SRI TACTITUS system. (Other similar work is that of Charniak and
Goldman (1988)). The basic mode of operation of TACITUS is to
attempt interpretation of the logical form by attempting to construct
a proof of its correctness using the background theory and the
``cheapest'' possible set of additional assumptions; for the idea to
make sense, a cost function is defined on the space of logical
formulas which makes intuitively plausible formulas cheaper than
intuitively implausible ones. We will make use of this general idea,
though in a somewhat different context, since we are dealing with
equivalence rather than implication.

\subsection{Relationship of AET to earlier work}

Having briefly described what we view as the main precursors of our
work, we now indicate more exactly how we consider that it combines
and improves on them. AET can be viewed as a version of the
equivalential translation framework used in PHLIQUA and Konolige's
paper, extended to be able to employ general inference methods of the
kind used in Horn-clause systems to justify equivalence between LF and
database form. This depends on coding the relationship between LF and
database forms not as Horn-clauses but as ``definitional
equivalences'', conditional if-and-only-if rules of a particular form.
Our approach retains computational tractability by limiting the way in
which the equivalences can take part in deductions, roughly speaking
by only using them to perform directed ``translation'' of predicates.
However we still permit nontrivial goal-directed domain reasoning in
justifying query derivation, allowing, for example, the translation of
an LF conjunct to be influenced by any other LF conjuncts, in contrast
to the basically local translation in PHLIQA.  This approach deals
with the first two points described at the beginning of the chapter
without recourse to the CWA, and simultaneously allows a clean
integration of a version of the ``abductive'' reasoning used in
TACITUS. As we shall see, it also makes it possible to use
substantially the same framework to achieve interfacing
functionalities other than question-answering.  The main technical
problems to be solved are caused by the fact that the left-hand sides
of the equivalences are generally not atomic.

The rest of this chapter is structured as follows.
Section~\ref{Ontology} begins by laying out the basic ontology
underlying linguistic domain theories of the types we will consider.
After this, Section~\ref{LDT-and-effective-trans} describes the
approximate formal appearance of these theories, and presents the key
technical concept of {\it effective translation}.  The central section
is~\ref{NL-functionalities}, where we give an overview of the
interface system, concentrating on the denotational and functional
aspects; taking them one at a time, we show how the interface
functionalities listed at the beginning of Section~\ref{AETIntro} can
be formalized as tasks of the form ``for some formula $P$, find a
formula $P^\prime$ of a specified type, such that $P$ and $P^\prime$
are equivalent given the linguistic domain theory and some permitted
assumptions.''  We will refer to tasks of this kind as ``performing
abductive equivalential translation (AET) on $P$''. Finally
section~\ref{Closed-World} considers the connection between AET and
the Closed World Assumption.

\section{Ontological issues}\label{Ontology}

We start by defining the basic ontology underlying the linguistic
domain theory; the ideas are closely related to those proposed in
(Konolige 1981).  We assume as usual that there is a set of objects,
$\bf O$, and a set of relations obtaining between them, $\bf R$. $\bf
O$ will contain all the things that the predicates in logical forms
range over --- the things, like suppliers, parts, projects, payments,
deliverables, etc. that natural language refers to.  This includes
events. Similarly, $\bf R$ will contain all the relations obtaining
between elements of $\bf O$ that we will wish to refer to.

There will be no particular need in what follows to be much more
specific about exactly what can and cannot belong to $\bf O$ and $\bf
R$, though this is obviously not a trivial matter. We will however be
interested in picking out certain distinguished subsets of these two
sets which have direct relevance to database interfacing. We take the
position that databases are also objects in the world, consisting of
finite collections of rows of marks; this is not a common viewpoint at
the moment, but we think it corresponds well to the naive view of
``what a database is''. We consequently assume the existence of a
subset $\bf O_{db}$ of $\bf O$ consisting of {\it database objects},
which will be numbers, strings, date representations, and other
``marks'' that can be found filling fields in database records.
Similarly, we will assume the existence of a subset $\bf R_{db}$ of
$\bf R$, which will be the {\it database relations}.  Database
relations are defined by the database's internal structure in terms of
tuples: a database relation $D$ holds of a list of arguments $Arg_i$
iff there is a tuple of type $D$ in the database whose fields are
filled by the $Arg_i$.  Since little can be done with database objects
without recourse to elementary arithmetic and an ability to display
results, we assume two more distinguished subsets of $\bf R$. $\bf
Arith$ will be the relevant arithmetic relations (like ``being
numerically less than'') that can obtain between members of $\bf O_{db}$.
$\bf Exec$ will be a primitive set of relations which the interface
can cause to hold by performing actions.  A typical relation in $\bf
Exec$ would be the one that holds between a database-object, and a
time and place at which the interface displayed it.

In what follows,
the distinction between database objects, the non-data\-base objects
they name, and the terms in the linguistic domain theory that refer to
both of them will be central. Thus for example we distinguish
between
\begin{itemize}
\item a transaction (a non-database object, an event in the exterior world)
\item the term referring to the transaction in the linguistic domain theory
\item the database object (a number) that is the transaction's ID
\item the term in the theory referring to the database object.
\end{itemize}
Although these distinctions may not immediately seem necessary, they
are in fact motivated by typical properties of real databases.
Database objects are often codes or non-unique names, that cannot simply
be treated as names in a first-order theory; for example, the same number,
occurring in different fields in a relation, can be related to distinct
real-world entities, or (if used as a code value), to a property of
an entity.

The linguistic domain theory relates database objects with real-world
objects, allowing us to draw conclusions about the properties of the
real-world objects by examining the database records.  Less obviously,
it can also allow us to draw conclusions about properties of
real-world objects from the {\it lack} of records of a particular type
in the database. This amounts to a principled generalization of the
``closed world assumption'', and is described further in
section~\ref{Closed-World}. The linguistic domain theory can also
partially model the interaction between system and user, by defining a
relation between predicates holding in the ``utterance situation''
(the real-world situation in which the user is interacting with the
interface), and the executable relations. For example there are
real-world predicates which correspond to verbs like ``show'' and
``list'', which are commonly used in commands (e.g. ``Show me all
payments over \pounds 500''); these are related by the linguistic
domain theory to the primitive executable relation of displaying
characters on the screen. Treating the utterance situation uniformly
as part of the domain adds both conceptual elegance and a real
increase in the interface's scope; so it is for instance possible to
deal with queries like ``Which of the payments that you showed me in
answer 12 were made this year?'' in a principled way.

\section{LDTs and effective translations}
\label{LDT-and-effective-trans}

At a pre-theoretical level, our characterization of the NL interfacing
functionalities is actually a very simple one: it is only the
technical details that will prove complex. The key notion is that of
``translation''. Given an interfacing task, described in one
vocabulary, we wish to {\it translate} it into an equivalent task,
described in another vocabulary. Thus for example the
``question-answering'' task is of the form ``produce an answer to the
question whose logical form is $P$''; we wish to translate this into
an equivalent task of the form ``execute the database query
$P^\prime$, and print the results in a readable way''.  Here, the
original task is expressed using predicates corresponding to
word-senses; the problem is to translate it into a task expressed
using predicates corresponding to database relations, arithmetical
operations and primitive executable relations. What we want to do now
is to express these ideas in formal terms, so that we can then realize
them as an inference problem. In particular, we want to introduce
linguistic domain theories so that translation can be regarded as
logical equivalence with respect to a theory.  We will start by
sketching the appearance of a linguistic domain theory, and defining
the formal concept of {\it effective translation}.  Throughout most of
the thesis, we will regard it as sufficient to allow the target
representation (the result of performing the translation) to be a
predicate calculus expression that treats database relations as
predicates; in appendix~\ref{SQL-convert}, we briefly review the module
that performs the final conversion into executable SQL queries.

\subsection{Basic form of linguistic domain theories}\label{Basic-LDT}

Since we are primarily interested in establishing equivalences (rather
than implications), it makes sense to write the LDT as far as possible
using formulas which themselves are equivalences. For various reasons,
it turns out that it is convenient to allow these equivalences to be
conditional; to limit computational complexity, we will only allow
their left- and right-hand sides to be existentially quantified
conjunctions of atomic formulas. Thus in summary, the greater part of
an LDT will be composed of formulas of the general appearance
$$\forall\vec{x}.(Left\equiv Right) \leftarrow Conds$$ where $Left$,
$Right$ and $Conds$ are existentially quantified conjunctions.
Linguistic domain theories typically also contain Horn-clause axioms
and declarations of functional relationships between arguments of
predicates (these are described in detail in
section~\ref{Functional-relations}).

It is permitted during inference to assume goals abductively, at a
cost depending both on the goal and the context in which the
assumption is made. We do not attempt to define a formal semantics for
the notion of abductively justified inference: we merely assume that
costs are assigned according to some scheme that generally makes
low-cost proofs intuitively more appealing than high-cost ones. Since
abductive assumptions are in any event made explicit to the user of
the system, we view the costs essentially as heuristics to control the
order in which the space of possible proofs is searched. The range of
goals that may be abductively assumed is controlled by declarations of
the form ``goal $G$ may be abductively assumed with cost $C$ and
justification $J$ in a context where $Conds$ can be proved to hold''.
Here, $G$ is an atomic formula, $C$ a number and $Conds$ an arbitrary
formula; $J$, the ``justification'', is a tag that can be used to
identify the assumption to the user. An assumption can be identified
as inconsistent if a proof of the negation of the assumed goal can be
found in the context in which it was assumed (cf
Section~\ref{Abductive-translation}). To the extent that this is
possible (cf. Section~\ref{Negated-proofs}), CLARE thus has a limited
ability to handle normal defaults.

Experimentation with CLARE seems to indicate that one can profitably
divide abductive assumptions into at least three distinct types; we
illustrate using the example PRM domain, which covers a project and
payment database for SRI Cambridge. In the PRM domain, it is
reasonable to assume (lacking evidence to the contrary) that
``project'' is equivalent with ``project at SRI Cambridge''. The
content of assumptions of this kind is that the speaker means
something more specific than what was actually said. We consequently
refer to them as ``specializations''.
In contrast, it is also reasonable to assume that ``payment'' means
``SRI payment made during the period covered by database records''.
In this case, however, it seems intuitively less clear that the speaker
intended to use the word in the more restricted sense, and it is more
appropriate to assume that the database is limited in a way which the
speaker may not be fully aware of. We will call assumptions of this kind
``limitations''.
Finally, it may sometimes be appropriate to make assumptions that can actually
be incorrect: for example, we assume that completion dates for future
deliverables have been correctly estimated. We call these assumptions
``approximations''.
Distinguishing between different kinds of assumption will become important
later on, when we consider issues regarding the completeness of answers
to questions.

\subsection{Effective translation}\label{Effective-translation}

We now introduce the key formal definition.
We write $\Gamma$ to symbolize a linguistic domain theory, and
let $F_{source}$ be an arbitrary formula; then we define
$F_{target}$ to be an {\it effective translation} of $F_{source}$
iff there exists a set of abductive assumptions $A$ such that
\begin{eqnarray}
\Gamma\cup A \Rightarrow (F_{source} \equiv F_{target})\label{Basic-equiv}
\end{eqnarray}
and
\begin{itemize}
\item Each assumption in $A$ is acceptable in the context in which it is made.
\item There is a finite proof procedure for determining the truth of
$F_{target}$.
\end{itemize}
The question of whether or not a finite proof procedure exists for a formula,
given a theory, is of course undecidable in general; CLARE only attempts to
solve a simple subcase of this problem.

The first criterion that must be met is that all predicates in
$F_{target}$ must be {\it potentially finite}: by this, we mean that
they should have the property that for at least some combinations of
instantiation of their arguments there are only finitely many
instantiations of the remaining arguments that make them true, and
that these new instantiations can be found within bounded time.  Thus
the minimum requirement is that if all the arguments are ground it is
possible to determine the truth of the relation within bounded time.
There are in practice three kinds of potentially finite predicates in
a linguistic domain theory: database predicates, arithmetic relation
predicates, and primitive executable predicates. We examine each of
these in turn:
\begin{description}
\item[Database predicates] These are predicates directly corresponding to the
database
relations $\bf R_{db}$; a database predicate holds of its arguments
iff the database relation has a row with those arguments. The
finiteness of database predicates follows from the finiteness of the
corresponding database relations. Database predicates are consequently
always finite, irrespective of their instantiation.
\item[Arithmetic relation predicates] These correspond the arithmetic
relations in $\bf Arith$, such as addition, subtraction, and
inequality. In general, arithmetic relation predicates are only finite
if all or all but one of their arguments are instantiated. For
example, if $X$ is instantiated to a given number, there is an
infinite set of values for $Y$ such that $X < Y$ holds. If both $X$
and $Y$ are fixed, however, the truth or falsity of $X < Y$ can in
practice be determined in bounded time. Similarly, if $X$ is
specified, then there is an infinite set of values for $Y$ and $Z$
such that $X + Y = Z$. If $Y$ is also specifie, however, there
only one such $Z$, which can again in practice be found in bounded
time.
\item[Primitive executable predicates] In order to be able to reason
about CLARE's ability to perform actions like displaying of objects,
there is a set of predicates which correspond to the primitive
executable relations $\bf Exec$.  We assume that these predicates are
finite for {\it instantiated} actions. Thus for example we will assume
that it is possible within a bounded time to ensure that any specified
database object is displayed.
\end{description}

Since the finiteness of some predicates depends on their
instantiation, the existence of a finite proof procedure generally
depends on being able to find a evaluation order which ensures that
conditionally finite predicates are sufficiently instantiated by the
time they are queried; CLARE can search for suitable evaluation orders
by permuting the order of evaluation of conjunctions. For example, if
$TRANS/3$ is a database predicate then the strategy ``find an $X$ such
that $X > 100$, then find values of $Y$ such that $TRANS(john,X,Y)$''
is not a finite strategy; however, reversing the order to make the
strategy ``find $X$ and $Y$ such that $TRANS(john,X,Y)$, then
determine whether $X > 100$ holds'' is finite. The search is carried
out using a simple abstract interpretation method, which uses
meta-information about finiteness of predicates with respect to
different instantiation patterns to mimic the behaviour of the real
execution module. This part of the system is described further in
section~\ref{Qopt}.

\section{Formalizing interfacing functionalities}\label{NL-functionalities}

\subsection{Yes/No questions}\label{YNQ-functionality}

We are now in a position to describe formally the various interfacing
functionalities. The simplest one to begin with is that of answering a
Y-N question.  The formal statement of the problem is as follows.  We
are given a formula, $F_{ling}$, which is the logical form of a Y-N
question to be answered.  We wish to find a formula $F_{eval}$ and a
set of permissible assumptions $A$, such that $F_{eval}$ is an
effective translation of $F_{ling}$ modulo $A$ in the sense defined in
Section~\ref{Effective-translation}.  There are several possibilities.
\begin{itemize}
\item No such $F_{eval}$ can be found: the answer is {\bf Don't know}.
\item $A$ is empty and $F_{eval}$
can be proved: the answer is {\bf Yes}.
\item $A$ is empty and $F_{eval}$
cannot be proved: the answer is {\bf No}.
\item $A$ is non-empty, but some member $\alpha$ of $A$ is
such that the negation of $\alpha$ can be proved: the answer is
{\bf Cannot answer question, because this violates the assumption
$\alpha$}.
\item $A$ is non-empty and $F_{eval}$
can be proved: the answer is {\bf Yes, conditional on the validity of
the assumptions}.
\item $A$ is non-empty and $F_{eval}$
cannot be proved: the answer is {\bf No, conditional on the validity
of the assumptions}.
\end{itemize}
If assumptions have been made, the user can be told what they were.
An example of each case, taken from the PRM domain, follows.
\begin{itemize}
\item The question is {\it Does Peter have a dog?}. There is no
effective translation of the logical representation of this question,
so the answer is {\bf Don't know}.
\item The question is {\it Is Peter an employee at BT?}. There is an
effective translation of the logical representation of this question,
but it assumes that Peter is an employee at SRI, which contradicts the
requirement that he is also an employee at BT.  So the answer is {\bf
Don't know, because this violates the assumption that all employees
referred to are at SRI}.
\item The question is {\it Does Peter have a car?}. Peter is an employee,
and there is an effective translation to a query that accesses the {\tt
EMPLOYEE}
relation. The translated query can be proved, so the
answer is {\bf Yes}.
\item The question is {\it Does Gordon have a car?}. This is the same as the
previous
case, except that Gordon does not have a car according to the database, and the
effective translation cannot thus be proved. The answer is {\bf No}.
\item The question is {\it Has Peter booked less than 200 hours to CLARE?}
Peter is an employee, and CLARE is a project, so there is an effective
translation to a query that accesses the {\tt TIMESHEET} relation
under the assumption that the time referred to was booked during the
period for which records were kept. The database reveals that only 165
hours were booked during this period, so the answer is {\bf Yes,
conditional on the validity of the assumptions}.
\item The question is {\it Has Peter booked less than 100 hours to CLARE?}
The assumption used when translating is the same as in the last example, and
the answer is {\bf No, conditional on the validity of the assumptions}.
\end{itemize}

\subsection{Commands and WH-questions}\label{Command-functionality}

The functionality of responding to a natural language command
can be captured by a fairly simple extension of the previous definition.
We will assume that the logical form representation of a command is
a formula which is true if the command will be carried out
at some future time. Once again, we let
$F_{ling}$ be the logical form of the command, and the task is to
find an effective translation $F_{eval}$ modulo a set of assumptions $A$.
$F_{eval}$ will usually be a formula that contains primitive evaluable
predicates.

Answering WH-questions can be treated as a special case of responding
to commands, if we assume that the logical form of a WH-question is of
the type $\lambda x.Q(x)$, where $Q(x)$ holds iff $x$ is an object
satisfying the question.  Thus for
example the logical form for {\it Which payments in June were over
\pounds 1000?} will be roughly
\begin{eqnarray}
\lambda x.payment^\prime(x)\wedge in^\prime(x,june^\prime)\wedge
          over^\prime(x,1000)
\end{eqnarray}
The {\it asking} of a WH-question can be regarded as a command to
display all the objects that would be answers to it. (The issues
involved are discussed at greater length in Appendix~\ref{qsem}; cf.
also (Rayner and Janson 1987)). So if $\lambda x.Q_{ling}(x)$ is the
logical form of the question and $displayed\_in\_future(x)$ is a
predicate that holds if $x$ is displayed at some future time, then the
question can be regarded as equivalent with a command whose logical
form is $$\forall x.Q(x)
\rightarrow \exists t.displayed\_in\_future(x)$$ The
different cases that can arise in responding to a command or
WH-question are fairly similar to those for responding to Y-N
questions. We summarize, this time without examples:
\begin{itemize}
\item No such $F_{eval}$ can be found: the answer is {\bf Don't know
how to respond}.
\item $A$ is non-empty, but some member $\alpha$ of $A$ is
such that the negation of $\alpha$ can be proved: the answer is
{\bf Cannot respond, because this violates the assumption
$\alpha$}.
\item $A$ is empty, or only contains ``specialization'' assumptions, and
$F_{eval}$
can be proved: the response is to perform all the necessary actions
(e.g. displaying objects) and
inform the user that the response is complete.
\item $A$ is empty, or only contains ``specialization'' assumptions, and
$F_{eval}$
cannot be proved: the response is to inform the user that it is impossible to
carry out the command. (This cannot occur with questions).
\item $A$ contains ``limitation'' or ``approximation'' assumptions, and
$F_{eval}$
can be proved: the response is to perform all the necessary actions and inform
the user that the completeness and/or accuracy of the answer depends on the
assumptions.
\item $A$ contains ``limitation'' or ``approximation'' assumptions, and
$F_{eval}$
cannot be proved: the response is to inform the user that it is impossible to
carry out the command if the assumptions are correct.
\end{itemize}

A few words are in order to explain why different types of assumption
can give rise to different responses. The intuitive justification is
that ``specialization'' assumptions are ones where the reasonable
default is to trust that the user intended them to be assumed;
``limitation'' and ``approximation'' assumptions are ones where this
default is not clearly appropriate. The lack of a clear dividing line
between the various types of assumptions mirrors the lack of clear
criteria that can be used to decide when a set of answers to a
WH-question can be regarded as potentially ``incomplete''; the
incompleteness is relative to the user's intent, which can only be
guessed at.

A more advanced treatment might examine the previous discourse when
deciding whether or not to warn the user about potential
incompleteness of information. For example, a typical ``limitation''
assumption is that all events of a certain type (e.g.  transactions)
that are referred to are within the period for which records are
available. The first time this point becomes relevant to translating a
query, it is reasonable to inform the user, specifying the exact
period in question and warning that the answer is based on potentially
incomplete information. On subsequent occasions, it may be more
helpful to assume that the user is already aware of the assumption and
has taken it into account when phrasing her question.

\subsection{Statements describing new information}
\label{Declaration-functionality}

At a pre-theoretical level, the intended behaviour when responding to
a declarative statement is to attempt to interpret it as a description
of a new tuple (or set of tuples) that are to be added to the
database. However, one of the immediate problems that arises when
formalizing this simple picture is that more than one sentence is
normally required to describe a tuple fully; the natural unit for
describing a tuple is perhaps the paragraph. This clashes to some
extent with the sentence-oriented processing strategy in the rest of
CLARE. The compromise position taken is to treat each sentence as a
component in a tuple description that is being built up successively,
as part of a small discourse. When responding to a given declarative
sentence $S$, we will thus assume that a logical formala
$\Lambda_{in}$ is available, which summarizes the content of the
sentences previously processed in the current tuple-description
discourse. We will call $\Lambda_{in}$ the {\it logical discourse
context}. If processing of $S$ does not result in a full tuple
description being produced, a new formula $\Lambda_{out}$ may be
produced which summarizes the content of the current discourse after
processing of $S$ (cf also Section~\ref{Simplify-assertions}).
We will also need a criterion for what it means for a formula
to be a tuple description: we will say that $F$ is a tuple description
iff $F$ is a predication whose predicate symbol refers to a
database relation, and all of whose arguments are terms referring to
database objects.

This brief discussion should serve to motivate the following breakdown
into cases. The starting point is a declarative sentence with logical
form $F_{ling}$, and the current contents of the logical discourse
context $\Lambda_{in}$.  We now attempt to find $A$, $\Delta$ and
$\Lambda^\prime$ such that $\Delta\wedge \Lambda^\prime$ is an
equivalential translation of $F_{ling}\wedge\Lambda_{in}$ modulo $A$,
and $\Delta$ is a conjunction of tuple descriptions.  The cases are:
\begin{itemize}
\item $F_{ling}\wedge\Lambda_{in}$ can be proved false. The user
is informed, and no other action is taken.
\item $F_{ling}\wedge\Lambda_{in}$ cannot be proved false, but no such
$\Delta$ and $\Lambda^\prime$ can be found: the user is informed that
it was so far impossible to construct any tuple, and the logical
discourse context is updated to be $F_{ling}\wedge\Lambda_{in}$.
\item A suitable $\Delta$ and $\Lambda^\prime$ can be found; the
tuples described by the $\Delta$ are added to the database, the user
is informed that tuples have been constructed using the assumptions
$A$, and the logical discourse context is updated to be
$\Lambda^\prime$.
\end{itemize}
The first case can normally only arise when function information can
be used in effect to show that two distinct records would have the
same primary key (cf Section~\ref{Functional-relations}
and the beginning of Section~\ref{DBLDT}).

It is arguable that in the second case the user should be informed if
it is impossible to find a translation $F_{db}$ of
$F_{ling}\wedge\Lambda_{in}$ all of whose predicates are potentially
finite, on the grounds that the statements already received during the
present discourse cannot be translated. The objection is that further
statements may resolve the problem and allow translation to database
form to occur. To take an example from the PRM domain, the first
sentence in the discourse could be {\it Clara is a woman}; this cannot
be translated into database predicates, since information about sex
occurs in two distinct relations, {\tt EMPLOYEE} and {\tt PAYEE}.  If,
however, the second sentence is {\it Clara is an employee}, the
conjunction of the two sentences can be translated. There are other
conceivable reasons why the logical discourse context at an
intermediate point in the discourse may be untranslatable, for
instance that a statement is being made only to provide information
that will allow a subsequent statement to be translated.  It is in
general clear that a tight characterization of the functionality of
responding to declarative sentences is not straightforward.  Although
the user's ultimate goal can be assumed to be that of describing a
tuple, there are many indirect ways in which this can be achieved, and
it can be difficult to distinguish cases where communication has
broken down from cases where the user's intended strategy for
describing the tuple is not yet apparent.

\subsection{Meta-questions}\label{Meta-knowledge-functionality}

Both Y-N and WH questions can occur embedded in meta-questions of the
form ``Do you know {\it Q}?'', e.g. ``Do you know how many people work
on CLARE?'', ``Do you know who received the largest payment?'', ``Do
you know whether there were any payments on 1/11/91?''. The strategies
used to respond to Y-N and WH-questions can extended in a
straightforward way to cover these cases; the intuitive idea is that
the system can be said to ``know {\it Q}'' if it would be able to
produce a complete answer to {\it Q} according to the criteria
described in Sections~\ref{YNQ-functionality}
and~\ref{Command-functionality}. There is an important point here on
which we are implicitly relying: we assume that ``knowing what X is''
is in the context of database-question-answering equivalent with
``knowing how to display X''. In general, there are many good reasons
why ``knowing what X is'' should be interpreted in a context where the
information will be put to a specific use; in a more advanced
treatment of meta-questions, it might be necessary to reason about
possible ways in which the information would be used. For example,
``knowing where X lives'' could mean any of ``knowing how to get to
X's house by car'', ``knowing how to get to X's house on foot'',
``knowing how to give instructions for reaching X's house'' or
``knowing how to display X's postal address''. It is a matter of
everyday experience that these conditions need not coincide.

With the caveats just mentioned, the problem of dealing with
meta-questions is simple. If $F_{ling}$ is the logical form of Q, it
follows from the discussion in~\ref{YNQ-functionality}
and~\ref{Command-functionality} that a sufficient criterion for
producing an affirmative answer to ``Do you know {\it Q}?'' is that
there is an effective translation $F_{eval}$ of $F_{ling}$ modulo a
set of assumptions $A$, where $A$ contains only ``specialization''
assumptions. There are also several ways in which inability to meet
this criterion can be meaningful to a user.  If translation can only
be carried out by assuming a ``limitation'' or ``approximation''
assumption, or by making an assumption whose negation is implied by
its context, then the offending assumption normally constitutes a good
explanation of ``why the system doesn't know''.

The cases for different types of response are simple variants of those
for normal questions.  So for example if database records only
extend back to 17/8/88 it may only be possible to translate the
predicate corresponding to the word ``payment'' into an expression
involving the transaction relation if it is assumed that the payment
was made after 17/8/88.  In this case the meta-question ``Do you know
whether there were any payments on 5/5/86?''  will receive the answer
{\bf No, because this violates the assumption that all payments
referred to were made after 17/8/88}.

\subsection{Generating statements and questions}
\label{Generation-functionality}

Finally, we consider the functionalities of generating statements and
questions.  If $S_{db}$ is a database formula that we wish to realize
in language, the problem is to find a set of NL statements $S_i$ ($i =
1...n$) with logical forms $L_i$, and an acceptable set of assumptions
$A$ such that $$\Gamma\cup A \Rightarrow (\bigwedge L_i \equiv
S_{db})$$ This is fairly obvious; more interestingly, a very similar
treatment works for questions. For example, if $Q_{db}$ is a formula
composed of database predicates, and containing a free variable $x$,
then $\lambda x.Q_{db}$ can reasonably be thought of as an abstract
representation of the WH-question ``What $x$ satisfies $Q_{db}$?''.
Assuming as before that the logical form for a WH-question is a
lambda-abstraction, the problem of finding a linguistic realization of
$\lambda x.Q_{db}$ can be formalized as that of finding a single
natural-language question $Q$ with logical form $\lambda x.
Q_{ling}(x)$, and an acceptable set of assumptions $A$ such that
$$\Gamma\cup A \Rightarrow \forall x.(Q_{ling}(x) \equiv Q_{db}(x))$$
The issues involved are discussed at greater length in
chapter~\ref{genass}.

\section{The Closed World Assumption}
\label{Closed-World}

This section will briefly consider the question of how the Closed
World Assumption fits into the picture we have constructed. To make
the discussion concrete, we will initially focus on the case of Y-N
questions.

We start with a logical form, $F_{ling}$, for which we assume that we
can find an effective translation $F_{eval}$, and assume further that
it turns out that there is no proof of $F_{eval}$.  Inability to prove
$F_{eval}$ will imply that it is false, since we have assumed the
existence of a finite proof procedure; this in turn implies (if the
abductive assumptions are granted) that $F_{ling}$ is false, since
$F_{eval}$ and $F_{ling}$ are equivalent.  With regard to the CWA, the
interesting thing to consider is the part played by the database
predicates that occur in $F_{eval}$.  Knowledge of the database
predicates is complete, since they have ``trivial'' semantics, only
referring to the existence of database tuples and not directly to the
rest of the world; they are closed by construction and the Closed
World Assumption may safely be used on them.  However, the predicates
in $F_{ling}$ are in general not closed. We will now examine in more
detail how the connection between the ``closed'' predicates in
$F_{eval}$ and the ``open'' predicates in $F_{ling}$ is defined.

In practice, the key step in proving the equivalence between
$F_{ling}$ and $F_{eval}$ is usually of the following form.  There are
two predicates $P(x,y)$ and $P_{rec}(n_{x},n_{y})$, whose intended
semantics are, respectively, ``$x$ and $y$ stand in relationship $P$''
and ``It is recorded in the database that objects named $n_{x}$ and
$n_{y}$ stand in relationship $P$''. Thus $P_{rec}$ is closed, but $P$
isn't. The two predicates are related by a domain axiom of the form
\begin{eqnarray}
C(y) \rightarrow
(P(x,y) \equiv \exists n_{x},n_{y}.name_1(x,n_{x}) \nonumber \\
\wedge name_2(y,n_{y}) \nonumber \\
\wedge P_{rec}(n_{x},n_{y})) \label{CW-key-equiv}
\end{eqnarray}
Here, each $name_i(\alpha,n_{\alpha})$ ($i = 1,2$) is a predicate that
relates an object $\alpha$ and the identifier $n_{\alpha}$ assigned to
it according to naming convention $i$. (As pointed out above in
section~\ref{Ontology}, it is important to take account of the fact
that the same identifier can name different objects in different
contexts).  $C(y)$ defines a sufficient condition for $P(x,y)$ to have
been recorded; for example, $y$ might be a date and $C(y)$ could say
that $y$ is within the period for which records have been kept.

Now suppose that $P(x,y)$ occurs in $F_{ling}$ conjoined with some
expression $D(y)$, where $D(y)$ can be proved to imply $C(y)$.
Then~(\ref{CW-key-equiv}) can be invoked to
justify replacing $P(x,y)$ in $F_{ling}$ with $\exists
n_{x},n_{y}.name_1(x,n_{x})
\wedge name_2(y,n_{y})\wedge P_{rec}(n_{x},n_{y})$ while maintaining
equivalence. (This will be justified in detail in
Section~\ref{Universal-equivalences}). The net effect is that the CWA
with regard to the predicate $P$ has been made conditional on the its
context of use.  Equivalences of type~(\ref{CW-key-equiv}) are
discussed further in section~\ref{Database-specific-axioms}.

The above analysis of how the CWA relates to Y-N question can be
extended fairly simply to WH-questions as well. We will in fact
distinguish between two types of WH-question: ``what'' or ``which''
questions, which involve finding objects, and ``how many'' or
``how much'' questions, which involve counting or summming.
In the first case, our treatment, as already indicated, involves
recasting the question in a form where it is a command to display
objets from the database to the user. Again, it will normally be
the case that the predictes used in the orginal query are open:
however, the effective translation will only refer to a closed
world in which names are transferred from the database to the
display screen.

With regard to ``how many'' questions, the original question will in
general refer to counting objects in the exterior world which satisfy
some property $P$.  Here, construction of an effective translation
will rely on axioms in the domain theory that place the counted
objects in one-to-one correspondence with database objects $P^\prime$.
One way of looking at this is that these axioms constitute a local
unique name axiom, which encodes the constraint that names are unique
{\it over the set of objects being counted}.  Once the problem is
reduced to counting database objects, it is again possible to solve it
by examining the actual database. The point is that the intended model
for the theory is finite and available for inspection, so it is
feasible to find and count the extension of the property $P^\prime$ --
an option not available for a general theory.

\chapter{AET}
\label{AET}

\section{Introduction}\label{AET-chapter-intro}

We will now consider the actual computational mechanisms used to
effect the task of carrying out abductive equivalential translation.
The guiding example we will have in mind here is that provided by
logic programming: one of the things that has made Prolog a success is
the clear procedural model that can be associated with a Horn-clause
program. This chapter will attempt to do the same thing for a slightly
richer kind of theory; it describes a kind of logic-programming with
equivalences.

More exactly, recall that the main body of the declarative knowledge
used is coded as a set of conditional equivalences, formulas of the
type\footnote{Quantification over the $\vec{x}$ on the left-hand side
will often in practice be vacuous. In this and other formulas, we
assume implicit universal quantification over free variables.}
\begin{equation}
Conds \rightarrow ((\exists \vec{x}.P_1\wedge P_2\wedge P_3 ...) \equiv
P^\prime) \label{MP}
\end{equation}
The attractive aspect of this type of equivalence stems from the fact
that it can be given a sensible interpretation in terms of the
procedural notion of ``translation''.  The intuitive idea is that is
that~(\ref{MP}) can be read as ``$P_1$ can be expanded to $P^\prime$
if it occurs in an environment where $P_2\wedge P_3 ...\wedge Conds$
can be inferred''. The ``environment'' is provided by the conjuncts
occurring together with $P_1$ in the original logical form, other
meaning postulates, and the contents of the database; we will call
it the {\it conjunctive context}. The result is a framework in which
arbitrary domain inference can play a direct role in justifying the
validity of the translation of an LF into a particular database query.

The chapter is structured as follows.
Section~\ref{Formalization-of-AET} defines the basic formalization of
AET in terms of {\it translation schemas}, formulas that allow
translation of a complex formula to be defined in terms of translation
of its components. Initially, higher order-order constructs and
abductive proofs are ignored; section~\ref{Simple-example} gives a
simple illustrative example. After this,
Section~\ref{Higher-order-AET} then extends the framework to cover
translation of higher-order formulas, and
Section~\ref{Abductive-translation} describes how translation can be
made abductive.  Section~\ref{Simplification} describes the
simplification module.  Section~\ref{Doctor-on-board} considers the
way in which AET's handling of existential quantification allows it to
deal successfully with the ``Doctor on Board'' problem, the standard
test-case in the literature.  The discussion up to this point has been
at an abstract level; Section~\ref{AET-Engine} now sketches the
implementation of the AET interpreter in Prolog. Finally,
Section~\ref{AET-debugger} describes a simple AET debugger, based on
the Prolog debugger.

{\bf Remark on notation.} This chapter describes the central ideas of
AET at an abstract and fairly implementation-independent level, and to
emphasize this we will usually present logical formulas in standard
notation. When referring to examples from the implemented system,
however, it will sometimes be convenient to employ the concrete Prolog
syntax used by CLARE, writing {\tt and(P,Q)} instead of $P\wedge Q$,
{\tt exists(X,P)} instead of $\exists X.P$, and so on.  In the next
chapter, where we refer constantly to examples from the system, we
will switch almost exclusively to the concrete syntax.  We hope that
this lack of consistency will not cause the reader undue difficulties.

\section{Formalization of AET}\label{Formalization-of-AET}

This section describes the formal basis for AET; it shows how
conditional equivalences of type~(\ref{MP}) can be used to perform
recursive goal-directed translation of formulas according to a simple
and well-defined scheme. We begin with a manoevre that will
turn out to result in a great simplification of the technical
problems. Instead of allowing arbitrary formulas of type~(\ref{MP}), we
constrain them by demanding that one of the following two
conditions holds:
\begin{enumerate}
\item The left-hand side is a conjunction of atomic formulas.
\item The left-hand side is an existentially quantified atomic
formula, and the conditions are trivial.
\end{enumerate}
The two types of permitted formulas are illustrated
in~(\ref{Equiv-univ-LHS}) and~(\ref{Equiv-ex-LHS}):
\begin{equation}
Conds \rightarrow ((P_1\wedge P_2\wedge P_3 ...) \equiv P^\prime)
\label{Equiv-univ-LHS}
\end{equation}
\begin{equation}
\exists\vec{x}.(P_1) \equiv P^\prime \label{Equiv-ex-LHS}
\end{equation}
We will refer to equivalences of type~(\ref{Equiv-univ-LHS}) as
{\it universal} equivalences, and those of type~(\ref{Equiv-ex-LHS})
as {\it existential} equivalences.
Allowing only these types of equivalence actually entails no loss of
generality, since an expression of the form
\begin{equation}
Conds \rightarrow ((\exists\vec{x}.P_1\wedge P_2\wedge P_3\ldots) \equiv
P^\prime)
\end{equation}
can be rewritten, introducing a new predicate $P_{123}$, as the two
equivalences
\begin{equation}
\forall\vec{x}.(Conds \rightarrow ((P_1\wedge P_2\wedge P_3\ldots) \equiv
P_{123}(\vec{x})))
\end{equation}
of type~(\ref{Equiv-univ-LHS}), and
\begin{equation}
(\exists\vec{x}.P_{123}) \equiv P^\prime
\end{equation}
of type~(\ref{Equiv-ex-LHS}).
The considerations relevant to use of universal and existential
equivalences turn out to be quite different in nature. We
will consequently discuss them seperately, starting with the
universal ones.

\subsection{``Universal'' equivalences and translation schemas}
\label{Translation-schemas}\label{Universal-equivalences}

We first present the basic motivating example, which we will quickly
generalize; we will show how a ``universal'' equivalence -- an
equivalence of type~(\ref{Equiv-univ-LHS}) -- can be used to translate
a conjunction $P\wedge R$ in a way that allows $R$ naturally to be part of
the context in which $P$ is translated. Suppose then that we are given
our equivalence
\begin{equation}
Conds \rightarrow (P_1\wedge P_2\wedge P_3 ...) \equiv P^\prime
\label{Universal-eq-instance}
\end{equation}
and that $P$
is a ground goal that unifies with $P_1$ with most general unifier
$\theta$, i.e.  $\theta(P_1) = P$. Suppose also that $R$ implies
$\theta(P_2\wedge P_3 ...\wedge Conds)$. Then we claim that
\begin{equation}
R \wedge P \equiv R \wedge \theta(P^\prime) \label{Eq-schema-0}
\end{equation}
The proof is simple. In the $\Leftarrow$ direction, we have that $R$
implies $\theta(Conds)$ by hypothesis.
Also,~(\ref{Universal-eq-instance}) means that $\theta(P^\prime \wedge
Conds)$ implies $\theta(P_1)$, which by hypothesis is equal to $P$,
from~\ref{Equiv-univ-LHS}. Hence $R \wedge \theta(P^\prime)$ implies
$P$ and thus $R \wedge P$.

In the $\Rightarrow$ direction we have by hypothesis that $P =
\theta(P_1)$ and $R$ implies $\theta(P_2\wedge P_3 ...\wedge Conds)$.
Hence $R \wedge P$ implies $\theta(P_1 \wedge P_2\wedge P_3 ...\wedge
Conds)$ which by~(\ref{Universal-eq-instance}) implies $\theta(P^\prime)$.
Hence $R \wedge P$ implies $R \wedge \theta(P^\prime)$. We can
summarize~(\ref{Eq-schema-0}), together with its associated
conditions, as the inference rule
\begin{eqnarray}
& & (Conds \rightarrow (P_1\wedge P_2\wedge P_3 ...) \equiv P^\prime) \wedge
\nonumber \\
& & R \rightarrow \theta(P_2\wedge P_3 ... \wedge Conds) \nonumber \\
& \Rightarrow & R \wedge \theta(P_1) \equiv R \wedge \theta(P^\prime)
\label{Eq-schema-1}
\end{eqnarray}

The next step is to generalize~(\ref{Eq-schema-1}) by making explicit
the concept of the ``conjunctive context''. We do this by splitting it
into two inference rules,~(\ref{Eq-schema-base})
and~(\ref{Eq-schema-conj}), as follows:
\begin{eqnarray}
& & (Conds \rightarrow (P_1\wedge P_2\wedge P_3 ...) \equiv P^\prime) \wedge
\nonumber \\
& & Context \rightarrow \theta(P_2\wedge P_3 ... \wedge Conds) \nonumber \\
& \Rightarrow & Context \rightarrow (\theta(P_1) \equiv \theta(P^\prime))
\label{Eq-schema-base} \\
& & \nonumber \\
& & {Context \wedge Q \rightarrow (P \equiv P^\prime)} \nonumber \\
& \Rightarrow & Context \rightarrow (P \wedge Q \equiv P^\prime \wedge Q)
\label{Eq-schema-conj}
\end{eqnarray}
The proofs of (\ref{Eq-schema-base}) and~(\ref{Eq-schema-conj}) follow
from that of~(\ref{Eq-schema-1}). The two rules can be used
recursively together to translate formulas composed using an arbitrary
number of occurrences of the conjunction operator. In both rules, the
formulas before the $\Rightarrow$ are the premises, and the formula
after the conclusion.  (\ref{Eq-schema-base}) is the base case: it
gives sufficient conditions for using~(\ref{Equiv-univ-LHS}) to
translate $P_1$ to $P^\prime$.  The other formula,
(\ref{Eq-schema-conj}), is the recursive case; it expresses
translation of a conjunction in terms of translation of one of its
conjuncts, adding the other conjunct to the conjunctive context as it
does so. The advantage of splitting~(\ref{Eq-schema-1}) into
(\ref{Eq-schema-base}) and~(\ref{Eq-schema-conj}) is that it is then
possible to add rules for other logical operators, each of which
passes around the conjunctive context; we will call rules of this kind
{\it translation-schemas}. We now present some more
translation-schemas, starting with those for quantified expressions.
(\ref{Eq-schema-exists}) and~(\ref{Eq-schema-forall}) express
translation of a quantified form in terms of translation of its body.
In each of them, $\theta$ substitutes a set of unique constants for the
$\vec{x}$.
\begin{eqnarray}
& & {Context \rightarrow (\theta(P) \equiv \theta(P^\prime))} \nonumber \\
& \Rightarrow & Context \rightarrow (\exists\vec{x}.P \equiv \exists
\vec{x}.P^\prime) \label{Eq-schema-exists} \\
& & \nonumber \\
& & {Context \rightarrow (\theta(P) \equiv \theta(P^\prime))} \nonumber \\
& \Rightarrow & Context \rightarrow (\forall\vec{x}.P \equiv \forall
\vec{x}.P^\prime) \label{Eq-schema-forall}
\end{eqnarray}
Next, the
schemas~(\ref{Eq-schema-disj}),~(\ref{Eq-schema-neg}),~(\ref{Eq-schema-impl1})
and~(\ref{Eq-schema-impl2}) express translations of disjunctions,
negations and implications in terms of translations of their component
sub-formulas. Note that the left-hand side is added to the context
when translating the right-side side in~(\ref{Eq-schema-impl2}), but
not {\it vice versa}.
\begin{eqnarray}
& & {Context \rightarrow (P \equiv P^\prime)} \nonumber \\
& \Rightarrow & Context \rightarrow (P \vee Q \equiv P^\prime \vee Q)
\label{Eq-schema-disj}\\
& & \nonumber \\
& & {Context \rightarrow (P \equiv P^\prime)} \nonumber \\
& \Rightarrow & Context \rightarrow (\neg P \equiv \neg P^\prime)
\label{Eq-schema-neg}\\
& & \nonumber \\
& & {Context \rightarrow (P \equiv P^\prime)} \nonumber \\
& \Rightarrow & Context \rightarrow (P \rightarrow Q \equiv P^\prime
\rightarrow Q) \label{Eq-schema-impl1}\\
& & \nonumber \\
& & {Context \wedge P \rightarrow (Q \equiv Q^\prime)} \nonumber \\
& \Rightarrow & Context \rightarrow (P \rightarrow Q \equiv P \rightarrow
Q^\prime) \label{Eq-schema-impl2}
\end{eqnarray}
The proofs of~(\ref{Eq-schema-exists})--(\ref{Eq-schema-impl2}) are
straight-forward.

(\ref{Eq-schema-base})--(\ref{Eq-schema-impl2}) thus define a set of
schemas which can be used to express translations of complex formulas
built up with the connectives $\wedge$, $\exists$, $\forall$, $\vee$,
$\neg$ and $\rightarrow$ in terms of translations of their components.
Together, they define a way in which ``universal'' equivalences can be
used as truth-preserving conditional rewriting rules, licencing
translation of one of the conjuncts on the left-hand side into the
right-hand side in suitable conjunctive contexts.

\subsection{``Existential'' equivalences}\label{Existential-equivalences}

We have so far only considered universal equivalences, i.e.
equivalences of type~(\ref{Equiv-univ-LHS}). As we have seen, the
information used to determine when an universal equivalence is
applicable is passed through the conjunctive context. For
existential equivalences (equivalences of type~(\ref{Equiv-ex-LHS})),
the problem is rather to make sure that they are used in
contexts where the variables are appropriately quantified.
We will basically adopt a minimal solution to this problem,
encoded in the easily-proved translation schema
\begin{eqnarray}
& & \exists\vec{x}.P_1 \equiv P^\prime \nonumber \\
& \Rightarrow & \theta(\exists\vec{x}.P_1) \equiv \theta(P^\prime)
\label{Base-schema-exists}
\end{eqnarray}
where $\theta$ is is one-to-one on $\vec{x}$ (i.e. it does not
map distinct existential variables in $\vec{x}$ into identical ones in
$\theta(\vec{x})$).  At first glance, this seems terribly restrictive,
since it virtually limits application of an existential equivalence
$E$ to sub-formulas that are syntactic variants of $E$'s left-hand
side. Things are actually rather better than they look, however, since
it is possible to exploit the fact that
$$\exists\vec{x}.\exists\vec{y}.(R\wedge P_1)$$ is equivalent with
$$\exists\vec{y}.(R\wedge
\exists\vec{x}.P_1)$$ if the $\vec{x}$ do not occur in $R$. This
equivalence makes it possible to ``move'' existential quantifiers to
the right place in the formula, so that they become associated with
goals that are to be translated using existential equivalences, and in
practice seems to provide all the flexibility needed. It is moreover
possible to ``move'' the existential quantifiers ``on demand'' in a
simple way which interacts cleanly with the rest of the AET
translation process.  The details are presented below in
Section~\ref{AET-process}.  First, however, we define a second way to
use the equivalences -- as normal Horn-clauses -- which as we soon
shall see is also essential to the translation process.

\subsection{Horn-clause readings of equivalences}\label{Horn-clause-readings}

An equivalence of the form
\begin{displaymath}
\forall\vec{x}.(P_1\wedge P_2\wedge \ldots \equiv \exists\vec{y}.Q_1\wedge
Q_2\wedge \ldots) \leftarrow Conds
\end{displaymath}
implies the validity, for any $k$, of all Horn-clauses either of the form
\begin{displaymath}
\forall\vec{x}.\forall\vec{y}.(P_k \leftarrow Q_1\wedge Q_2\wedge \ldots \wedge
Conds)
\end{displaymath}
or
\begin{displaymath}
\forall\vec{x}.(\theta(Q_k) \leftarrow P_1\wedge P_2\wedge \ldots \wedge Conds)
\end{displaymath}
where $\theta$ replaces the $\vec{y}$ with Skolem functions of the $\vec{x}$.
We will refer to these, respectively, as {\it normal} and {\it backward}
Horn-clause readings of the equivalence. For example, the rule
\begin{equation}
woman^\prime(X)\wedge employee^\prime(X)) \equiv
\exists HasCar.EMPL(X,w,HasCar)
\end{equation}
produces two normal Horn-clause readings,
%
\begin{equation}
woman^\prime(X) \leftarrow EMPL(X,w,HasCar)
\end{equation}
\begin{equation}
employee^\prime(X) \leftarrow EMPL(X,w,HasCar)
\end{equation}
and one backward Horn-clause reading,
\begin{equation}
EMPL(X,w,sk_1(X)) \leftarrow woman^\prime(X)\wedge employee^\prime(X)
\end{equation}
where $sk_1$ is a Skolem function.
In the implementation,
each equivalence
is compiled in three different ways, to yield the
normal Horn-clause, backward Horn-clause and equivalential readings.
The Horn-clause readings are used to support proofs from the
conjunctive context. We have experimented with using both kinds of
Horn-clause; at present, though, only the normal Horn-clauses
are used in the implemented system, for reasons discussed in
Chapter~\ref{reasoning}.

\subsection{The abstract AET process}\label{AET-process}

We can now sketch out the basic translation process. The
actual process of translation of a complex formula $F$ is a series of
single translation steps, each of which consists of the translation of
an atomic constituent of $F$. A translation step contains the
following sub-steps, with {\it either} {\bf Translate-universal},
{\it or} {\bf Translate-existential} being applied.
\begin{description}
\item[Recurse:] Descend through $F$ using the translation-schemas, until
an atomic sub-formula $A$ is reached. During this process, a
conjunctive context $E$ has been accumulated in which conditions will
be proved, and some bound variables will have been replaced by unique
constants. Constants resulting from bound variables are specially
marked if i) they come from existentially bound variables that occur
only in $A$, and ii) the only connectives between $A$ and the
binding existential quantifier are conjunctions. We will refer to
these as {\it marked existential constants}.
\item[Translate-universal:] Find a rule $(H\wedge R \equiv B) \leftarrow C$
such that $H$ unifies with $A$ with m.g.u.~$\theta$.  If it is then
possible to prove $\theta(R\wedge C)$, replace $A$ with $\theta(B)$.
The leaves of the proof may include elements of the conjunctive
context $E$, facts from the database,
Horn-clauses from the linguistic domain theory $\Gamma$, or
Horn-clause readings of conditional equivalences from $\Gamma$.
\item[Translate-existential:] Find a rule $\exists\vec{x}.H \equiv B$
such that $H$ unifies with $A$ with m.g.u.~$\theta$. If all the
elements of $\theta(\vec{x})$ are distinct marked existential
constants (in the sense defined immediately above in the {\bf Recurse}
step), then replace $A$ with $\theta(B)$.
\item[Simplify:] if possible, apply simplifications to the resulting formula.
\end{description}
When describing translation, we will often find it convenient, by
analogy with the corresponding terminology for logic programming, to
refer to $A$ in the description above as the {\it selected goal}, $H$
as the {\it head} of the rule used, and $R\wedge C$ (in the case of a
universal rule) as its {\it conditions}.

\section{A simple example}\label{Simple-example}

An example follows to illustrate how the AET process works. In
the interests of expositional clarity we use a grossly over-simplified
version of the actual CLARE domain rules. In particular, we will fail
to distinguish between database objects and real-world objects, since
our current focus is the AET process itself rather than the structure
of the linguistic domain theory. We start with the sentence (S2),
\begin{description}
\item[(S2)] Do any women work on CLARE?
\end{description}
which receives the LF
\begin{equation}
\exists X\exists E.woman^\prime(X)\wedge
work\_on^\prime(E,X,clare^\prime)   \label{Simple-ex-lf-1}
\end{equation}
(\ref{Simple-ex-lf-1}) has to be mapped to a query which accesses two
database relations. The first is the employee relation
$$EMPL(Empl,Sex,HasCar)$$
where $Empl$ represents employees, $Sex$ can be $w$ or $m$, and
$HasCar$ can be $y$ or $n$. The second is the project-member relation
$$PROJMEM(Empl,Project)$$
where $Empl$ again represents the employee and $Proj$ the project.
The desired result is
(\ref{Simple-ex-lf-final}):
\begin{equation}
\exists X\exists HasCar.EMPL(X,w,HasCar)\wedge
             PROJMEM(clare^\prime,X)
\label{Simple-ex-lf-final}
\end{equation}
The most clearly non-trivial part is justifying the conversion between
the linguistic relation $woman^\prime(X)$ and the database relation
$EMPL(X,w,\_)$ Even in a very restricted context, it may well be
incorrect to state that ``woman'' is equivalent to ``employee classed
as being of female sex''; in the implemented PRM domain theory, it is
definitely wrong, since both employees and payees are classified by
sex. It is thus more correct to say that a tuple of type
$EMPL(X,w,\_)$ is equivalent to the conjunction of two pieces of
information: firstly that $X$ is a woman, and secondly that she is an
employee. This can be captured in the rule
\begin{eqnarray}
& woman^\prime(Person)\wedge employee^\prime(Person) \equiv \nonumber \\
& \exists HasCar.DB\_EMPLOYEE(Person,w,HasCar))     \label{Simple-ex-eq-1}
\end{eqnarray}
In the left-to-right direction, the rule can be read procedurally as
follows:
$$woman^\prime(X)$$ translates to $$\exists Y.EMPL(X,w,Y)$$
in contexts where it is possible to prove $$employee^\prime(X)$$
For the rule to be of use in the present example, we must therefore
provide a justification for $employee^\prime(X)$'s holding
in the context of the query.
The simplest way to ensure that this is so is to provide a
Horn-clause meaning postulate,
\begin{equation}
employee^\prime(X) \leftarrow PROJMEM(Project,X)   \label{Simple-ex-hc-1}
\end{equation}
which encodes the fact that people listed in the database as project
members are employees.

Similarly, we will need
an equivalence rule to convert between $work\_on^\prime$
and $PROJMEM$. Here the fact we want to state
is that project-members are precisely people who work on projects,
which we write as follows: we split the equivalence into two
formulas for the reasons explained at the beginning of
Section~\ref{Translation-schemas}.
\begin{eqnarray}
 & work\_on^\prime(Event,Person,Project)\wedge
      project^\prime(Project) \equiv \nonumber \\
 & work\_on\_project^\prime(Event,Person,Project)
\label{Simple-ex-eq-2-1}\\
\nonumber \\
& \exists Event. work\_on\_project^\prime(Event,Person,Project)
  \equiv \nonumber \\
& PROJMEM(Project,Person)
\label{Simple-ex-eq-2-2}
\end{eqnarray}
We will also make indirect use of a rule that states that
projects are objects that can be found in the first field of
a $PROJ$ tuple,
\begin{eqnarray}
& project^\prime(Proj) \equiv \nonumber \\
& \exists ProjNum\exists Start\exists
End.PROJ(Proj,ProjNum,Start,End)
\label{Simple-ex-eq-3}
\end{eqnarray}
since this will allow us to infer (by looking in the database relation
$PROJ$) that the predicate $project^\prime$ holds of $clare^\prime$.

Three translation steps now produce the desired transformation. In
each, the schema for existential quantification
(\ref{Eq-schema-exists}) and the schema for conjunction
(\ref{Eq-schema-conj}) are used in turn to reduce to the base case of
expanding an atom.  Remember that existential quantification schema
replaces variables with unique constants; when displaying the results
of such a transformation, we will consistently write $X*$ to symbolize
the new constant associated with the variable $X$.

The first translation step selects the atomic subformula
of~(\ref{Simple-ex-lf-1}) $$work\_on^\prime(E,X,clare^\prime)$$ which
after replacement of bound variables by constants becomes
$$work\_on^\prime(E*,X*,clare^\prime)$$ The associated conjunctive
context is $\{woman^\prime(X*)\}$. Now the
equivalence~(\ref{Simple-ex-eq-2-1}) can be used, taking the first
conjunct on its LHS as the head; its conditions after unification with
the selected atom are the remaining conjuncts on the LHS, i.e.
$$project^\prime(clare^\prime)$$ The normal Horn-clause reading
of~(\ref{Simple-ex-eq-3}) can be used to reduce the conditons to
$$\exists ProjNum\exists Start\exists
End.PROJ(clare^\prime,ProjNum,Start,End)$$ which can be proved
directly by a call to the database relation $PROJ$. Since the
conditions have been proved to hold,
$work\_on^\prime(E,X,clare^\prime)$ can be replaced by
$work\_on\_project(E,X,clare^\prime)$ in~(\ref{Simple-ex-lf-1}),
yielding the result
\begin{equation}
\exists X\exists E.woman^\prime(X)\wedge
work\_on\_project^\prime(E,X,clare^\prime)   \label{Simple-ex-lf-2}
\end{equation}
The second translation step now uses~(\ref{Simple-ex-eq-2-2}) to
translate the second conjunct of~(\ref{Simple-ex-lf-2}).
(\ref{Simple-ex-eq-2-2}) is an existential equivalence, so there are
no conjunctive conditions; it is on the other hand necessary
to check that the variables are appropriately
bound. The first variable in the LHS of~(\ref{Simple-ex-eq-2-2}) is
existentially bound, so the first variable in the selected goal
must also be existentially bound by a quantifier which is seperated from
the goal only by conjunctions. Recall that these conditions
are required to be able to rewrite~(\ref{Simple-ex-lf-2}) into the form
\begin{equation}
\exists Xwoman^\prime(X)\wedge
(\exists E.work\_on\_project^\prime(E,X,clare^\prime)) \label{Simple-ex-lf-2-1}
\end{equation}
As this is possible,
$work\_on\_project(E,X,clare^\prime)$ in~(\ref{Simple-ex-lf-2})
can be replaced by the body of~(\ref{Simple-ex-eq-2-2}), giving
\begin{equation}
\exists X\exists E.woman^\prime(X)\wedge
PROJMEM(clare^\prime,X)   \label{Simple-ex-lf-3}
\end{equation}
and as quantification over $E$ has become vacuous,~(\ref{Simple-ex-lf-3})
can be simplified to
\begin{equation}
\exists X.woman^\prime(X)\wedge
PROJMEM(clare^\prime,X)   \label{Simple-ex-lf-4}
\end{equation}
We are now in a position to be able to translate the first conjunct
of~(\ref{Simple-ex-lf-4}),
$$woman^\prime(X)$$
After replacing bound variables by
constants, this becomes
$$woman^\prime(X*)$$
and the conjunctive
context is $\{PROJMEM(clare^\prime,X*)\}$. $woman^\prime(X*)$ unifies
with the first conjunct on the left-hand side
of the equivalence~(\ref{Simple-ex-eq-1}), making its conditions
$employee^\prime(X*)$. Using the Horn-clause meaning
postulate~(\ref{Simple-ex-hc-1}), the conditions can be reduced to
$PROJMEM(Project,X*)$. Note that $X*$ in this formula is a constant,
while $Project$ is a variable.  This new goal can be proved directly
from the conjunctive context, instantiating $Project$ to
$clare^\prime$. So the selected goal can be replaced
in~(\ref{Simple-ex-lf-4}) by the body of~(\ref{Simple-ex-eq-1}),
yielding the final result
\begin{equation}
\exists X\exists HasCar.EMPL(X,w,HasCar)\wedge
             PROJMEM(clare^\prime,X)
\end{equation}

\section{Translating higher-order formulas}\label{Higher-order-AET}

It is possible to extend AET without much trouble to provide an adequate
coverage of the higher-order constructs that commonly occur in the
context database querying: the most important are those for counting,
summing and ordering. We will represent these respectively by the
operators $count$, $sum$ and $order$, with the following syntax and
semantics:
$$count(N,\lambda X.P(X))$$
holds if there are precisely $N$
values of $A$ such that $P(A)$ holds;
$$sum(S,\lambda X.P(X))$$
holds if all
the objects $A$ of which $P(A)$ holds are summable quantities, and $S$
is their sum; and
$$order(Selected,\lambda X.\lambda D.P(X,D),Ordering)$$
holds if $Ordering$ is an ordering relation, $D_{max}$ is the maximal
$D$ under the relation $Ordering$
such that $P(X,D)$ holds for some $X$, and $Selected$ is such an $X$.

Expressions formed using these operators can be translated by
translating their bodies, replacing the bound variables by
unique constants as in the schemas for universal and existential
quantifiers shown earlier. For example, the schema for $count$
is
\begin{eqnarray}
& & {Context \rightarrow (\theta(P) \equiv \theta(P^\prime))} \nonumber \\
& \Rightarrow & Context \rightarrow (count(N,\lambda X.P) \equiv
count(N,\lambda X.P^\prime)) \label{Eq-schema-count}
\end{eqnarray}
where $\theta$ replaces $X$ with a unique constant.

As can be seen, the conjunctive context is passed into the body of the
higher-order form and is available for rules used to translate it.
What is more interesting to consider is the extent to which the
material in the bodies of these higher-order forms can be ``exported''
out to add to the context of goals conjoined with higher-order forms;
this is frequently desirable.  The solution we have adopted is to
assume that the $\lambda$-bound properties appearing in the bodies of
$sum$, $count$ and $order$ forms always have non-empty extensions, in
other words that the following hold:
\begin{eqnarray}
&count(N,\lambda X.P(X)) \rightarrow \exists X_1.P(X_1) \\
\nonumber \\
&sum(N,\lambda X.P(X)) \rightarrow \exists X_1.P(X_1) \\
\nonumber \\
&order(Selected,\lambda X.\lambda D.P(X,D),Ordering) \rightarrow
\nonumber \\
&\exists D_1. P(Selected,D_1) \label{Order-context-add-formula}
\end{eqnarray}
This means, for instance, that given a formula of type
$$C\wedge count(N,\lambda X.P(X))$$
it is permissible to assume $\exists Y.P(Y)$ when translating $C$.
A special case is presented by the $order$ operator; as can be
seen from~(\ref{Order-context-add-formula}), it is possible to make
a stronger assumption, substituting the maximal element (rather than
an existentially quantified one) in the outermost lambda-binding.

\section{Abductive translation}\label{Abductive-translation}

We now put the ``A'' into ``AET'', and describe how the basic
framework can be extended further to allow abductive translation.
There are essentially two problems. Firstly, we want if necessary to
be able to make abductive assumptions when performing the {\bf
Translate-universal} step of the AET process described
Section~\ref{AET-process}: in other words, we want to be able to
define situations where some of the information required to license a
translation step is assumed rather than proved.  Goals will be classed
as potential abductive assumptions if they ought to be assumed true
when there is no specific evidence to the contrary. This brings us to
our second requirement: that we should be able to define ways in which
an abductive assumption is contradicted by its context of occurrence.

These two functionalities are achieved as follows. We include
declarations of the form
\begin{verbatim}
assumable(Goal,Cost,Justification,Type,Cond)
\end{verbatim}
where {\tt Goal} and {\tt Cond} are atomic formulas, {\tt Cost} is a
non-negative integer, {\tt Justi\-fic\-ation} is a unique tag that
names the assumption, and {\tt Type} is the assumption-type (cf
Section~\ref{Basic-LDT}).  The intended semantics are that {\tt Goal}
may be assumed at cost {\tt Cost} in an environment where {\tt Cond}
holds. The ``justification'' can be used to identify the assumption to
the user. If an abductive assumption $A$ is made while translating the
goal $G$, it is saved together $G$'s associated conjunctive context
$Context$. Assumptions are considered to be inconsistent if
$$\Gamma\cup Context \rightarrow \neg A$$ holds, where $\Gamma$ is the
linguistic domain theory. The implemented system has so far only a
fairly primitive ability to find proofs of negated goals (cf
Sections~\ref{Negated-proofs} and~\ref{Horn-clause-axioms}).

Examples of how assumption declarations are used will be presented
later in Sections~\ref{Database-specific-axioms}
and~\ref{Assumption-declarations}.

\section{Simplification}\label{Simplification}

AET, like many systems based on the idea of rewriting, tends to suffer
from the problem that expressions can rapidly grow in size as they
pass through successive stages of translation.  The standard solution
to the problem is to include a {\it simplifier}, a module which takes
the output from a translation stage and attempts to reduce it to a
more compact form (cf e.g. (Williams 1991)). This section will describe
CLARE's simplifier. Although the simplifier's primary use is in
conjunction with the translator, it has turned out to provide a
functionality that several other parts of the system utilise.  In
section~\ref{Simplify-assertions} we describe how the simplifier is
used in the processing of assertions; here, the problem is to take the
conjunction of the TRL representations of several assertions, and
combine them into a compact form.

The simplifier consists of two main pieces. The first, described in
section~\ref{Normal-simplification}, is a collection of standard
logical rewriting techniques, none of which use inference, that
have the common function of reducing expressions to a canonical
form. Simplification can also be carried out by using the inference-based
equivalential translation methods described earlier in this chapter,
in which inference justifies the replacement of sub-expressions by
other forms, the equivalence of which is implied by the context.
This second type of simplification is described in
section~\ref{Functional-relations}.

\subsection{Non-inferential simplification}\label{Normal-simplification}

The non-inferential simplification module consists of the set of
rewriting methods listed below. Most of them are fairly obvious.  We
present the methods in the order in which they are applied by the
system.

\subsubsection{Moving existential quantifiers outwards}

This method consists of two kinds of rewriting rule.  The first
simplifies an expression built out of conjunction and existential
quantification operators into an equivalent form with a single
existential operator on the outside binding all the variables (it has
proven convenient to slightly extend standard predicate calculus
notation by allowing a quantifier to bind an arbitrary number of
variables).  Thus for example $$\exists x. p(x) \wedge\exists y.q(y)$$
will get rewritten to $$\exists x,y. p(x) \wedge q(y)$$ The other
rewriting rule is applied when an existentially quantified form occurs
on the LHS of an implication; in this case, the existentially
quantified variables are moved upwards to become universally
quantified with wide scope over the implication.  The justification
for this manoeuvre is the equivalence
\begin{displaymath}
((\exists x.P) \rightarrow Q)) \equiv \forall x.(P \rightarrow Q)
\end{displaymath}
which is easily proved. The result of recursively applying the two
rules is to give existential quantifiers as wide scope as possible,
if necessary turning them into universals on the way.

\subsubsection{Simplification of equalities}

The next simplification step removes equalities from the expression
when this is possible. The basic equivalences that licence this
type of simplification are the following:
\begin{displaymath}
(\exists x.\exists y.P\wedge (x = y)) \equiv (\exists x. P[y/x])
\end{displaymath}
\begin{displaymath}
(\forall x.\exists y.P\wedge (x = y)) \equiv (\forall x. P[y/x])
\end{displaymath}
and
\begin{displaymath}
(\exists y.P\wedge (y = a)) \equiv P[y/a]
\end{displaymath}
where $a$ is a constant. The methods are implemented by recursively
descending through the expression to find equality sub-forms, and
in suitable cases unifying together their two arguments, replacing
the equality with an occurrence of the identically true predicate.
After this, a second pass removes the occurrences of $true$ and the
vacuous and repeated quantifications introduced.

If the expression contains an equality whose arguments consist of two
non-unifiable constants $Arg_1$ and $Arg_2$, its value can be known to
be identically false (assuming non-identity of reference of distinct
constants).  In this case, the simplifier replaces the equality with
the form $$mismatch(Arg_1,Arg_2)$$ Subsequent behaviour depends on the
nature of the expression being manipulated.  If it is a question, the
$mismatch$ form is later replaced with an occurrence of the
identically false predicate, which will generally lead to a ``No''
answer being produced. If the expression however represents an
assertion, the mismatch normally represents a presupposition failure.
In this case, the interface handler will inform the user that the
expression was translated to an identically false form due to the
occurrence of the mismatch, and attempt to print the incompatible
arguments in an informative way (cf.
Section~\ref{Declaration-functionality}).

\subsubsection{Simplification of ground sub-expressions}

The final type of simplification involves ground sub-expressions,
that is to say sub-expressions containing no variables. Since
these can be assigned a truth-value irrespective of their context,
the expression can be simplified by computing it and replacing
by an occurrence of either $true$ or $false$. At present, the method is
restricted in two ways: ground sub-expressions are only replaced
if their predicates are declared to be executable relations
(cf. section~\ref{Effective-translation}); also, only {\it true}
ground expressions are substituted.

\subsection{Inferential simplification}\label{Functional-relations}
\label{Inferential-simplification}

We will now consider methods that use equivalential translation to
carry out simplification; the notion of {\it conjunctive context},
defined in section~\ref{Translation-schemas}, will play a key role.
We can immediately describe one way to achieve this end. Suppose that
$F$ is a formula containing an occurrence of the subformula
$F^\prime$, and suppose further that $F^\prime$ is implied by its
conjunctive context. It follows that replacing $F^\prime$ with $true$
in $F$ will result in a formula equivalent with $F$. This gives a
method can remove duplicated conjuncts and the like, and also perform
less obvious simplifications. To take a simple example, consider the
formula $$P\wedge Q\rightarrow P\wedge R$$ The conjunctive context of
the second occurrence of $P$ contains $P$, so the formula can be
simplified to $$P\wedge Q\rightarrow R$$ A slightly more complex
example is provided by the formula $$P\rightarrow Q\wedge R$$ where we
are also told that $Q$ can be derived from $P$. Then as we know that
$P$ is in $Q$'s conjunctive context, it is possible to remove $Q$ and
simplify to $$P\rightarrow R$$

A similar, but more refined, form of simplification can also be
performed, which exploits functional relationships between arguments
in predicates. We start by re-examining our example {\bf (S1)} from the
beginning of chapter~\ref{translation}, reproduced here for
convenience.
\begin{description}
\item[(S1)] Show all payments made to BT during 1990.
\end{description}
In section~\ref{Examples}, we will see that, when the example is
processed by the implemented CLARE system, the logical form originally
derived from {\bf (S1)} is
\begin{verbatim}
forall([Trans],
       impl(x([Agnt,Ev],
              and(payment1(Trans),
                  and(make2(Ev,Agnt,Trans,payee1#bt),
                      during1(Ev,year1#1990)))),
            x([ShowEv],
              and(show1(ShowEv,clare,Trans),
                  before1(<Now>,ShowEv)))))
\end{verbatim}
After a few translation steps, the resulting formula contains three
separate instances of the \verb!transaction! relation, one from each
of the original linguistic predicates
\verb!payment1!, \verb!make2! and
{\tt during1}. Assuming that the arguments of the {\tt transaction}
relation are
\begin{verbatim}
transaction(Transaction,Payer,Date,Payee,Amount)
\end{verbatim}
{\tt payment1(Payment)} expands, roughly speaking, to
\begin{verbatim}
transaction(Payment,_,_,_,_)
\end{verbatim}
{\tt make2(Event,Agent,Payment,Payee)} expands to
\begin{verbatim}
transaction(Payment/Event,Agent,_,Payee,_)
\end{verbatim}
(the {\tt Payment} and {\tt Event} arguments map into the same
variable). Finally, {\tt during1(Payment,Interval)} expands to
\begin{verbatim}
and(transaction(Payment,_,Date,_,_),
    time_during(Date,Interval))
\end{verbatim}
The database query will
conjoin all three instances together. It is clearly preferable, if
possible, to merge them instead,
yielding a composite predication
\begin{verbatim}
transaction(Payment/Event,Agent,Date,Payee,_)
\end{verbatim}
The information that licences this step as a valid simplification is
that \verb!transaction! is a function from its first argument
(the payment) to the remaining ones (the agent, the date,
the payee and the amount); in other words, a given payment is made by a unique
agent, on a unique date, to a unique payee, for a unique amount.
The system allows the information
to be entered as a ``function'' meaning postulate in the form
\begin{verbatim}
function(transaction(Transaction,Payer,Date,Payee,Amount),
         [Transaction] -> [Payer,Date,Payee,Amount])
\end{verbatim}
which is treated as a concise notation for the conditional equivalence
\begin{eqnarray}
&transaction(Trans,Payer^\prime,Date^\prime,Payee^\prime,Amount^\prime)
\rightarrow \nonumber \\
&(transaction(Trans,Payer,Date,Payee,Amount) \equiv \nonumber \\
&Payer = Payer^\prime\wedge Date = Date^\prime\wedge \nonumber \\
&Payee = Payee^\prime\wedge Amount = Amount^\prime)
\end{eqnarray}
It is thus possible to view ``merging'' simplification of this kind as
just another instance of equivalential translation. In the current
version of the system, the transformation process operates in a cycle,
alternating normal translation followed by simplification using the
same basic interpreter; simplification consists of functional
``merging'' followed by reduction of equalities where this is
applicable.

\subsection{Simplification for processing
assertions}\label{Simplify-assertions}

The simplification process plays an important role in the processing
of assertions (cf Section~\ref{Declaration-functionality}).
Consider, for example, what would happen to the pair of sentences {\bf
(Assertion1)} and {\bf (Assertion2)} without simplification:
\begin{description}
\item[(Assertion1)] Clara is an employee who has a car.
\item[(Assertion2)] Clara is a woman.
\end{description}
Assuming a suitable extension of the axioms used in
Section~\ref{Simple-example} above, {\bf Assertion1} translates into
the database form $$\exists A.EMPL(clara^\prime,A,y)$$ (Recall
that the second field in the $EMPL$ relation indicates sex, and the
third whether or not the employee has a company car).  This can then
be put into Horn-clause form as $$EMPL(clara^\prime,sk_1,y)$$ (where
$sk_1$ is a Skolem constant) to make it accessible to inferencing
operations. Since Clara is now known to be an employee, {\bf
Assertion2} will eventually translate into the unit clause
$$EMPL(clara^\prime,w,sk_2)$$ with $sk_2$ a second Skolem constant.
The two clauses produced would contain all the information entered,
but they could not be entered into a relational database as they
stand; a normal database has no interpretation for the Skolem
constants $sk_1$ and $sk_2$.

However, it is possible to use function information to merge the two
clauses into a single record. The trick is to arrange things so that
the system can when necessary recover the existentially quantified
form from the Skolemized one; assertions which contain Skolem
constants are kept together in a cache called the {\it logical
discourse context}.  Simplification of assertions then proceeds
according to the following sequence of steps:
\begin{enumerate}
\item Retrieve all assertions from the logical discourse context cache.
\item Construct a formula $A$, which is their logical conjunction.
\item Let $A_0$ be $A$, and let $\{sk_1\ldots sk_n\}$ be the
Skolem constants in $A$. For $i = 1\ldots n$, let
$x_i$ be a new variable, and let $A_i$ be
the formula $\exists x_i.A_{i-1}[sk_{i}/x_i]$, i.e. the result of
replacing $sk_i$ with $x_i$ and quantifying existentially over
it.
\item Perform normal function merging on $A_n$, and call
the result $A^\prime$.
\item Convert $A^\prime$ into Horn-clause form, and replace
the result in the logical discourse context cache.
\end{enumerate}
In the example above, this works as follows. After {\bf (Assertion1)}
and {\bf (Assertion2)} have been processed, the logical discourse
context contains the clauses
$$EMPL(clara^\prime,sk_1,y)$$
$$EMPL(clara^\prime,w,sk_2)$$
$A = A_0$ is then the formula
$$EMPL(clara^\prime,sk_1,y)\wedge EMPL(clara^\prime,w,sk_2)$$
and $A_2$ is
$$\exists X_1,X_2.EMPL(clara^\prime,X_1,y)\wedge EMPL(clara^\prime,w,X_2)$$
If $EMPL$ is declared functional on its first
argument, the second conjunct can be reduced to two equalities, giving
the formula
$$\exists X_1,X_2.EMPL(clara^\prime,X_1,y)\wedge X_1 = w\wedge y = X_2$$
which finally simplifies to $A^\prime$,
$$EMPL(clara^\prime,w,y)$$
a record without Skolem constants, which can be added to a normal
relational database.

\section{The ``Doctor on Board'' problem}\label{Doctor-on-board}

One of the best-known problems in the area of logic-based
natural-language database interfacing is posed by so-called ``Doctor
on Board'' queries (Perrault and Grosz, 1988). These are the standard
examples of the problems that arise when existentially quantified
variables in the logical form fail to correspond to any quantified
variable in the final evaluable expression. The original example came
from a naval information database, in which the main relation contained
information about the position and other attributes of a set of ships;
one field was set to $y$ if there was a doctor on board the ship in
question.  Now consider the three queries {\bf (DOB1--3)} below:
\begin{description}
\item[(DOB1)] Is there a doctor on board ship FOO?
\item[(DOB2)] Who is the doctor on board ship FOO?
\item[(DOB3)] Is there a doctor within 500 miles of ship FOO?
\end{description}
It is clear that a query like {\bf (DOB1)} should be answerable; it is
equally clear that a query like {\bf (DOB2)} should not. Most
interestingly, queries like {\bf (DOB3)} should also be answerable,
since the position of a doctor can be assumed to be the same as the
position of his ship.

In this section, we will sketch out a treatment of the ``Doctor on
board'' problem within the AET framework; we will abstract away as
many of the details as possible. In accordance with the ideas
described in Sections~\ref{YNQ-functionality}
and~\ref{Command-functionality}, we begin by assigning the original
logical forms~(\ref{DOB1-lf-1})--(\ref{DOB3-lf-1}) to {\bf DOB1--3}:
\begin{eqnarray}
\exists X.doctor^\prime(X)\wedge on\_board^\prime(X,foo^\prime)
\label{DOB1-lf-1}\\
\nonumber \\
\forall X.(doctor^\prime(X)\wedge on\_board^\prime(X,foo^\prime) \rightarrow
\nonumber \\
displayed\_in\_future(X)) \label{DOB2-lf-1}\\
\nonumber \\
\exists X.doctor^\prime(X)\wedge within^\prime(X,foo^\prime,500)
\label{DOB3-lf-1}
\end{eqnarray}
We will assume that the basic
database predicate we are mapping to is
$$SHIP(ShipId,Position,DoB)$$
where $Position$ is a position expressed according to some suitable
co-ordinate system, and $DoB$ has the value $y$ if there is a doctor
on board the ship, $n$ otherwise. We will require $SHIP$ to be
functional on its first argument (i.e. the field $ShipId$ will be a
primary key). There will also be two evaluable predicate
$$dist(Position_1,Position_2,Distance)$$
and
$$X < Y$$
The first of these computes the
distance (as a number) between two positions, according some suitable
distance metric; the second is the standard arithmetic inequality
relation.

We are now in a position to write down the axioms we will need. The
first three define the relationship between the word-sense predicates
$ship^\prime$, $doctor^\prime$ and $on\_board^\prime$, and the
database predicate $SHIP$:
\begin{eqnarray}
&ship^\prime(S) \equiv \exists Pos\exists DoB.SHIP(S,Pos,DoB)
\label{DOB-eq-ship} \\
\nonumber \\
&doctor^\prime(D)\wedge on\_board^\prime(D,S) \equiv \nonumber \\
&doctor\_on\_board(D,S) \label{DOB-eq-doctor} \\
\nonumber \\
&\exists D.doctor\_on\_board(D,S) \equiv \nonumber \\
&\exists Pos.SHIP(S,Pos,y) \label{DOB-eq-doctor-on-board}
\end{eqnarray}
Axiom~(\ref{DOB-eq-within}) defines the semantics of the word-sense predicate
$within^\prime$ in terms of positions and distances:
\begin{eqnarray}
&within^\prime(X_1,X_2,Radius) \equiv \nonumber \\
&pos(X_1,P_1) \wedge pos(X_2,P_2) \wedge dist(P_1,P_2,D)
\wedge D < Radius \label{DOB-eq-within}
\end{eqnarray}
The last two axioms translate the intermediate predicate
$pos$; two rules are provided, one for positions of ships and
the other for positions of doctors.
\begin{eqnarray}
&ship^\prime(S) \rightarrow \nonumber \\
&(pos(S,P)\equiv \exists DoB.SHIP(S,P,DoB)) \label{DOB-eq-pos-ship} \\
\nonumber \\
&doctor^\prime(D) \rightarrow \nonumber \\
&(pos(D,P)\equiv (\exists S.ship^\prime(S)\wedge
on\_board(D,S)\wedge pos(S,P))) \label{DOB-eq-pos-doctor}
\end{eqnarray}
Finally, we will assume that the database shows $foo^\prime$ to be a
ship, i.e. that there is a $SHIP$ record whose first field contains
an instance of $foo^\prime$.

Let us now first consider how the axioms we have just presented can be
used to translate the logical form corresponding to the query {\bf
(DOB1)}. The original LF is
\begin{eqnarray}
\exists X.doctor^\prime(X)\wedge on\_board^\prime(X,foo^\prime)
\label{DOB1-lf-1-1}
\end{eqnarray}
Using the existential and conjunction translation schemas in the way
described in Section~\ref{Simple-example}, we choose as the selected
goal the second conjunct: after replacement of bound variables by
constants, this is
$$on\_board^\prime(X*,foo^\prime)$$
and the conjunctive context is $$\{doctor^\prime(X*)\}$$
We use the equivalence~(\ref{DOB-eq-doctor}) to translate; the
conditions can be proved directly from the conjunctive context,
and the result after substitution is:
\begin{eqnarray}
\exists X.doctor^\prime(X)\wedge doctor\_on\_board(X,foo^\prime)
\label{DOB1-lf-2}
\end{eqnarray}
The first conjunct of~(\ref{DOB1-lf-2}) can be simplified away
(see section~\ref{Inferential-simplification}); since
one of the normal Horn-clause readings
of~(\ref{DOB-eq-doctor}) is
\begin{eqnarray}
doctor^\prime(X) \leftarrow doctor\_on\_board(X,S)
\end{eqnarray}
we have that $$doctor^\prime(X)$$ is implied by its
conjunctive context $$\{doctor\_on\_board(X,foo^\prime)\}$$
and can thus be removed.
This reduces~(\ref{DOB1-lf-2}) to
\begin{eqnarray}
\exists X.doctor\_on\_board(X,foo^\prime) \label{DOB1-lf-3}
\end{eqnarray}
Since $X$ is existentially quantified and only occurs in the atomic
sub-formula $$doctor\_on\_board(X,foo^\prime)$$
it is now possible to apply~(\ref{DOB-eq-doctor-on-board}), to translate
to the final evaluable form
\begin{eqnarray}
\exists Pos.SHIP(foo^\prime,Pos,y) \label{DOB1-lf-4}
\end{eqnarray}
We next examine {\bf (DOB2)}, which receives the initial LF
\begin{eqnarray}
\forall X.(doctor^\prime(X)\wedge on\_board^\prime(X,foo^\prime) \rightarrow
\nonumber \\
displayed\_in\_future(X)) \label{DOB2-lf-1-1}
\end{eqnarray}
By a similar sequence of steps to those used for {\bf (DOB1)}
above,~(\ref{DOB2-lf-1-1}) can be translated as far as the formula
\begin{eqnarray}
\exists X.doctor\_on\_board(X,foo^\prime) \rightarrow
displayed\_in\_future(X) \label{DOB2-lf-2}
\end{eqnarray}
In contrast to the previous case, however, it is impossible to
translate~(\ref{DOB2-lf-2}) any further;~(\ref{DOB-eq-doctor-on-board})
is this time not applicable, as $X$ is bound by a universal
quantifier rather than an existential, and moreover occurs on
the RHS as well.

We finally turn to {\bf (DOB3}), with initial LF
\begin{eqnarray}
\exists X.doctor^\prime(X)\wedge within^\prime(X,foo^\prime,500)
\label{DOB3-lf-1-1}
\end{eqnarray}
Omitting the details, we can first use~(\ref{DOB-eq-within}) to
translate~(\ref{DOB3-lf-1-1}) into
\begin{eqnarray}
\exists X,P_d,P_{foo},D.doctor^\prime(X)\wedge pos(X,P_d) \wedge
pos(foo^\prime,P_{foo}) \wedge \nonumber \\
dist(P_d,P_{foo},D)\wedge D < 500 \label{DOB3-lf-2}
\end{eqnarray}
Since the goal $doctor^\prime(X*)$ is a member of the conjunctive
context of the second conjunct, $pos(X,P_d)$, it is possible to
translate it using~(\ref{DOB-eq-pos-doctor}), giving
\begin{eqnarray}
\exists X,P_d,P_{foo},S,D.doctor^\prime(X)\wedge ship^\prime(S)\wedge
on\_board(X,S)\wedge \nonumber \\
pos(S,P_d)\wedge pos(foo^\prime,P_{foo})
\wedge dist(P_d,P_{foo},D)\wedge D < 500
\label{DOB3-lf-3}
\end{eqnarray}
By the same methods as those used for {\bf (DOB1)} above, we can then
translate the conjunct $on\_board(X,S)$, and simplify away
the conjuncts $doctor^\prime(X)$ and $ship^\prime(S)$, yielding
\begin{eqnarray}
\exists X,P_d,P_s,P_{foo},S,D.SHIP(S,P_s,y)\wedge
pos(S,P_d)\wedge \nonumber \\
pos(foo^\prime,P_{foo}) \wedge
dist(P_d,P_{foo},D)\wedge D < 500
\label{DOB3-lf-4}
\end{eqnarray}
We now want to translate the two $pos$ conjuncts,
using~(\ref{DOB-eq-pos-ship}). In the first case, the selected atom
is $$pos(S*,P_d*)$$ The conditions deriving from~(\ref{DOB-eq-pos-ship})
are $$ship^\prime(S*)$$ which can be proved using the normal Horn-clause
reading of~(\ref{DOB-eq-ship}), i.e.
$$ship^\prime(X) \leftarrow \exists P,DoB.SHIP(X,P,DoB)$$
The atom
$$SHIP(S*,P_s*,y)$$
can be proved from the conjunctive context. The result is
\begin{eqnarray}
\exists X,P_d,P_s,DoB,P_{foo},S,D.SHIP(S,P_s,y)\wedge
SHIP(S,P_d,DoB)\wedge \nonumber \\
pos(foo^\prime,P_{foo}) \wedge dist(P_d,P_{foo},D)\wedge D < 500
\label{DOB3-lf-5}
\end{eqnarray}
Since $SHIP$ is functional on its first argument it is possible
to use functional merging (cf. Section~\ref{Functional-relations})
to merge the first and second conjuncts, giving
\begin{eqnarray}
\exists X,P_s,P_{foo},S,D.\wedge SHIP(S,P_s,y)\wedge
\nonumber \\
pos(foo^\prime,P_{foo}) \wedge dist(P_s,P_{foo},D)\wedge D < 500
\label{DOB3-lf-6}
\end{eqnarray}
Lastly, the remaining $pos$ conjunct can be translated,
using the equivalence~(\ref{DOB-eq-pos-ship}), the normal Horn-clause
reading of~(\ref{DOB-eq-ship}) and the database relation
$SHIP$. This allows $$pos(foo^\prime,P_{foo})$$ to be
replaced with $$\exists DoB_{foo}.SHIP(foo^\prime,P_{foo},DoB_{foo})$$
giving the final evaluable form
\begin{eqnarray}
\exists X,P_s,P_{foo},DoB_{foo},S,D.SHIP(S,P_s,y)\wedge \nonumber \\
SHIP(foo^\prime,P_{foo},DoB_{foo}) \wedge \nonumber \\
dist(P_s,P_{foo},D)\wedge D < 500
\label{DOB3-lf-7}
\end{eqnarray}

\section{The translation engine}\label{AET-Engine}

This section sketches the CLARE translation engine, the existing
Prolog implementation of the abstract AET process described earlier in
Section~\ref{AET-process}. The structure of the translation engine is
fairly similar to that of a Prolog meta-interpreter. There is a main
predicate,
\begin{verbatim}
translate(+In,-Out,+Context,+MarkedConstants,+Ain,-Aout)
\end{verbatim}
The intended semantics are that {\tt Out} is a translation of {\tt In}
when it occurs with conjunctive context {\tt Context}, and the
variables in the list {\tt Marked} are constants deriving from
existentially bound variables, that are ``specially marked'' in the
sense described in Section~\ref{AET-process}; {\tt Ain-Aout} is a
difference list holding the abductive assumptions made while
performing the translation. There is one or more {\tt translation}
clause for each logical operator {\tt Op}, each of which encodes an
appropriate ``translation schema'' for {\tt Op} (cf.
Section~\ref{Translation-schemas}).

The clause encoding the base case is applied when the main functor of
{\tt In} is a predicate symbol, rather than a logical operator.  Here,
an attempt is made to find an equivalence that can be used to
translate {\tt In}. Equivalences are stored in the compiled
form
\begin{verbatim}
equiv(Head,Conds,ExVars,Body,Group)
\end{verbatim}
where
\begin{itemize}
\item {\tt Head} is a potential ``head'' of the rule, i.e. an atomic goal on
its LHS;
\item for universal rules, {\tt Conds} are the remaining LHS
conjuncts;
\item for existential rules, {\tt ExVars} are the existentially
quantified LHS variables;
\item {\tt Body} is the RHS, and
\item {\tt Group} is an atom that associates the rule with a named
``rule-group''.
\end{itemize}
Prolog variables are used to represent both universal and existential
variables in the rule. The last argument, {\tt Group}, also demands
some explanation; it is frequently useful, for modularity reasons, to
be able to divide equivalences into distinct named groups. A
declaration is also supplied that arranges the groups in a linear
sequence $Group_1...  Group_n$. Translation is first carried out as
far as possible using only the rules from $Group_1$; then again using
only the rules from $Group_2$; and so on.

When trying to apply either a universal or an atomic rule to
translate an atomic goal, the first step is to unify {\tt Head} with
{\tt In}, which consequently instantiates {\tt Conds}, {\tt ExVars}
and {\tt Body} as well.  Behaviour now diverges depending on whether
the rule is universal or existential:
\begin{itemize}
\item If the rule is universal, an
attempt is made to prove {\tt Conds} in the current conjunctive
context; this in effect consists of sending the formula
\begin{verbatim}
impl(Context,Conds)
\end{verbatim}
and the current list of abductive assumptions {\tt Ain} to the
inference component (cf. Chapter~\ref{reasoning}). If a proof can be
found, the call returns with a set of bindings for the free
variables in {\tt Conds} and an updated list of abductive assumptions
{\tt Aout}.
\item  If the rule is existential, then the instantiated list {\tt
ExVars} from the rule and the list {\tt Marked\-Constants} from the
original call to {\tt translate} are passed to the predicate {\tt
check\_\-equiv\_\-exvars/2}, which succeeds if the members of {\tt
ExVars} are distinct members of {\tt Marked\-Constants}.
\end{itemize}

\section{Debugging equivalential theories}\label{AET-debugger}

To be practically useful, a programming language normally has to
support some kind of interactive debugging tool. It has proved simple
to build such a tool on top of the translation engine described in the
last section; the functionality it offers is similar to that provided
by the standard Prolog ``Byrd-box'' debugger. This has been enough to
make it feasible to construct and debug large equivalential linguistic
domain theories, containing several hundred equivalences and
Horn-clauses.

The top-level commands available to the user are similar to those
in the normal Byrd-box debugger, and are summarized below:
\begin{quote}
\begin{verbatim}
aet_debug_init
\end{verbatim}
\end{quote}
Initializes AET debugger. It is then possible to use the other
commands.
\begin{quote}
\begin{verbatim}
aet_trace
\end{verbatim}
\end{quote}
Switches on the AET debugger, in ``trace'' mode.
\begin{quote}
\begin{verbatim}
aet_debug
\end{verbatim}
\end{quote}
Switches on the AET debugger, in ``debug'' mode.
\begin{quote}
\begin{verbatim}
aet_nodebug
\end{verbatim}
\end{quote}
Switches off the AET debugger.
\begin{quote}
\begin{verbatim}
aet_spy(+PredSpec,+Mode)
\end{verbatim}
\end{quote}
Puts an AET spypoint of type {\tt Mode} on {\tt PredSpec}.
{\tt Mode} must be either {\tt translation} or {\tt inference}.
{\tt PredSpec} must be of the form {\tt F/N}.
\begin{quote}
\begin{verbatim}
aet_nospy(+PredSpec,+Mode)
\end{verbatim}
\end{quote}
Removes an AET spypoint of type {\tt Mode} on {\tt PredSpec}.
{\tt Mode} must be either {\tt translation} or {\tt inference}.
{\tt PredSpec} must be of the form {\tt F/N}.
\begin{quote}
\begin{verbatim}
aet_nospyall
\end{verbatim}
\end{quote}
Removes all AET spypoints.

The AET debugger is implemented on top of the normal Quintus Prolog
debugger, using the Quintus Prolog ``advice'' package. In view of the
extremely implementation-dependednt nature of the debugger, we will
confine our description to a very brief sketch.  The basic idea is to
put normal spypoints on the central predicates (in particular {\tt
translate} and {\tt prove}) that implement the translation and
reasoning engines. ``Advice'' is added at all four ports of these key
predicates, which checks the arguments passed to the predicate, the
defined AET spypoints and the current mode, and if necessary switches
the associated spypoint on or off. A special ``portray'' predicate is
employed to format the predicate calls shown by the normal Quintus
debugger in an easily readable way. Thus for example a completed call
to the base case of {\tt translate} is formatted as illustrated below:
\begin{verbatim}
Translating atomic sub-formula

<InFormula>

<Result>

Assumptions made: <Assumptions>
\end{verbatim}
where \verb!<InFormula>!, \verb!<Result>! and \verb!<Assumptions>! are
formatted representations of the input and output of the translation,
and the abductive assumptions needed to perform it.  Similarly, a
completed attempt by the inference component to prove a goal is
formatted as
\begin{verbatim}
Prove (<CostIn> units in, <CostOut> units out):

<Formula>

Assumptions made: <Assumptions>
\end{verbatim}
where \verb!<Formula>! and \verb!<Assumptions>! are the formula that
has been successfully proved and the abductive assumptions used to
prove it, \verb!<CostIn>! was the number of cost units available
before the goal was attempted (cf Section~\ref{Inference-search}), and
\verb!<CostOut>! was the number of cost units left afterwards.


\chapter{Linguistic Domain Theories}
\label{claredm}\label{dmi}

\section{Introduction}

 From the examples presented in the previous chapter, the reader will
already have gained some impression of what a linguistic domain theory
looks like. In this chapter, we will consider the subject in more
detail.  We first describe the various types of information that
compose an LDT: equivalences, Horn-clauses, declarations of functional
relationships and assumption declarations. The rest of the chapter
describes the LDT for the PRM domain, which we will use
as a generic example.

There are four main kinds of information in the PRM LDT. The greater
part of the theory is composed of conditional equivalences, which have
already been discussed at length in chapter~\ref{AET}.  In this
chapter, it will be convenient to allow equivalences to be written in
the concrete TRL notation illustrated by the following example (cf
also Appendix~\ref{trlsem})
\begin{verbatim}
exists([Event],
  and(work_on1(Event,Person,Project),
      project1(Project))) <->
works_on_project(Event,Project,Person)
\end{verbatim}
Apart from the conditional equivalences, there are Horn-clause
formulas, declarations of functional relationships between arguments
of predicates, and declarations of what assumptions are permissible.
We will first review each of these briefly; later on in the chapter,
we explain more exactly how they fit into the LDT.
\begin{description}
\item[Horn-clauses]
Certain relationships in the PRM LDT are inherently ones of
implication rather than equivalence; in these cases, Horn-clauses,
rather than equivalences, are appropriate. For example, it is
necessary to capture the fact that everyone who books time to a
project is an employee.  This cannot easily be made into an
equivalence, since being an employee does not directly imply that one
books time to any particular project. Instead, we use the Horn-clause
\begin{verbatim}
employee_Worker(Person) <-
  booking_to_project(Event,Person,Proj,Hours,Date)
\end{verbatim}
\item[Functional relationships]
Many relations have the property of being functional on some subset of
their arguments; that is to say, selecting a particular given value
for that subset determines a unique value for the remaining arguments.
In practice, the most important case is that of {\it primary key}
fields in database predicates. Functional relationships are defined by
declarations of the form
\begin{verbatim}
function(<Template>,<FunctionalArgs> -> <RemainingArgs>)
\end{verbatim}
In all the examples occurring in the PRM LDT, \verb!<FunctionalArgs>!
is a list consisting of a single argument. For example, the first
argument of the {\tt TRANS} relation is an identifier that functions
as a primary key: this is captured by the declaration
\begin{verbatim}
function(TRANS(TransId,ChequeNum,Date,Payee,Acct,Amount),
         [TransId] -> [ChequeNum,Date,Payee,Acct,Amount])
\end{verbatim}
\item[Assumption declarations]
Assumption declarations are used to control the abductive proof
mechanism: it is only permissible to make an abductive assumption when
this is explicitly licenced by an assumption declaration.
An assumption declaration has the general form
\begin{verbatim}
assumable(<Goal>,<Cost>,<Justification>,<Type>,<Conds>)
\end{verbatim}
The intended semantics are that \verb!<Goal>! may be assumed at
cost \verb!<Cost>! in an environment where
\verb!<Conds>! hold. \verb!<Type>! can be either {\tt specialization},
{\tt limitation} or {\tt approximation}; this is explained further in
section~\ref{Basic-LDT}. \verb!<Justification>! is a tag which can be
used to identify instances of use of this declaration.  Assumption
declarations are mainly used in the PRM LDT to deal with the problem
of ``contingent completeness'' of information. Examples can be found
in Sections~\ref{Database-specific-axioms}
and~\ref{Assumption-declarations}.
\end{description}
The rest of the chapter will explain in detail how these various kinds
of information can be used to build up an LDT; we will focus on the
specific PRM domain for lack of other large-scale examples, but we
believe that many of the principles would carry over to other
relational databases.

\section{Overview of the example LDT}
\label{DBLDT}

This section gives an overview of how the LDT for the PRM domain is
structured. The PRM relational database is based on a real SRI
projects and payments database; the LDT contains about 300
equivalences, 70 Horn clauses, 45 declarations of functional
relationships, and 25 assumption declarations.

It will be easiest to start by describing the various types of
predicates used in the PRM domain's LDT. At one end are the
{\it linguistic} predicates, mostly corresponding to word-senses;
the names of these will be written in the form
\verb!<LowerCaseSurfaceForm><Number>!
e.g. {\tt cheque1}, {\tt project1}, {\tt make2}, {\tt on1}, etc.  At
the other end, we have the predicates meaningful to the database.  As
explained in Section~\ref{Effective-translation}, these are of three
kinds.

Firstly, we have the {\it database predicates} themselves, which
directly correspond to the database relations. We describe these in
more detail immediately below. Secondly, we have {\it arithmetical
relation predicates}, which define the permissible arithmetical
operations that can be carried out on database objects. The LDT
contains predicates corresponding to addition and subtraction,
inequality between numbers, and temporal precedence between
representations of dates. Thirdly, we have the single {\it executable
predicate}, instances of which the system can cause to hold in the
world. The executable predicate is
\begin{verbatim}
execute_in_future(Action)
\end{verbatim}
which holds if {\tt Tuple} is a term representing an action which the
system will perform at some future time. The type of action we will be
most interested is the displaying of a list of database objects, which
we write {\tt display(List)}.

We now return to consider the database relations, the names of which
will be written entirely in upper case. We will refer to the following
specific predicates:
\begin{verbatim}
TRANS(TransId,ChequeId,Date,PayeeId,AccNum,Amt)
\end{verbatim}
Non-payroll payments made by SRI Cambridge between 17/8/89 and 1/4/91.
For each record, the transaction ID is {\tt TransId} and cheque ID is
{\tt ChequeId}: if {\tt Amt} is positive, SRI paid the payee with name
{\tt PayeeId} the quantity {\tt Amt} in pounds sterling on day {\tt
Date}, charging it to account {\tt AccNum}. Records with negative {\tt
Amt} turned out to represent an abuse of the database system to record
reimbursements of small debts to petty cash; this is characteristic of
the sort of arbitrary condition that can easily apply in a real
database.
\begin{verbatim}
PROJECT(Name,ProjectNum,Start,End)
\end{verbatim}
Projects at SRI Cambridge. The project's name is {\tt Name} and its
account number is {\tt ProjectNum}; the start date is {\tt Start} and
the end date is {\tt End}.
\begin{verbatim}
PAYEE(PayeeId,Type)
\end{verbatim}
Payees in listed transactions, classified by the field {\tt Type} as
being men ({\tt m}), women ({\tt w}), companies ({\tt c}),
universities ({\tt u}) or others ({\tt o}).
\begin{verbatim}
EMPLOYEE(Name,Sex,HasCar)
\end{verbatim}
Employees at SRI Cambridge, classified by their sex ({\tt m} or {\tt
w}) and whether or not they have a company car ({\tt y} or {\tt}).
\begin{verbatim}
TIMESHEET(EmployeeId,Time,Date,Project)
\end{verbatim}
Timesheet entries for the periods that had been entered when the
database was compiled (these periods turned out to be
project-dependent).  The fields give in order the employee ID for the
person booking the time, the number of hours booked, the date for
which time was booked, and the account number charged to.

To sum up, we have on one hand the linguistic predicates, which are
directly related to language, and on the other the database,
arithmetic relation and executable predicates, which are directly
related to the database. Between these, it has proved convenient to
define a number of other ``intermediate'' predicates.  There are in
particular two specific types of intermediate predicates, which we
will shortly describe, that play a major rule in the LDT; we call
these {\it conceptual} predicates and {\it attribute} predicates.
Translation, in the direction language to database, proceeds very roughly
as follows. Linguistic predicates are translated into a mixture of
conceptual and attribute predicates. The attribute predicates are
then translated into conceptual predicates; finally, the conceptual
predicates are tranlated into database predicates.

The rest of the chapter will make this thumbnail sketch more explicit.
In Section~\ref{Names-and-objects}, we begin by defining the types
of terms used in the theory: in particular, we consider the
relationship between terms representing names and terms representing
the objects referred to by the names.
Sections~\ref{Conceptual-predicates} and~\ref{Attribute-predicates}
describe conceptual and attribute predicates. We will then have
presented enough definitions to be able to describe the structure of
the LDT in detail.  The LDT's axioms are stratified into three
distinct groups, listed in order of increasing specificity: this
reflects the basic strategy of translating the original logical form
predicates into predicates which are increasingly closer to the
database.  We refer to the groups as {\it general}, {\it
domain-specific} and {\it database-specific}, and describe them in
turn in Sections~\ref{Basicdomain},~\ref{Domain-specific-axioms}
and~\ref{Database-specific-axioms}. The next three sections describe
particular types of information in the LDT.
Section~\ref{Specific-linguistic-constructs} describes the axioms used
to deal with some of the more interesting linguistic constructions;
Section~\ref{Assumption-declarations} discusses assumption
declarations; and Section~\ref{Horn-clause-axioms} briefly examines
Horn-clause axioms. In the final section, we present an extended
example showing translation of a complex query.

\subsection{Names and objects}\label{Names-and-objects}

For reasons explained in Section~\ref{Ontology}, we regard the terms
appearing as arguments to database predicates as referring not to
entities in the exterior world, but rather to ``database objects'',
which in practice can be taken to mean strings of characters. Since
the linguistic predicates are intended to range over real-world
objects, it is necessary to provide some way of relating objects to
their database identifiers. This is done through the {\it naming
predicates}.  The intuition is that database identifiers are names, but
not unique names: they need to be associated with some extra information
to become unique. For example, there are some numbers which are both
cheque and transaction IDs; the fact that a cheque ID happens to be
identical with a transaction ID does {\it not} imply that the cheque
and transaction are identical, or even that they are related.

The LDT deals with these problems by assuming that the named objects
referred to by the theory can be divided up into a set of discrete
subsets called {\it types}, such that names are unique within each
type; it is also convenient to introduce identifiers for the types, and
to allow terms of the form
\begin{quote}
\begin{verbatim}
<Type>#<Name>
\end{verbatim}
\end{quote}
to be regarded as referring to ``the object of type
\verb!<Type>!, with identifier \verb!<Name>!''. If the types can
be identified with the extensions of English NBAR senses, then the
typed objects can be described by noun-phrases consisting of the NBAR
followed by the name.  Thus for example the typed object
\verb!transaction1#123! can be referred to by the English noun-phrase
{\it transaction 123}.  In practice, all the types used in the PRM LDT
are identified with NBAR senses in the way just described.

Three kinds of database objects that need to be treated specially are
those explicitly referring to numbers, amounts and dates. To deal with
these, there are three predicates that relate these database objects
to terms in the theory that represent the objects they name; for
uniformity there is also a fourth predicate, which encodes the
relationship between a database ID, a type, and the resulting typed
object. The fourth predicates are
\begin{description}
\item[{\tt sql\_number\_convert(Num,DBNum)}] {\tt Num} is the number
of which {\tt DBNum} is the name. In the current implementation, {\tt
Num} is a Prolog number, and {\tt DBNum} is a Prolog atom whose
print-name is the same as that of {\tt Num}. Surprisingly,
distinguishing between these two types of object actually turns out to
be very helpful in terms of simplifying the mechanics of the low-level
SQL interface.

\item[{\tt amount\_object(Amount,Unit,Num)}] {\tt Amount\-Object}
is an object representing the quantity of size {\tt Num}, measured
in the unit {\tt Unit}.

\item[{\tt sql\_date\_convert(Date,DBDate)}] {\tt DBDate} is
the database representation of the date {\tt Date}. In the current
implementation, {\tt Date} is a structured term whose components
represent the day, month and year of the date; {\tt DBDate} is a
Prolog atom whose print-name encodes the SQL representation of
the date.

\item[{\tt named\_object(TypedObject,Type,Id)}] {\tt Typed\-Object}
is the unique typed object (cf Section~\ref{Names-and-objects}) with
type {\tt Type} and database ID {\tt Id}.
\end{description}

\subsection{``Conceptual'' predicates}\label{Conceptual-predicates}

Because of the very conservative semantics we have assigned to the
database predicates, they are difficult to work with directly. It is
consequently useful to pair each database predicate $P_{db}$ with an
``idealized'' version which we call the corresponding {\it conceptual}
predicate, $P_{conc}$.  The exact general relationship between the
``database'' and ``conceptual'' versions of predicate is not meant to
be strictly defined, but the following guiding priniples have been
found to be useful:
\begin{itemize}
\item Each argument $Arg_{db}$ in $P_{db}$ should have a corresponding
argument $Arg_{conc}$ in $P_{conc}$. If $Arg_{db}$ is an identifier,
then $Arg_{conc}$ is the object it identifies; if it is a ``code''
(e.g. a yes/no value), then the values of the two arguments should be
equal.
\item An extra argument should be added to $P_{conc}$ if $P_{db}$
has no primary key field. This argument will often end up
corresponding to the ``entity'' argument in verbs that translate
into $P_{conc}$.
\item In some cases, other extra arguments may need to be added to
$P_{conc}$ corresponding to objects that can be referred to in natural
language in connection with $P_{db}$, but which are always assumed to
have the same value. For example, all the transactions in the database
have been made by SRI, and this is implicit in the semantics of the
{\tt TRANS} relation. Since SRI can however be {\it explicitly}
referred to in English as making payments, it is helpful to have an
argument in the conceptual predicate corresponding to {\tt TRANS}
which holds the ``payer'' in the transaction.
\item ``Contingent'' restrictions relating to $P_{db}$ should not
apply to $P_{conc}$, for example that the record only holds of
events occurring within a specified period. The contingent
completeness information is added instead as part of the formulas
relating $P_{db}$ and $P_{conc}$.
\end{itemize}
The brief discussion above should motivate the following definitions
of the conceptual predicates associated with the database predicates
in the PRM LDT. We will describe the axioms that relate each database
predicate to its associated conceptual predicate later, in
Section~\ref{Database-specific-axioms}. For the moment, we leave the
connection between them informal.
\begin{verbatim}
transaction(Trans,Payer,Cheque,Date,Payee,Account,Amt)
\end{verbatim}
{\tt Trans} is a payment made by {\tt Payer} to {\tt Payee} on the day
{\tt Date}. The amount was {\tt Amt} and the payment was accomplished
using {\tt Cheque}; it was charged to {\tt Account}. Note that the new
argument {\tt Payer} has been added, and the restrictions concerning
the permitted range of dates dropped. Most importantly, the arguments
are all objects in the exterior world rather than strings.
\begin{verbatim}
project(Project,Organisation,Account,Start,End)
\end{verbatim}
{\tt Project} is a project at {\tt Organisation}. It has the
associated account {\tt Account}, the start date is {\tt Start}, and
the end date is {\tt End}. Once again, a new argument ({\tt
Organisation}) has been added.
\begin{verbatim}
payee(Payee,Type)
\end{verbatim}
{\tt Payee} is a payee of the type defined by {\tt Type}. Here, the
difference between the conceptual and database predicates is
minimal.
\begin{verbatim}
employee(Employee,Organisation,Sex,HasCar),
\end{verbatim}
Employees at {\tt Organisation}, classified by their sex and whether
or not they have a company car.
\begin{verbatim}
booking_to_project(Event,Person,Time,Date,Project)
\end{verbatim}
Since the {\tt TIMESHEET} predicate has no primary key, the argument
{\tt Event} is added, corresponding to the event of {\tt Person}'s
booking the quantity of time {\tt Time} to {\tt Project} on {\tt
Date}.

Database relationships are usually functions on at least one of
their arguments (those that correspond to primary keys); as just
discussed, conceptual relations always have this property, if
necessary by construction. The translation mechanism is able to make
use of information about functional relationships in several ways
(cf~\ref{Functional-relations}). Functional relations are defined by
declarations of the form
\begin{verbatim}
function(<Template>,<From> -> <To>)
\end{verbatim}
This is to be read as stating that in
\verb!<Template>!, an assignation of values to the variables making up
the list \verb!<From>! uniquely determines the values of the variables in
the list \verb!<To>!.

Function declarations of this type are consequently supplied both for
database and conceptual predicates. For example, the declarations for
{\tt TRANS} and {\tt transaction} are
\begin{verbatim}
function(TRANS(TransId,ChequeId,Date,Payee,Acc,Amt),
         ([TransId] -> [ChequeId,Date,Payee,Acc,Amt]))
function(TRANS(TransId,ChequeId,Date,Payee,Acc,Amt),
         ([ChequeId] -> [TransId,Date,Payee,Acc,Amt]))

function(transaction(Trans,Payer,Cheque,Date,Payee,Acc,Amt),
         ([Trans] -> [Payer,Cheque,Date,Payee,Acc,Amt]))
function(transaction(Trans,Payer,Cheque,Date,Payee,Acc,Amt),
         ([Cheque] -> [Payer,Trans,Date,Payee,Acc,Amt]))
\end{verbatim}
Either the transaction ID or the cheque number could be used as
a primary key, and hence there are two functional declarations
for each predicate.

\subsection{``Attribute'' predicates}\label{Attribute-predicates}

There is a second general class of intermediate predicates that we
have found it helpful to introduce; we call these ``attribute''
predicates. Attribute predicate receive their name because they are
used to associate an object or event with a characteristic attribute,
such as time of occurrence or magnitude. There is one attribute
predicate for each such attribute. Introducing the attribute
predicates makes the LDT more modular. For example, most temporal
prepositions translate directly or indirectly into expressions
involving the time associated with an object; however, the target
predicates used to express the relationship between an object and its
associated time depend on the nature of the object.  By introducing
the attribute predicate {\tt associated\_\-time}, it is possible to
produce a modular solution. The predicates representing the
preposition senses are first translated, using domain-independent
equivalences, into a representation expressed in terms of the {\tt
associated\_\-time} predicate; then domain-dependent equivalences (one
for each relevant type of object) translate {\tt associated\_\-time}
into conceptual predicates.

In the PRM LDT, we use four attribute predicates.  There are three
temporal attribute predicates, {\tt associated\_\-time}, {\tt
associated\_\-start\_\-time} and {\tt associated\_\-end\_\-time},
which are produced as intermediate translations of word-sense
predicates for words like {\it when}, {\it during}, {\it after}, {\it
start} and {\it end}.  Each temporal attribute predicate is a
three-place predicate, whose arguments are, in order
\begin{itemize}
\item the object (which will often be an event)
\item the associated date or time
\item the ``granularity'' of the date or time, which at present must be either
{\tt days} or {\tt seconds}
\end{itemize}
The fourth attribute predicate {\tt associated\_\-size}, has two
arguments. It is produced as a result of translating linguistic
predicates denoting the ``size'' or ``magnitude'' of an objects, which
mainly arise in constructions that compare objects with reference to
their extent. Expressions of this form include comparatives
({\it larger}, {\it smaller}) superlatives ({\it largest}, {\it smallest}),
and some prepositions like {\it over} ({\it over \pounds 250}).

\section{General axioms}\label{Basicdomain}

In the next three sections, we describe the axioms of the PRM LDT: as
indicated earlier, these are divided into three groups, which we refer
to as ``general'', ``domain-specific'' and ``database-specific''
respectively.  About 195 of the axioms fall into the first,
``general'' group.  About 45 of them translate words used to refer to
concepts in the utterance situation ({\it show}, {\it question}, {\it
answer} etc.) into representations in terms of the primitive
executable predicates. For the sort of modularity reasons already
referred to, translation is in fact carried out in three stages. The
linguistic word-sense predicates are first translated into an
intermediate form, expressed in terms of the predicates {\tt execute}
and {\tt display\_\-format}.  For example a typical equivalence
of this type would be the following one, that translates the
ditransitive verb {\it show} ({\it Show me the payments}):
\begin{verbatim}
show2(Event,Agent,X,Audience) <->
exists([Time,Format],
  and(display_format(X,Format)
      execute(Event,display(Format),Agent,Audience,Time))
\end{verbatim}
The right-hand side of the above equivalence expresses the fact that a
``showing'' of an object {\tt X} by {\tt Agent} to {\tt Audience}
involves finding a suitable display format {\tt Format}, and then
displaying {\tt Format}. In practice, {\tt Agent} will be {\tt clare}
and {\tt Audience} a term referring to the user.  In the second round
of translation, the predicate {\tt display\_format} is replaced by a
representation in terms of ``conceptual'' predicates, in a way that is
dependent on the nature of {\tt X}: this is discussed later, in
Section~\ref{Equiv-identifiers}. Finally, goals of the form
\begin{verbatim}
execute(Event,Action,clare,<User>,Time)
\end{verbatim}
are translated into one of the two forms
\begin{verbatim}
execute_in_future(Action)
\end{verbatim}
or
\begin{verbatim}
executed_in_past(Action)
\end{verbatim}
depending on whether the context makes {\tt Time} after or before the
present moment. This division mirrors the important pragmatic
distinction between future actions, which can be caused to occur if
the system so desires, and past actions, which can be ``remembered''
by consulting the action log.

Of the remaining ``general'' axioms, about 100 translate basic word
senses, like temporal prepositions and size expressions, into the
``attribute predicates'' described in
Section~\ref{Attribute-predicates} above.  A simple example is
provided by the predicate corresponding to the the temporal
preposition {\it during}, which is translated by the equivalence
\begin{verbatim}
during1(E1,E2) <->
exists([T1,T2,Gran1,Gran2],
  and(associated_time(E1,T1,Gran1),
  and(associated_time(E2,T2,Gran2),
      time_during([T1,Gran1],[T2,Gran2]))))
\end{verbatim}
This expresses the fact that an event {\tt E1} can be regarded as
being ``during'' another event {\tt E2} if the time associated with
the first event is contained in the time associated with the second
one. The arguments {\tt Gran1} and {\tt Gran2} express the conceptual
level of granularity of the events {\tt E1} and {\tt E2}; at present
they can be either {\tt days} (appropriate for database events) or
{\tt seconds} (appropriate for utterance situation events).  Note that
the definition of the {\tt associated\_time} predicate is provided by
domain-specific axioms; the equivalence performs the
domain-independent function of translating the linguistic predicate
{\tt during1} into a more convenient form expressed in terms of the
general predicates {\tt associated\_\-time} and {\tt time\_\-during}.

The inherently vague nature of many prepositions, however, means that
an extra level of indirection is often required to translate them. The
predicate corresponding to the preposition is first translated by a
conditional equivalence into a second predicate representing a
specialized preposition-sense; the conditions reflect suitable
constraints on the preposition's arguments. The preposition sense
predicate can then be translated into ``attribute predicates'' by
equivalences like the one for {\tt during1} shown above. For instance,
the following conditional equivalence is in the PRM LDT sufficient to
deal with the temporal sense of the preposition {\it from}, as in {\it
from 1/3/90 to 15/5/90}:
\begin{verbatim}
(from1(X,Y) <-> from_Temporal(X,Y)) <-
  temporal_entity(Y).
\end{verbatim}
The predicate {\tt temporal\_entity} is defined by Horn-clauses to
hold of dates, times and a few other similar objects.

The final set of about 30 general rules deal with translations of
predicates like {\tt time\_\-during}, which relate times; the
complications are caused by the fact that ``times'' can be either
points or intervals, and can also be expressed at different levels of
granularity.  The temporal axioms translate all the temporal
predicates into a uniform representation in terms of a single
predicate, {\tt time\_\-point\_\-precedes}, that compares time-points
with regard to temporal precedence.

\section{Domain-specific axioms}\label{Domain-specific-axioms}

The axioms of the second group, which for the PRM domain are somewhat
less numerous than the first, are domain-specific but not
database-specific.  They translate predicates representing senses of
domain words ({\it project}, {\it transaction}, {\it deliverable}
etc.)  and ``attribute'' predicates like {\tt asso\-ci\-ated\_\-time},
into representations expressed in terms of the conceptual predicates
of Section~\ref{Conceptual-predicates}.

Translation of both domain word-sense predicates and attribute
predicates is frequently dependent on the nature of their arguments, and
the corresponding equivalences are thus either conditional or have
complex left-hand sides.  For example, the time associated with a
payment is the transaction date, but that associated with a project is
the interval spanned by the start- and end-dates. The equivalences
that express this are
\begin{verbatim}
and(associated_time(Trans,Date,Granularity),
    transaction1(Trans)) <->
exists([Payer,Cheque,Payee,Acc,Amt],
   and(transaction(Payer,Trans,Cheque,Date,Payee,Acc,Amt),
       Granularity = days))

and(associated_time(Project,Interval,Gran),
    project1(Project)) <->
exists([StartDate,EndDate],
    and(project(Org,Project,Account,StartDate,EndDate),
        Interval = interval(StartDate,EndDate))
\end{verbatim}
Other examples of domain-specific equivalences can be found at the
beginning of Section~\ref{Specific-linguistic-constructs}.

\section{Database-specific axioms}\label{Database-specific-axioms}

The third group of axioms relate the conceptual predicates to the
associated concrete database predicates.  They capture specific coding
peculiarities which relate database-objects and their associated
real-world counterparts, which, as explained above in
section~\ref{Closed-World}, includes contingent limitations on the
availability of recorded data. The database-specific group is small,
consisting of about 25 axioms; these consist of one or more equivalence
linking each conceptual predicate to its concrete counterpart,
together with some axioms that define the ``contingent limitations''
when necessary. For example, the equivalences that relate the database
predicate {\tt TIMESHEET} and its conceptual counterpart {\tt
book\_\-time\_\-to\_\-project} (slightly simplified) are
\begin{verbatim}
(book_time_to_project(Ev,Emp,Amt,Date,Acc) <->
 book_time_to_project_recorded(Ev,Emp,Amt,Date,Acc)) <-
   timesheet_data_available(Date,Acc)

exists([Ev],
  book_time_to_project_recorded(Ev,Emp,Amt,Date,Acc) <->
exists([EmpId,DBDate,AccId],
  and(TIMESHEET(EmpId,Amt,DBDate,AccId),
      and(named_object(Emp,employee1,EmpId),
          and(sql_date_convert(Date,DBDate),
              named_object(Acc,account1,AccId)))))
\end{verbatim}
The first of these is a ``universal'' equivalence, that encodes the
contingent completeness restrictions. It consults the predicate {\tt
timesheet\_\-data\_\-available} to find out whether {\tt Date} is
within the range for which timesheet data on account {\tt Acc} has
been recorded. {\tt timesheet\_\-data\_\-available} is defined by a
set of Horn-clauses (one for each account), of the form
\begin{verbatim}
timesheet_data_available(D,<CLAREAccount>) <-
  and(time_point_precedes(nonstrict,<CLAREStartDate>,D),
      time_point_precedes(nonstrict,D,<CLARELastRecordedDate>))

timesheet_data_available(D,<SLTAccount>) <-
  and(time_point_precedes(nonstrict,<SLTStartDate>,D),
      time_point_precedes(nonstrict,D,<SLTLastRecordedDate>))
...
\end{verbatim}
where \verb!<CLAREAccount>! is a ground term representing the CLARE
account, and so on. The second, ``existential'' equivalence above uses
the ``naming'' predicates {\tt named\_object} and {\tt
sql\_date\_convert} to relate the real-world objects referred to by
{\tt book\_\-time\_\-to\_\-project} and the database identifiers
in {\tt TIMESHEET}.

Most of the permissible assumptions that can be made are related to
the database-specific axioms, and are of one of two types.  The first
type, a ``specialization'' (see section~\ref{Basic-LDT}), permits the
assumption that arguments like ``employer'' in the {\tt project}
conceptual relation or ``payer'' in the {\tt transaction} relation are
filled by {\tt SRI} unless explicitly filled by something else. There
is a predicate, {\tt assumably\_=}, to deal with this situation:
we illustrate with the {\tt project} relation. The equivalence that
translates it into the associated database predicate is schematically
of the form
\begin{verbatim}
(project(Proj,Org,Acc,StartDate,EndDate) <->
 ... <Right-hand side> ...) <-
  assumably_=(Org,
              organization1#sri,
              all_projects_referred_to_are_at_SRI)
\end{verbatim}
The content of the condition on this equivalence is that {\tt Org}
should be equal to \verb!organization1#sri!. If it is not, it may be
assumed to be, with the ``justification'' {\tt
all\_\-projects\_\-referred\_\-to\_\-are\_\-at\_\-SRI}.  The intended
semantics of the predicate {\tt assumably\_=}, as the name suggests,
are that it should hold if the first two arguments are equal;
otherwise, they can be assumed equal, with the third argument serving
as a ``justification''. This is specified by the following Horn-clause
and assumption declaration:
\begin{verbatim}
assumably_=(X,Y,Justification) <- X=Y

assumable(assumably_=(X,Y,Justification),
          0,
          Justification,
          specialization,
          true)
\end{verbatim}

The second main type of assumption, a ``limitation'', is related to
temporal incompleteness of database records: here, the declaration
is to the effect that it is permissible to assume that data is
within the period for which records have been kept unless the
conjunctive context implies the contrary. We illustrate with
the {\tt timesheet\_\-data\_\-available} predicate above. There
is one assumption declaration for each account. The declaration for
the CLARE account is for example
\begin{verbatim}
assumable(
  timesheet_data_available(D,<CLAREAcc>),
  15,
  time_was_booked_after_<CLAREStartDate>_and_\\
                  before_<CLARELastRecordedDate>,
  limitation,
  true))
\end{verbatim}

\section{Specific linguistic constructs}
\label{Specific-linguistic-constructs}

In this section, we will describe the types of equivalences used to
translate some specific types of linguistic constructs. We begin with
the unproblematic cases, in which a linguistic predicate can directly
be mapped into a relation on one more slots of a conceptual or
attribute predicate. (We will use the term ``conceptual slot'' as a
handy abbreviation for ``argument of a conceptual predicate''). The
simplest case is perhaps that of a common noun which describes a
conceptual slot.  In the following example, the rule says that an
object which fills the third argument place in the conceptual {\tt
transaction} relation can be described by the predicate {\tt cheque1}.
\begin{verbatim}
cheque1(Cheque) <->
exists([Payer,Trans,Date,Payee,Acc,Amt],
  transaction(Trans,Payer,Cheque,Date,Payee,Acc,Amt))
\end{verbatim}
Slightly more complex rules are used to translate many domain-specific
preposition senses. The following are the equivalences that cover
``cheque for amount'' ({\it cheques for more than \pounds 200}),
``cheque to payee'' ({\it cheque to Maxwells}) and ``cheque on
account'' ({\it cheques on account 8468}):
\begin{verbatim}
and(for(Cheque,Amt),
    cheque1(Cheque)) <->
exists([Trans,Payer,Date,Payee,Acc,Amt],
  transaction(Trans,Payer,Cheque,Date,Payee,Acc,Amt))

and(to(Cheque,Payee),
    cheque1(Cheque)) <->
exists([Trans,Payer,Date,Payee,Acc,Amt],
  transaction(Trans,Payer,Cheque,Date,Payee,Acc,Amt))

and(on(Cheque,Acc),
    and(cheque1(Cheque),
        account1(Acc))) <->
exists([Trans,Payer,Date,Payee,Acc,Amt],
  transaction(Trans,Payer,Cheque,Date,Payee,Acc,Amt))
\end{verbatim}
In all these examples, the conditions on applicability are provided by
extra conjuncts on the LHS. Similar considerations apply to many verb
senses: the following, for example, is the rule that covers ``pay
money to'', as in {\it SRI paid \pounds 120 to Maxwells}.
\begin{verbatim}
and(pay2(Trans,Payer,Amt,Payee),
    money1(Amt)) <->
exists([Payer,Date,Acc],
  transaction(Trans,Payer,Cheque,Date,Payee,Acc,Amt))
\end{verbatim}
Note that the ``paying event'' is regarded as being the same as the
transaction, and that the agent of the paying event becomes the {\tt
Payer} in the conceptual {\tt transaction} relation.

In the remainder of this section, we consider constructions that
require equivalences of a different or more complex form.

\subsection{Translating ``attribute'' predicates}

``Attribute'' predicates can be translated in much the same way as the
linguistic predicates we have seen above.  Domain-dependent
equivalences are provided to translate the attributes predicates in
the contexts provided by the domain word-senses: the right-hand sides
of these equivalences will usually contain conceptual predicates.  The
following rule, for example, translates the {\tt associated\_\-time}
predicate in a context where its first argument is a transaction; it
states that the fourth field of a {\tt transaction} relation is the
time (at the level of granularity of days) associated with the
transaction.
\begin{verbatim}
and(associated_time(Trans,Date,Granularity),
    transaction1(Trans)) <->
exists([Cheque,Payer,Payee,Acc,Amt],
  and(transaction(Trans,Payer,Cheque,Date,Payee,Acc,Amt),
      Granularity = days))
\end{verbatim}
Similarly, the following pair of rules state that the start and end
dates of a project are encoded in the fourth and fifth fields of the
{\tt project} relationship respectively.
\begin{verbatim}
and(associated_start_time(Project,StartDate,Granularity),
    project1(Project)) <->
exists([Org,ProjNum,EndDate],
  and(project(Project,Org,ProjNum,StartDate,EndDate),
      Granularity = days))

and(associated_end_time(Project,EndDate,Granularity),
    project1(Project)) <->
exists([Org,ProjNum,StartDate],
  and(project(Project,Org,ProjNum,StartDate,EndDate),
      Granularity = days))
\end{verbatim}
The {\tt associated\_\-size} primitive predicate takes two arguments:
the first is the object in question, and the second is its ``size'',
which is represented as number. Finding a suitable number usually
involves including a goal in the RHS which extracts the number from a
term representing an amount. In the following example, which defines
the ``size'' of a cheque, the extraction is performed using the {\tt
amount\_object} predicate (cf Section~\ref{Names-and-objects}).
\begin{verbatim}
and(associated_size(Cheque,N),
    cheque1(Cheque)) <->
exists([Trans,Payer,Date,Payee,Acc,Amt],
  and(transaction(Trans,Payer,Cheque,Date,Payee,Acc,Amt),
      amount_object(AM,sterling_unit,N)))
\end{verbatim}

\subsection{Support verbs}

Support verb constructions like ``make a payment'' also demand a
slightly different treatment.  The next example shows the relevant
equivalence:
\begin{verbatim}
and(make2(MakeEvent,Payer,Trans),
    payment1(Trans)) <->
exists([Cheque,Date,Payee,Acc,Amt],
  and(transaction(Trans,Payer,Cheque,Date,Payee,Acc,Amt)
      MakeEvent = Trans))
\end{verbatim}
Here the ``making'' event is once again regarded as being the same as
the payment. In cases like this, where two variables from the LHS
({\tt MakeEvent} and {\tt Trans}) are to be mapped onto a single event
on the RHS, the most straightforward solution is to write the rule in
the way shown, with an equality on the right. It is worth noting
that the apparently plausible formula
\begin{verbatim}
and(make2(Trans,Payer,Trans),
    payment1(Trans)) <->
exists([Cheque,Date,Payee,Acc,Amt],
  transaction(Trans,Payer,Cheque,Date,Payee,Acc,Amt))
\end{verbatim}
does not achieve the desired end, and is not in fact equivalent with
the previous one. The easiest way to convince oneself of the truth of
this statement (which may seem counter-intuitive to readers used to
Horn-clauses) is to consider the pair of formulas
\begin{verbatim}
p(X,Y) <-> and(q(X),X = Y)
\end{verbatim}
and
\begin{verbatim}
p(X,X) <-> q(X)
\end{verbatim}
The first of them means that {\tt p(a,b)} implies {\tt q(a)}, but
not the second.

\subsection{Predicates referring to identifiers}\label{Equiv-identifiers}

Some linguistic predicates can refer to names or numbers, which in the
PRM LDT are equated with database identifiers.  For example, the
equivalence used to translate the relational sense of the compound
noun {\it project number} (e.g. {\it CLARE's project number}) is
\begin{verbatim}
project_number_of(N,Project),
exists([Org,Start,End,Account],
  and(project(Project,Org,Account,Start,End),
      named_object(Account,account1,N)))
\end{verbatim}
Note the use of the predicate {\tt named\_\-object} on the RHS. This
expresses the fact that {\tt Account} is the ``typed object'' (cf
Section~\ref{Names-and-objects}) of type {\tt account1} with identifier
{\tt N}, i.e.  the object referred to by the English expression
{\it account {\tt N}}.

Equivalences of this type are also used to help translate
predicates corresponding to ``displaying'' verbs like {\it show} and
{\it list}. As explained in Section~\ref{Basicdomain} above, these are
first translated into a uniform representation that uses the
predicates {\tt executable\_\-action/6} and {\tt display\_\-format/2}.
The second of these is a predicate which determines the appearance of
the term printed to represent a database object: the intention is that
\begin{quote}\begin{verbatim}
display_format(Object,Format)
\end{verbatim}\end{quote}
should hold if {\tt Format} is
either a database object, or a list containing one or more database
objects, that can be used together as a representation of {\tt Object}
suitable for displaying to the user. For example, a transaction can be
conveniently displayed as a list consisting of its transaction ID, the
date of the payment, the payee ID, and the amount. This is captured in
the following equivalence,
\begin{verbatim}
and(display_format(Trans,Format),
    transaction1(Trans)) <->
exists([Payer,C,Date,Payee,AC,Amt,TransId,DBDate,PayeeId,AmtNum],
  and(transaction(Trans,Payer,C,Date,Payee,AC,Amt),
  and(named_object(Trans,transaction1,TransId),
  and(sql_date_convert(Date,DBDate),
  and(named_object(Payee,payee1,PayeeId),
  and(amount_object(Amt,pounds_sterling,AmtNum)
      Format = [TransId,DBDate,PayeeId,AmtNum]))))))
\end{verbatim}
There is one {\tt display\_\-format} definition for each type of
object. Many objects are ``typed objects'' whose names serve as
adequate descriptions; this is for example the case for sponsors in
the PRM domain. In cases like these, an equivalence like the following
one for ``sponsors'' is sufficient:
\begin{verbatim}
(display_format(Sponsor,Format) <->
 named_object(Sponsor,sponsor1,Format)) <-
   sponsor1(Sponsor).
\end{verbatim}

\subsection{Expressions that map onto code values}\label{Equiv-codes}

The filler of a field in a database relationship is often a code, and
an English expression can map onto a value, or set of values, for this
code; thus for example it might easily be the case that a word like
{\it man} or {\it woman} ultimately maps onto a field being filled by an
{\tt m} or a {\tt w}. The problems involved have already been
discussed at some length in Section~\ref{Doctor-on-board}.  Here, we
present some more examples from the PRM LDT.

The first and second equivalences below are straightforward. The first
states that {\it man} when applied to an object known to be a payee,
maps onto the first field of the {\tt payee} relation being filled by
an {\tt m} (recall that {\tt payee1} is a predicate corresponding to a
wods-sense, while {\tt payee} is a ``conceptual'' predicate).
The second states that an overhead payment is a transaction
where the ``account'' field is filled by a numbered object of type
``account'' whose number satisfies the predicate {\tt
overhead\_\-code}.
\begin{verbatim}
and(man1(Payee),
    payee1(Payee)) <->
payee(m,Payee)

overhead_payment1(Trans) <->
exists([Payer,Cheque,Date,Payee,Acc,Amt,Num],
  and(transaction(Trans,Payer,Cheque,Date,Payee,Acc,Amt),
  and(numbered_object(Account,account1,Num),
      overhead_code(Num))))
\end{verbatim}
For the second equivalences, a further definition is needed to specify
the range of permitted values for overhead codes: they are
defined to be exactly those numbers which are members of the set {\tt
\{775, 675\}}. The equivalence used is
\begin{verbatim}
overhead_code(Num) <-> member(Num,[775,675])
\end{verbatim}
The third example uses the {\tt employee} relation, whose third
argument is set to {\tt y} if the employee in question has a company
car.  It illustrates the problems discussed earlier in
Section~\ref{Doctor-on-board}, and involves a code which indicates
existence or non-existence of an object.  The following two rules
provide an appropriate translation for the relevant predicates
{\tt company\_car1} and {\tt have1}:
\begin{verbatim}
and(company_car1(Car),
    and(employee1(Empl),
        have1(Event,Empl,Car))) <->
employee_has_car(Event,Empl,Car)

exists([Event,Car],
  employee_has_car(Event,Empl,Car)) <->
exists([Sex]
  employee(Empl,Sex,y))
\end{verbatim}

\subsection{Transferring properties between objects}
\label{Equiv-transfer-properties}

There are several cases in the PRM LDT where two related objects $X$
and $X^\prime$ can both have the same property $P$. We have already
seen one example of this phenomenon earlier in our discussion of the
``Doctor on Board'' problem (Section~\ref{Doctor-on-board}), where
both ships and doctors can have positions, and the position of a
doctor is the same as that of his ship. In all the cases in the PRM
domain, it was possible, just as with the ships and doctors, to
express the relationship between $P$, $X$ and $X^\prime$ with a simple
conditional equivalence.

We illustrate with the preposition {\it on}, used with regard to
payments; linguistically, payments can be either on accounts ({\it
Show me all payments on account 8468}) or on projects ({\it Which
payments on CLARE were over \pounds 1000?}).  This is of course a
simple example of metonymy. The ``payment on account'' case can be
mapped into the {\tt TRANS} relation, which associates payments with
account numbers. To deal with the ``payment on project'' case we
must first find the project's account number.  This can be achieved
with the following conditional equivalence:
\begin{verbatim}
(on1(X,Project) <-> on1(X,Account)) <-
  project(Project,Org,Account,Start,End).
\end{verbatim}
which compactly captures the metonymic relationship between the two
contexts of use of {\tt on1}.

\section{Assumption declarations}\label{Assumption-declarations}

Most of the assumption declarations in the PRM LDT are of the form
described earlier in Section~\ref{Database-specific-axioms}:
they make it permissible to assume that contingently complete
information really is complete for the purposes of the question.
Assumption declarations are also used in two other contexts.

Firstly, there are a couple of cases where a word can be assumed
to be used in a specialized sense: for example, {\it car} is
normally assumed to mean {\it company car}. This is achieved by
using a conditional equivalence
\begin{verbatim}
(car1(X) <-> company_car1(X)) <-
  car_is_company_car(X)
\end{verbatim}
together with the assumption declaration
\begin{verbatim}
assumable(car_is_company_car(X),
          0,
          all_cars_referred_to_are_company_cars,
          specialization,
          true)
\end{verbatim}
Here, {\tt
all\_\-cars\_\-referred\_\-to\_\-are\_\-company\_\-cars} is the
``justification'' tag used to describe the assumption to the user.

The second type of assumption declaration is more interesting, and is
related to the ``executable predicates''. Recall that these are
predicates which the system can cause to become true, by executing
actions in the world. If they refer to future events, ground
executable predicate goals can thus be assumed to be true. This is
specified by the assumption declaration (slightly simplified)
\begin{verbatim}
assumable(execute_in_future(Action),
          0,
          perform_action(Action),
          action,
          ground_action(Action)).
\end{verbatim}
The content of the declaration is that the execution of {\tt Action}
by CLARE at some future time can be assumed to be true, if {\tt
Action} is a ground term representing an action. The assumption is of
the type {\tt action}, and the justification is the term
\begin{verbatim}
perform_action(Action)
\end{verbatim}
The justification in the case of an assumption of type {\tt action} is
of a special kind: it is in a literal sense a promissory note that the
action will be carried out by the system at some time in the future.
After a proof has been completed, CLARE collects together all these
action assumptions, and discharges them by executing all the
actions described in the justification tags.

\section{Horn clause axioms}\label{Horn-clause-axioms}

A number of Horn-clause axioms are also used in the PRM LDT.  We have
already seen some examples of Horn-clause axioms in the preceding
section, used to define conditions under which conceptual and database
predicates correspond. Two other types of Horn-clause axioms found in
the LDT are also worth discussing.  The first results from the
predicate {\tt personal/1}, which holds of entities that can be
referred to by the interrogative pronoun {\it who}.  In a normal context
like {\it Who works on CLARE?} or {\it Who booked time to project 8744
this month?}, no information is contributed by the occurrence of {\tt
personal}; it is implicit that entities who can work on projects, or
book time to projects, can be referred to as ``who''. This fact is
captured in the Horn-clauses
\begin{verbatim}
personal(Person) <-
  works_on_project(Event,Person,Project).

personal(Person) <-
  booking_to_project(Event,Person,Time,Date,Account).
\end{verbatim}
The translation process uses these clauses to deduce that the
occurrences of {\tt personal} are implied by their contexts, and can
safely be thrown away, in the way described in
Section~\ref{Inferential-simplification}.

The second type of Horn-clauses are the ``negated goal'' axioms used
to find inconsistent uses of assumptions
(cf~Section~\ref{Abductive-translation}). These are formulas of the
type
\begin{verbatim}
neg(Goal) <- Body
\end{verbatim}
They are not Horn-clauses according to the formal definition of the
term, but are treated as such by the inference engine
(cf Section~\ref{Negated-proofs}).

In the current PRM LDT, negated goal clauses are mainly used in
connection with conditions on ``limitation'' assumptions. For example,
we saw in Section~\ref{Database-specific-axioms} above that
the ``open'' predicate {\tt book\_time\_to\_project} and its
``closed'' counterpart {\tt book\_time\_to\_project\_recorded}
are linked by the equivalence
\begin{verbatim}
(book_time_to_project(Ev,Emp,Amt,Date,Acc) <->
 book_time_to_project_recorded(Ev,Emp,Amt,Date,Acc)) <-
   timesheet_data_available(Date,Acc)
\end{verbatim}
where {\tt timesheet\_data\_available} is a predicate, defined by
normal Horn-clauses, that gives sufficient conditions for timesheet
data having been recorded. A typical clause, giving sufficient
conditions for timesheet data on CLARE to have been recorded,
was
\begin{verbatim}
timesheet_data_available(D,<CLAREAccount>) <-
  and(time_point_precedes(nonstrict,<CLAREStartDate>,D),
      time_point_precedes(nonstrict,D,<CLARELastRecordedDate>))
\end{verbatim}
Now we can make these conditions necessary as well, by adding the
corresponding negated clauses
\begin{verbatim}
neg(timesheet_data_available(D,<CLAREAcc>)) <-
  time_point_precedes(strict,D,<CLAREStartDate>)

neg(timesheet_data_available(D,<CLAREAcc>)) <-
  time_point_precedes(strict,<CLARELastRecordedDate>,D)
\end{verbatim}
As can be seen, these state that timesheet data is explicitly {\it
not} recorded for dates outside the specified period.

The following negated goal clause is provided for the predicate {\tt
assumably\_=} described in Section~\ref{Database-specific-axioms}:
\begin{verbatim}
neg(assumably_=(X,Y,Justification)) <-
  different_ground_terms(X,Y)
\end{verbatim}
This makes it inconsistent to assume that distinct ground terms are
equal. The predicate {\tt different\_\-ground\_\-terms} is defined in
a way that treats specially the unique constants introduced by
translation schemas for constructions that bind variables
(cf~Sections~\ref{Universal-equivalences} and~\ref{Higher-order-AET}).  These
unique constants are not considered to be ground terms for the
purposes of the {\tt different\_\-ground\_\-terms} predicate.

\section{A full example}\label{Examples}

In this final section, we present a detailed example showing
how CLARE uses the PRM linguistic domain theory
to find an effective translation of
the TRL representation of sentence {\bf (S1)} (repeated here for convenience).
\begin{description}
\item[(S1)] Show all payments made to BT during 1990.
\end{description}
We will take a few liberties with the exact representation of the
domain axioms in the interests of increased readability.  In
particular we have suppressed the complications caused by the
treatment of granularity of time, and removed some arguments from
predicates when they were irrelevant to the example. To make the
formulas more compact, we also introduce the minor notational
abbreviation of writing {\tt E} instead of {\tt exists} to represent
existential quantification, and {\tt A} instead of {\tt forall}
to represent universal quantification.

The original TRL representation of (S1) is
\begin{verbatim}
A([Trans],
  impl(E([Agnt,Ev],
         and(payment1(Trans),
             and(make2(Ev,Agnt,Trans,payee1#bt),
                 during1(Ev,year1#1990)))),
       E([ShowEv],
         and(show1(ShowEv,clare,Trans),
             before1(<Now>,ShowEv)))))
\end{verbatim}
(We write \verb!<Now>! to stand for the TRL representation of the
present moment).  The formula can be glossed as ``For all {\tt Trans}
such that {\tt Trans} is a payment made by some {\tt Agent} during the
year 1990, it is the case that CLARE will show {\tt Trans} at some
time after the present time''.

The desired effective translation will be expressed in terms of three
evaluable predicates. The database information is accessed by the
predicate {\tt TRANS(Trans\-Id,DB\-Date,Payee\-Name,Amount)}.  We have
omitted the {\tt Cheque\-Id} and {\tt Account} arguments, since they
play no part in the derivation; we have also removed the corresponding
arguments from the associated conceptual predicate, {\tt transaction}.
Dates are manipulated by the arithmetic predicate {\tt
time\_\-point\_\-precedes} (shortened to {\tt t\_precedes}), which
holds if its arguments are two representations of calendar dates such
that the first is temporally before the second.  Execution of actions
is represented as usual by the predicate
\begin{verbatim}
execute_in_future(Action)
\end{verbatim}
(cf Section~\ref{Basicdomain}).  The actions we are interested in here
are displaying actions carried out by CLARE: these will be represented
as terms of the form {\tt display(FormatList)} where {\tt FormatList}
is a list of objects to be displayed.

The first step is to use equivalences which translate the linguistic
predicates \verb!payment1!, \verb!make2!, {\tt during1}, {\tt show1}
and {\tt before1} into conceptual predicates. Replacing bound
variables with unique constants marked with asterisks,
\begin{verbatim}
payment1(Trans*)
\end{verbatim}
is first translated to
\begin{verbatim}
E([Payer,Date,Payee,Amt],
  transaction(Trans*,Payer,Date,Payee,Amt))
\end{verbatim}
using the domain-specific equivalence
\begin{verbatim}
transaction1(Trans) <->
E([Payer,Date,Payee,Amt],
  transaction(Trans,Payer,Date,Payee,Amt))).
\end{verbatim}
Then
\begin{verbatim}
make2(Ev*,Agnt*,Trans*,payee1#bt)
\end{verbatim}
is translated into
\begin{verbatim}
E([Date1,Amt1],
  and(transaction(Ev*,Agnt*,Date1,payee1#bt,Amt1),
      Ev*=Trans*))
\end{verbatim}
using the rule
\begin{verbatim}
and(make2(Payer,Event,Payment,Payee),
    transaction1(Payment)) <->
E([Date,Amt],
  and(transaction(Event,Payer,Date,Payee,Amt),
      (Event = Payment))
\end{verbatim}
The next rule to be applied
is the one that translates \verb!during1!. The intended semantics of
{\tt during1(E1,E2)} are
``\verb!E1! and \verb!E2! are events, and the time
associated with \verb!E1! is inside that
associated with \verb!E2!''. The relevant equivalence is
\begin{verbatim}
during1(E1,E2) <->
E([Date1,Date2],
  and(associated_time(E1,Date1),
  and(associated_time(E2,Date2),
      time_during(Date1,Date2))))
\end{verbatim}
This translates
\begin{verbatim}
during1(Ev*,year1#1990)
\end{verbatim}
into
\begin{verbatim}
E([Date1,Date2],
  and(associated_time(Ev*,Date1),
      and(associated_time(year#1990,Date2),
          time_during(Date1,Date2))))
\end{verbatim}
This can now be translated further by using equivalences for {\tt
associated\_time}. For the first conjunct, the appropriate one is
\begin{verbatim}
and(associated_time(Trans,Date),
    transaction1(Trans)) <->
E([Payer,Payee,Amt],
  transaction(Trans,Payer,Date,Payee,Amt))
\end{verbatim}
which says that the {\tt associated\_\-time} of a {\tt transaction} is its
third argument; for the second conjunct, we have
\begin{verbatim}
associated_time(year1#YearNum,Time) <->
Time = interval(date(YearNum,1,1),date(YearNum,12,31))
\end{verbatim}
which says that the time associated with a year is the interval
starting on the first of January and ending on the 31st of December.
Applying these in turn, we get a representation of
\begin{verbatim}
during1(Ev*,year1#1990)
\end{verbatim}
in terms of conceptual predicates, namely
\begin{verbatim}
E([Payer2,Date2,Payee2,Amt2],
  and(transaction(Ev*,Payer2,Date2,Payee2,Amt2),
      time_during(Date2,interval(date([1990,1,1]),
                                 date([1990,12,31])))))
\end{verbatim}
Next
\begin{verbatim}
show1(ShowEv*,clare,Trans*)
\end{verbatim}
is translated by the following two equivalences. The first, a
``general'' axiom,
\begin{verbatim}
show1(E,Agent,X) <->
E([DisplayT,Format],
  and(execute(E,display(Format),Agent,DisplayT),
      display_format(X,Format)))
\end{verbatim}
can be glossed ``A showing of {\tt X} by {\tt Agent} is the execution
of a displaying of {\tt Format}, where {\tt Format} is the
display-format of {\tt X}''. The second,
\begin{verbatim}
and(display_format(Trans,Format),
    transaction1(Trans)) <->
E([TransId,Payer,Date,DBDate,Payee,PayeeId,Amt,AmtNum],
  and(transaction(Trans,Payer,Date,Payee,Amt),
  and(named_object(Trans,TransId,transaction),
  and(sql_date_convert(Date,DBDate),
  and(named_object(Payee,PayeeId,payee1),
  and(amount_object(Amt,pounds,AmtNum),
      Format = [TransId,DBDate,PayeeId,AmtNum]))))
\end{verbatim}
(see section~\ref{Equiv-identifiers}) is a context-sensitive
domain-dependent
equivalence which defines the display-format for a transaction to
be a four-element list consisting of the transaction ID, the transaction
date, the payee's name, and the amount. The result of applying both
of these is
\begin{verbatim}
E([TransId,Date3,DBDate,Payee3,PayeeId,Amt3,AmtNum,DisplayT],
  and(execute(ShowEv*,display([TransId,DBDate,PayeeId,AmtNum]),
              clare,DisplayT),
   and(transaction(Trans*,Payer3,Date3,Payee3,Amt3),
    and(Trans*=transaction1#Id,
     and(sql_date_convert(Date,DBDate)
      and(Payee3=payee1#PayeeId,
          Amt3=amount(AmtNum,pounds)))))))
\end{verbatim}
Finally,
\begin{verbatim}
before1(<Now>,ShowEv*)
\end{verbatim}
is translated into
\begin{verbatim}
E([Format1,DisplayAgent,DisplayT]
  and(execute(ShowEv*,Format1,DisplayAgent,DisplayT),
      time_before(<Now>,DisplayT)))
\end{verbatim}
using rules similar to those used above to translate {\tt during1}.
When all the rules mentioned so far
have been applied, the translated form is
\begin{verbatim}
A([Payer,Trans,Date,Payee,Amt,Payer1,Date1,Amt1,Payer2,
   Date2,Payee2,Amt2],
  impl(and(transaction(Trans,Payer,Date,Payee,Amt),
        and(transaction(Trans,Payer1,Date1,payee1#bt,Amt1)
         and(transaction(Trans,Payer2,Date2,Payee2,Amt2)
             time_during(Date2,
                         interval(date([1990,1,1]),
                                  date([1990,12,31])))))),
       E([TransId,Date3,DBDate,Payee3,PayeeId,Amt3,AmtNum,
          DisplayEv,Format1,DisplayAg,DisplayT3,DisplayT],
         and(execute(DisplayEv,
                     display([Id,DBDate,PayeeId,Amt3]),
                     clare,
                     DisplayT1),
           and(execute(DisplayEv,Format1,DisplayAg,DisplayT),
            and(time_before(<Now>,DisplayT),
             and(transaction(Trans,Payer3,Date3,Payee3,Amt3),
              and(Trans=transaction1#Id,
               and(sql_date_convert(Date,DBDate),
                and(Payee3=payee1#PayeeId,
                    Amt=amount(AmtNum,pounds)))))))))))
\end{verbatim}
Exploiting the fact that {\tt transaction} and {\tt execute} are both
functional on their first argument, this reduces after simplification
involving functional merging (see section~\ref{Functional-relations})
to
\begin{verbatim}
A([Payer,Trans,Date,Payee,Amt],
  impl(and(transaction(Trans,Payer,Date,payee1#bt,Amt),
           time_during(Date,
                       interval(date([1990,1,1]),
                                date([1990,12,31])))),
       E([TransId,DBDate,AmtNum,DisplayT],
         and(Trans=transaction1#TransId,
          and(sql_date_convert(Date,DBDate),
           and(Amt=amount(AmtNum,pounds),
            and(execute(DisplayEv,
                        display([TransId,DBDate,bt,AmtNum]),
                        clare,
                        DisplayT),
                time_before(<Now>,DisplayT))))))))
\end{verbatim}
Applying ``general''equivalences to translate {\tt time\_before}
and {\tt time\_during} into {\tt t\_precedes}, we get
\begin{verbatim}
A([Payer,Trans,Date,Payee,Amt],
  impl(and(transaction(Trans,Payer,Date,payee1#bt,Amt),
        and(t_precedes(date([1990,1,1]),Date),
            t_precedes(Date,date([1990,12,31])))),
       E([TransId,DBDate,AmtNum,DisplayT],
         and(Trans=transaction1#TransId,
          and(sql_date_convert(Date,DBDate),
           and(Amt=amount(AmtNum,pounds),
            and(execute(DisplayEv,
                        display([TransId,DBDate,bt,AmtNum]),
                        clare,
                        DisplayT),
                t_precedes(<Now>,DisplayT))))))))
\end{verbatim}
We then apply more ``general'' equivalences to translate {\tt execute}
into {\tt execute\_\-in\_\-future}, reducing the previous form to
\begin{verbatim}
A([Payer,Trans,Date,Payee,Amt],
  impl(and(transaction(Trans,Payer,Date,payee1#bt,Amt),
        and(t_precedes(date([1990,1,1]),Date),
            t_precedes(Date,date([1990,12,31])))),
       E([TransId,DBDate,AmtNum],
         and(Trans=transaction1#TransId,
          and(sql_date_convert(Date,DBDate),
           and(Amt=amount(AmtNum,pounds),
            and(execute_in_future(
                 display([TransId,DBDate,bt,AmtNum])))))))))
\end{verbatim}
Now we apply the equivalence that translates the conceptual predicate
{\tt trans\-action} into the database predicate {\tt TRANS}:
\begin{verbatim}
(and(transaction(Trans,Payer,Date,Payee,Amount),
     transaction_data_available(Date)) <->
 E([TransId,AMNum,PayeeName,DBDate],
   and(TRANS(TransId,DBDate,PayeeName,AMNum),
    and(named_object(Trans,TransId,transaction1),
     and(named_object(Payee,PayeeName,payee1),
      and(sql_date_convert(Date,DBDate),
       amount_object(Amoun,pounds,AMNum))))))) <-
assumably_=(Payer,
            organization#sri,
            payments_referred_to_are_from_SRI)
\end{verbatim}
When the equivalence is applied to the goal
\begin{verbatim}
transaction(Trans*,Payer*,Date*,payee1#bt,Amt*)
\end{verbatim}
its conjunctive context is the set
\begin{verbatim}
{t_precedes(date(1990,1,1),Date*),
 t_precedes(Date*,date(1990,12,31)}
\end{verbatim}
There are two conditions that must be proved from the context for
the rule to be applicable,
\begin{verbatim}
assumably_=(Payer*,
            organization1#sri,
            payments_referred_to_are_from_SRI)
\end{verbatim}
and
\begin{verbatim}
transaction_data_available(Date*)
\end{verbatim}
The first of these cannot be proved, but the assumption declaration
\begin{verbatim}
assumable(assumably_=(X,Y,Justification)
          0,
          Justification,
          specialization,
          true)
\end{verbatim}
(cf Section~\ref{Database-specific-axioms}) makes it permissible to
assume it with justification {\tt
payments\_\-referred\_\-to\_\-are\_\-from\_\-SRI}. The validity of the
other condition can be inferred from the context. The axioms needed to
do this are two Horn-clauses: one defines the period for which transaction
records exist, viz.
\begin{verbatim}
transaction_data_available(Date) <-
   and(t_precedes(date(1,8,89),Date2),
       t_precedes(Date,date(1,4,91)))
\end{verbatim}
and the other encodes the fact that
\verb!t_precedes! is transitive,
\begin{verbatim}
t_precedes(Date1,Date3) <-
   and(t_precedes(Date1,Date2),
       t_precedes(Date2,Date3))
\end{verbatim}
By chaining backwards through these to the
context, it is then possible to prove that
{\tt transaction\-\_data\-\_available(Date*)} holds,
and translate to the final form
\begin{verbatim}
A([TransId,Date,DBDate,AmtNum],
  impl(and(TRANS(TransId,DBDate,bt,AmtNum)
           and(sql_date_convert(Date,DBDate),
               and(t_precedes(date([1990,1,1]),Date),
                   t_precedes(Date,date([1990,12,31])))))
       execute_in_future(display([TransId,DBDate,bt,AmtNum]))))
\end{verbatim}
The final formula admits a finite proof procedure, (cf
sections~\ref{Effective-translation} and~\ref{Qopt}), since its
truth can be ascertained by an algorithm that can be summarized as
\begin{enumerate}
\item
Search for all tuples from the finite {\tt TRANS} relation.
\item
Check each to find the ones where the {\tt Payee}
field is equal to {\tt bt} and the {\tt Date} field is suitably
constrained by the arithmetic predicates {\tt sql\_date\_convert}
and {\tt t\_precedes}
\item
Assume that appropriate lists will in each such case be displayed by
the interface's causing the corresponding instantiated instance of the
{\tt execute\_\-in\_\-future} predicate to hold.
\end{enumerate}
The system has also managed to prove, under an abductive assumption
whose justification is {\tt
payments\_\-referred\_\-to\_\-are\_\-from\_\-SRI}, that the final
formula is equivalent to the original query.  Granted this assumption,
the user can be informed that the response is complete (cf
section~\ref{Command-functionality}).

\chapter{The Reasoning Engine}
\label{ReasoningEngine}
\label{reasoning}

CLARE's inference engine is used by several other components.  It
supports proofs from conjunctive contexts when performing translation
(cf. section~\ref{Translation-schemas}), and evaluates translated
queries; it is also employed to compile the lemmas used when
generating facts and questions (Chapter~\ref{genass}), and for various
other tasks during reference resolution which will not concern us
here. In each case, the basic functionality is some version of the
following.  We are provided with a Horn-clause theory $\Gamma$, a
formula $F$ (which may contain free variables $\bf \vec{x}$), and a
criterion which determines permissible abductive assumptions. The task
is to find a substitution $\theta$ on $\bf \vec{x}$ and a set of
permissible assumptions $A$ such that
\begin{eqnarray}
\Gamma \Rightarrow (A \rightarrow \theta(F))\label{Basic-infer}
\end{eqnarray}
or terminate with the information that no such $\theta$ and $A$
exist. This chapter describes how the inference functionalities are
realized in CLARE as a type of Prolog meta-interpreter.
Since these have been
described many times in the literature (e.g. (Sterling and Shapiro, 1986)),
we will concentrate on the unusual features of the CLARE inference
engine. The expositional strategy used will be to describe a simplified
version of the inference engine, beginning with a
the basic meta-interpreter for pure Horn-clause theories
shown in figure~\ref{Basic-interpreter} and adding features as we
go along to cover new functionalities.
\begin{figure}
\hrule
\begin{verbatim}
prove(and(L,R)) :-
  prove(L),
  prove(R).

prove(G) :-
  horn_clause_in_theory(G <- Body),
  prove(B).

prove(G) :-
  unit_clause_in_theory(G).
\end{verbatim}
\hrule
\caption{Basic Horn-clause interpreter\label{Basic-interpreter}}
\end{figure}

\section{Logical structure}\label{Reasoning-basic}

The basic form of the inference engine is that of
a slightly extended Prolog meta-interpreter. Superimposed
on top of it, there is an iterated-deepening A* search strategy.
We will only consider the logical aspects of
the inference engine in this section,
postponing discussion of the search strategy
until section~\ref{Inference-search}. There are two important
extensions over the standard Prolog meta-interpreter: these
are to cover abductive proofs and proofs of universally quantified
implications. We discuss these in Sections~\ref{Abductive-proof}
and~\ref{Prove-implications}. In Section~\ref{Negated-proofs}
we briefly consider the treatment of negation.

\subsection{Abductive proofs}\label{Abductive-proof}

We will use the term ``abductive proof'' in this section to mean
construction of a proof-tree in which some leaves are undischarged
assumptions. CLARE performs abductive proofs for two essentially
distinct reasons. The first has been referred to many times already in
chapters~\ref{translation},~\ref{AET} and~\ref{dmi}; when using logic
to reason about common-sense concepts, such as those expressed in
natural language, it is frequently necessary to assume background
knowledge which is not part of the explicit content of the utterance's
logical representation. In this case, the abductively assumed
conditions are normally shown to the user in order to check their
appropriateness.  Abductive proofs are also used in order to compile
``implicational lemmas'', which are utilized by the generation
component (see Chapter~\ref{genass}) and several other parts of the
system not described in the thesis. In this mode of operation, the
reasoning engine is given a goal $F$ containing free variables
$\vec{x}$ and a set of criteria for making abductive assumptions; the
assumptions may also contain free variables. If an abductive proof can
be found with assumptions $A$ and a substitution $\theta$ for the
$\vec{x}$, this can be regarded as a strict (non-abductive) proof for
the Horn-clause
\begin{displaymath}
\forall \vec{y}.(A \rightarrow \theta(F))
\end{displaymath}
where $\vec{y}$ are the free variables in $\theta(\vec{x})$. As part
of the process of compiling a linguistic domain theory, all
implicational lemmas of a particular type are generated.  At present,
the criterion used for limiting abductive assumptions restricts $A$ to
being a set of which all but at most one member are {\it arithmetic
relation goals} (cf section~\ref{Effective-translation}). The
remaining member of $A$ must be an atomic predication whose predicate
is a conceptual predicate.  The net result is to prove an implication
of the form $$Arith \rightarrow (Conceptual \rightarrow G)$$ where $G$
is an atomic goal, $Conceptual$ is a conceptual goal, and $Arith$ is a
conjunction of arithmetical relation goals. These lemmas can be used
to find quickly the ``consequences'' of a goal that matches
$Conceptual$ (cf~\ref{Talk-about}).

Two minor changes
are needed to extend a Prolog meta-interpreter to allow abductive
proofs: another argument is added to the main
predicate, which passes around a difference-list holding the assumptions,
and an extra clause is added which allows ``proof'' of an atomic
goal by adding it to the assumption-list, if the abductive assumability
criteria permit this. The basic interpreter, modified in the way described,
is shown in figure~\ref{Abductive-interpreter}.
\begin{figure}
\hrule
\begin{verbatim}
prove(and(L,R),AssumptionsIn-AssumptionsOut) :-
  prove(L,AssumptionsIn-AssumptionsNext),
  prove(R,AssumptionsNext-AssumptionsOut).

prove(G,AssumptionsIn-AssumptionsOut) :-
  horn_clause_in_theory(G <- Body),
  prove(Body,AssumptionsIn-AssumptionsOut).

prove(G,AssumptionsIn-AssumptionsIn) :-
  unit_clause_in_theory(G).

prove(G,AssumptionsIn-[G|AssumptionsIn]) :-
  abductively_assumable(G).
\end{verbatim}
\hrule
\caption{Abductive Horn-clause interpreter\label{Abductive-interpreter}}
\end{figure}

\subsection{Proving implications}\label{Prove-implications}

We now consider the task of extending the inference engine to
be able to prove formulas of type
\begin{quote}
\begin{verbatim}
forall(Vars,impl(LHS,RHS))
\end{verbatim}
\end{quote}
where the variables in {\tt Vars} occur free in {\tt LHS} and {\tt RHS}.
Two well-known strategies exist for attempting to prove such expressions;
they can be proved either {\it intensionally} or {\it extensionally}.
Intensional proofs are carried out by substituting unique constants
for the {\tt Vars}, assuming {\tt LHS}, and attempting to prove {\tt RHS};
extensional proofs by listing all values of {\tt Vars} for which a proof
of {\tt LHS} can be found, and for each one ascertaining that a proof
of {\tt RHS} exists. The CLARE inference engine can use either strategy.

To implement the intensional proof strategy it is necessary to add yet another
argument to the {\tt prove} predicate in the interpreter, to hold the
set of assumptions derived from the {\tt LHS}; it is tempting to try
and include them in the abductive assumption list, but in practice this
manoeuvre does not seem to simplify matters.
The need to
take account of possible abductive assumptions also introduces
some complications in the extensional
strategy, since it is necessary to collect the
assumptions used for each separate proof of {\tt RHS}. The skeleton
interpreter, modified to allow proof of universally quantified
implications by the intensional and extensional strategies,
is presented in figure~\ref{Implic-interpreter}. Here the first
clause of form {\tt prove(forall...)} is for the intensional
strategy; {\tt replace\_\-by\_\-unique\_\-constants} is assumed to be a
predicate which replaces all the variables in its argument
by unique ground terms. The following clause is for the
extensional strategy. Here
{\tt safe\_bagof(X,G,Xs)} is like
{\tt bagof(X,G,Xs)}, with the difference that it first binds any
free variables in G, and succeeds with {\tt Xs=[]} if there is no
proof of {\tt G}; {\tt union\_\-n\_\-1(List,Union)} takes a list of lists
as its first argument and returns their union as the second one.

\begin{figure}
\hrule
{\tt (All clauses from the interpreter in figure~\ref{Abductive-interpreter})}
\begin{verbatim}

prove(G,AssumptionsIn-AssumptionsIn,ImplAssumptions) :-
  member(G,ImplAssumptions).

prove(forall(Vars,impl(LHS,RHS)),AssIn-AssOut,ImplAss) :-
    replace_by_unique_constants(Vars),
    add_implicational_assumptions(LHS,ImplAss,ImplAss1),
    prove(R,AssIn-AssOut,ImplAss1).

prove(forall(Vars,impl(LHS,RHS)),AssIn-AssOut,ImplAss)
    safe_bagof(environment(Vars,AssOutLeft),
               prove(LHS,AssIn-AssOutLeft,ImplAss)
               LeftEnvironmentList),
    prove_forall_cases(LeftEnvironmentList,Vars^RHS,
                       ImplAss,AssOutList),
    union_n_1(AssOutList,AssOut).

prove_forall_cases([],_Vars^_RHS,_ImplAss,[]).
prove_forall_cases([EnvFirst|EnvRest],Right,ImplAss,
                   [AssOutFirst|AssOutRest]) :-
    prove_forall_case(EnvFirst,Right,ImplAss,AssOutFirst),
    prove_forall_cases(EnvRest,Right,ImplAss,AssOutRest).

prove_forall_case(environment(VarsLeft,AssOutLeft),
                  Vars^RHS,ImplAss,AssOut) :-
    copy_term([Vars^RHS,ImplAss],[Vars1^RHS1,ImplAss1]),
    VarsLeft = Vars1,
    prove(RHS1,AssOutLeft-AssOut,ImplAss1).

add_implicational_assumptions(and(P,Q),ImplAssIn,ImplAssOut) :- !,
    add_implicational_assumptions(P,ImplAssIn,ImplAssNext),
    add_implicational_assumptions(Q,ImplAssNext,ImplAssOut).
add_implicational_assumptions(G,ImplAss,[G|ImplAss]) :-
    atomic_goal(G), !.
add_implicational_assumptions(_Other,ImplAss,ImplAss).
\end{verbatim}
\hrule
\caption{Abductive interpreter for Horn-clauses and implications
\label{Implic-interpreter}}
\end{figure}

\subsection{Proving negated goals}\label{Negated-proofs}

The inference engine contains no general treatment of negation, but in
order to deal with cases where {\tt Goal} is explicitly contradicted
by its context it is possible to define rules of the form
\begin{verbatim}
neg(Goal) <- Body
\end{verbatim}
where {\tt Goal} is an atomic formula and {\tt Body} is an arbitrary
formula.  A rule of this type has the content ``the negation of {\tt
Goal} is implied by {\tt Body}'', and is formally a normal
Horn-clause; note that there is no connection, as far as the inference
engine is concerned, between the goals {\tt G} and {\tt not(G)}.  Rules
with negated heads are only invoked to attempt to identify
inconsistent uses of assumptions (cf.
Section~\ref{Abductive-translation}).

\section{Search strategies for inference}\label{Inference-search}

In this section we consider the search strategy used in the inference
engine, which as already stated is an iterated-deepening A* (IDA*)
search strategy (Korf, 1986, Stickel, 1986); that is, search proceeds
by first defining a cost limit $D$, and then performing a depth-first
search where each branch is cut off when the sum of the cost of goals
already processed, together with an estimate of the cost of solving
pending goals, exceeds $D$. The cost function used at present charges
one unit for each rule application, and conservatively estimates one
unit for each pending goal; as explained in
section~\ref{Abductive-translation}, it is possible to define extra
costs for abductively assumed goals.  The motivation for using IDA*,
rather than simple depth-first search, is that it avoids getting stuck
in infinite recursions; it essentially performs the function of
simulating breadth-first search, and is thus theoretically complete.
However, experiment quickly shows that infinite recursion still causes
severe practical difficulties without further elaboration of the
method.

There are basically three problems, which to some extent are related.
Firstly, it is easy to spend excessive amounts of time in infinite
branches of the search tree, especially ones resulting from
``transitivity'' axioms (cf (Appelt and Hobbs 90)), and Horn-clause
readings of axioms of the form $$(p(X,Y) \equiv p(X,Z)) \leftarrow
q(Y,Z)$$ (cf~Section~\ref{Equiv-transfer-properties}). The problem
becomes even more acute when both ``normal'' and ``backward''
Horn-clause readings of equivalences are used
(cf~Section~\ref{Horn-clause-readings}); it is then easy for the
inference engine to get into situations where it wanders aimlessly
around in the search-space, alternately using ``normal'' and
``backward'' Horn-clause versions of equivalences.  Secondly, even
when infinite branches as such are not a problem, it may be the case
that there are multiple redundant proofs of a goal. The third problem
arises from inefficient ordering of conjunctions. When trying to prove
a conjunction, it can sometimes be the case that there are many proofs
of the first conjunct (which take a correspondingly long time to
generate), but that the second conjunct can quickly be discovered to
have either one or no proofs; in this case attempting to prove the
conjuncts in a different order can pay dividends.  The non-obvious
aspects of the search strategy, which we will now describe, stem from
attempts to counteract these problems. We describe them one at a time.

\subsection{Avoiding infinitely recursive branches}

The approach we currently use to help minimize time spent in
infinitely recursive branches of the search-tree is based on a
standard loop-avoidance method. It can be proved that it is sound to
cut off search if a goal is {\it identical} to one of its ancestors.
The inference engine uses this criterion, both in its simple form and
in a ``look-ahead'' mode: when expanding a goal $G$ to a conjunction
$B_1\wedge B_2 \ldots$, it checks before adding the $B_i$ to the stack
of pending goals, to make sure that none of them are identical to an
ancestor.  Perhaps surprisingly (since this is a fairly expensive
operation), it turns out that performing the ``look-ahead'' check
increases efficiency quite substantially.

It would be pleasant if the ``identity'' condition could be
relaxed a little, for example by replacing it with a test of
unification or subsumption with one of the ancestor goals.
Unfortunately, it is easy to prove that weakening the condition
is in general not correct. However,
a strong syntactic similarity between a goal and an ancestor is
often a good heuristic indicator that a loop has arisen, and can be taken as
a justification for exacting a ``penalty'', causing the
cost limit to be reached more quickly. In the current version,
the penalty is applied if the goal is subsumed by an ancestor,
a strategy which we have found to work well in practice. The method is
sound, since its only effect is to devote
less effort to search of the affected branches without cutting
them off completely, and appears rather more general than
the approach advocated in (Appelt and Hobbs 90), which involves recoding
of the meaning postulates.

The loop avoidance method described above is strong enough that it is
at any rate possible to consider using both forward and backward
versions of equivalences. Use of the backward versions still slows
down the inference engine by more than a factor of ten, when it is
used to support translation with the PRM LDT. For this reason, we
decided to remove backward Horn clauses. It turned out in practice
that the potential inferences lost due to this step were fairly
unimportant, and that the ``inference holes'' which appeared could
easily be filled by adding a few appropriate hand-coded clauses.

\subsection{Avoiding redundant search}

The second problem, that of avoiding redundant
recomputation, is currently only
applied to ground goals, that is goals which contain
no uninstantiated variables. We have observed in many cases
that, without some kind of restrictions,
a great deal of effort is wasted in producing multiple
proofs for goals of this kind. Since a proof of a ground goal
is basically just a ``yes'' or ``no'', it is tempting to
adjust the backtracking mechanism so as to
block attempts to find new proofs of ground goals that have already
been successfully proved. However, this is not quite right;
the problem is that the ordering of the search-space may mean
that the first proof produced is an expensive one, which
leaves the search process on the verge of exceeding the cost
limit, and prevents further progress. The correct realization of
the idea seems rather to be to block retrying of ground goals
only if the cost of the existing proof falls under a given
bound, which we have at present arbitrarily fixed as a third
of the total cost limit.

\subsection{Dynamic re-ordering of conjuncts}

The time taken to find a proof for a conjunctive goal (or to ascertain
that there is no proof) can often be strongly dependent on the order
in which the conjuncts are attempted. We have made no attempt to solve
this problem in its general form, but two special cases are
sufficiently important that the inference engine needs to take account
of them. Suppose first that the interpreter is about to attempt to
prove the conjunction $L\wedge R$.  If it is the case that there is no
proof of $R$, then time lost establishing this can be recovered by
omitting the attempt to prove $L$. Thus it can pay to perform
``look-ahead'' and search for goals which can quickly be shown to be
unachievable. Similarly, if $L$ contains free variables which also
occur in $R$ then the time needed to find a proof can be reduced if
the number of proofs for $R$ is substantially less than the number of
proofs for $L$, since instantiating free variables nearly always has
the function of reducing the number of possible proofs. In particular,
if it can quickly be shown that there is only {\it one} proof of $R$
then it is very likely that it will be correct to attempt it first.

It is possible to include declarations of the form
\begin{quote}\begin{verbatim}
quick_failure_test(<Goal>) :- <Conds>
\end{verbatim}\end{quote}
or
\begin{quote}\begin{verbatim}
quick_determinism_test(<Goal>) :- <Conds>
\end{verbatim}\end{quote}
where \verb!<Goal>! is an atomic goal and \verb!<Conds>! an arbitrary
Prolog form; the intended semantics are that there should be, respectively,
no proofs or exactly one proof of \verb!<Goal>! if \verb!<Conds>! hold.
For declarations of this kind to be useful, \verb!<Conds>! should be
quickly evaluable. In the present version of CLARE, instances of
\verb!<Conds>! never do more than check instantiations of variables,
or attempt to unify them with other terms.
Tests for both cases are applied when a
Horn-clause is used to expand a goal {\tt G}; after the head of
the clause has been unified with {\tt G}, the goals in the body
are examined. If a {\tt quick\_\-failure\_\-test} succeeds then
proof of the whole conjunction fails; if a {\tt quick\_\-determinism\_\-test}
succeeds on a subgoal, then that subgoal is evaluated first. There
are currently about 30 ``quick failure and determinism'' declarations,
all of them for the ``naming'' predicates described in
section~\ref{Names-and-objects}.

\section{Finding finite proof procedures}\label{Qopt}

In the final section of this chapter we describe the module responsible
for searching for finite proof procedures
(see also section~\ref{Effective-translation}). Since the inference
engine is basically a Prolog meta-interpreter, it is sensitive to
the ordering of conjuncts: it attempts to prove a goal of the form
{\tt and(P,Q)} by first proving {\tt P} and then proving {\tt Q}. If
{\tt P} and {\tt Q} share a free variable {\tt X}, the order can be important.
To repeat the example from~\ref{Effective-translation}, if $TRANS/3$ is a
database predicate then the strategy ``find an $X$
such that $X > 100$, then find values of
$Y$ such that $TRANS(john,X,Y)$'' is not a finite strategy; however, reversing
the order to make the strategy ``find $X$ and $Y$ such that
$TRANS(john,X,Y)$, then determine whether $X > 100$ holds'' is finite.

The search for finite proof strategies is carried out by another extended
Prolog meta-interpreter, using an abstract interpretation method.
The information needed to deal with the base case of attempting to
prove an atomic goal is provided by declarations of the form
\begin{quote}
\begin{verbatim}
call_pattern(<Pred>(Arg1,Arg2,...Argn),InArgs,OutArgs)
\end{verbatim}
\end{quote}
where \verb!<Pred>! is a predicate and {\tt InArgs} and {\tt OutArgs}
are subsets of the set {\tt \{Arg1, ..., Argn\}}.
The intended semantics are that there are finitely many proofs of
the goal
\begin{quote}
\begin{verbatim}
<Pred>(Arg1,Arg2,...Argn)
\end{verbatim}
\end{quote}
if the variables in {\tt InArgs} are already instantiated,
and that each of them will
result in the arguments {\tt OutArgs} becoming instantiated.
The basic structure of the interpreter is shown in
figure~\ref{Qopt-interpreter}; only the clauses for conjunctive
and atomic goals are displayed.
\begin{figure}
\hrule
\begin{verbatim}
rearrange(OldForm,NewForm) :-
  copy_term(OldForm,EvalForm),
  rearrange0(OldForm,NewForm,EvalForm,[]-[],[]-_), !.

rearrange0(Old,New,Eval,FrozenI-FrozenO,DoneI-DoneO) :-
  rearrange1(Old,New0,Eval,FrozenI-FrozenN,DoneI-DoneN),
  thaw(New0,New,FrozenN-FrozenO,DoneN-DoneO).

rearrange1(and(LOld,ROld),and(LNew,RNew),and(L,R),
           FrozenIn-FrozenOut,DoneIn-DoneOut) :- !,
  rearrange0(LOld,LNew,L,FrozenIn-FrozenNext,DoneI-DoneN),
  rearrange0(ROld,RNew,R,FrozenNext-FrozenOut,DoneN-DoneO).
rearrange1(OldG,OldG,G,Frozen-Frozen,Done-[goal_done|Done]) :-
  abstract_call(G),!.
rearrange1(OldG,true,G,Frozen-[frozen(G,OldG)|Frozen],D-D).

thaw(Form,Form,[]-[],Done-Done) :- !.
thaw(Form,OutForm,
     [frozen(FrozenEval,FrozenOld)|FrozenNext]-FrozenOut,
     DoneI-DoneO) :-
  rearrange1(FrozenOld,FrozenNew,FrozenEval,[]-[],[]-[_|_]),!,
  thaw(and(Form,FrozenNew),OutForm,FrozenNext-FrozenOut,
       [goal_done|DoneI]-DoneO).
thaw(Form,OutForm,
     [FrozenFirst|FrozenNext]-[FrozenFirst|FrozenOutRest],
     DoneI-DoneO) :-
  thaw(Form,OutForm,FrozenNext-FrozenOutRest,DoneI-DoneO).

abstract_call(X=Y) :-
   X = Y, !.
abstract_call(_X=_Y) :- !.
abstract_call(G) :-
   call_pattern(G,InArgs,OutArgs),
   is_instantiated(InArgs),
   make_instantiated(OutArgs).
\end{verbatim}
\hrule
\caption{Interpreter for finding finite proof strategies
\label{Qopt-interpreter}}
\end{figure}
In the implemented interpreter,
there is a clause for each logical operator of the TRL language.
The central idea is to rearrange
the expression by using an abstract version of the Prolog
``freeze'' or delayed-goal primitive.
The arguments of the main predicates
{\tt rearrange0} and {\tt re\-arr\-ange1} are
\begin{quote}
\begin{verbatim}
rearrange0/1(Original,Rearranged,Eval,Frozen,Done)
\end{verbatim}
\end{quote}
where
\begin{description}
\item[{\tt Original}] is the original expression.
\item[{\tt Rearranged}] is the rearranged expression.
\item[{\tt Eval}] is a copy of {\tt Original} used to keep track of
      the variables currently instantiated in the abstract interpretation.
\item[{\tt Frozen}] is a difference-list of delayed goals, i.e. goals which
     were insufficiently instantiated to be executed when first called.
\item[{\tt Done}] is a difference-list that keeps track of progress made
     in evaluating goals, to prevent infinite postponement of goal
     execution.
\end{description}
The two predicates call each other recursively, with {\tt rearrange0}
calling {\tt re\-arr\-ange1} and then trying to ``thaw'' any frozen goals
that may have been created.
Note the special treatment of equality goals: if {\tt Eval} is
an equality, its arguments are unified if possible.


\chapter{Generation of Utterances}
\label{gentrl}\label{genass}

\section{Introduction}

So far, we have concentrated on interfacing functionalities which
involve translating from natural language primitives into database
primitives.  This chapter describes a simple method that has been
implemented in CLARE to handle translation in the reverse direction;
the input is expressed in terms meaningful to the database, and the
output is in the form of natural language utterances. We will as usual
mainly be concerned with the part of the translation process that
involves inferences, but to put the work in its proper context we
first briefly discuss CLARE's overall strategy with regard to
generation.

Recall that CLARE has two main abstract levels of representation:
Quasi Logical Form (QLF) and TRL. QLF is the representation that is
output by the first, compositional, phase of semantic analysis.  The
rules used to relate QLFs to surface strings are coded entirely in
terms of unification, and are both in theory and in practice fully
reversible. Generation of surface form from QLF is performed by the
Semantic Head-Driven generation algorithm (Shieber {\it et al} 1990),
and will not be discussed further here; we will instead concentrate on
the problem of producing suitable QLF representations from database
input, expressed in terms of TRL formulas. We will consider generation
of both assertions and questions, as described in
Section~\ref{Generation-functionality}.  Assertions will be generated
to describe complete database records; questions to ask for values of
unfilled fields in incomplete records.

To understand the problems involved in synthesizing an appropriate
QLF representation for a TRL formula, it will be helpful to begin by
considering it as the inverse of the analysis process, which has QLF
as input and database TRL as output. As described in
Section~\ref{CLE-overview}, the analysis process has two main phases:
the resolution phase, in which referentially vague expressions are
replaced by concrete values; and the translation phase, in which
linguistic predicates are translated into database predicates. Both of
these phases need to be inverted. The reason why this task is
non-trivial is that a good deal of information pertaining to surface
linguistic form is lost in the transition from QLF to TRL. There are
two main problems.

The first is that the resolution phase can in general be arbitrarily
non-deterministic in the reverse direction. For example, a given
definite NP will normally only be able to refer to a small number of
possible domain objects; conversely, however, there are a potentially
infinite number of definite noun-phrases that can be used to refer to
a given domain object. The generation of definite descriptions is a
major research topic in its own right, and in order to keep the focus
on our primary area of concern, the translation problem, we restrict
ourselves by only considering resolution rules that are deterministic
in the reverse direction.  Typical examples of such rules are those
which resolve proper name expressions, numbers and date expressions.
In the reverse (TRL to QLF) direction, these take as input a TRL term
representing a typed object, number, or date, and return a QLF {\tt
term} construct which encodes the surface semantic structure of an
appropriate noun-phrase.  We will see examples of use of these rules
later in the chapter.

The second problem is caused by the loss of hierarchical structure
involved in the transition from QLF to TRL. For example, an NP
modified by a relative clause will at QLF level contain the
contribution of the relative as a distinct sub-form. After conversion
to TRL, however, the structure is ``flattened''; the head-noun and all
its modifiers, including the relative, will be at the same level. Our
initial implementation solves the second problem by tightly limiting
the range of possible QLFs that can be produced; the process of
producing QLF from TRL is not free, but is rather constrained to
follow one of a set of ``template'' rules. These templates are
automatically produced by generalizing specific training examples,
using a variant of the so-called ``case-based learning'' method. The
learning procedure is described in Section~\ref{Case-based-learning}.
It produces rules of the approximate form
\begin{quote}\begin{verbatim}
qlf_for_trl(<QLF>) :- <Body>
\end{verbatim}\end{quote}
where \verb!<Body>! is a set of TRL goals. (We refer to these as TRL
description generation rules, or simply TRL generation rules.) The
goals in \verb!<Body>!  can be of several different types: they
include the component conjuncts in the TRL form that is being
generating from, and conditions constraining {\tt term}s in
\verb!<QLF>! to refer to TRL constants. Examples of learned TRL
description rules are given in the next section.

The set of TRL description rules is not complete; a QLF can be produced
only if it is of the same form as one of the training examples.
They are not necessarily guaranteed to be sound either, though
we have not in practice observed any unsound rules to have been
inferred by the process. The status of the rules is thus that of
{\it generation heuristics}, induced by analysis of specific
examples. To check the soundness of a rule, it necessary to make
sure that the original TRL can be recovered from the proposed QLF.

Loss of completeness is not necessarily important, since there is no
need to be able to produce {\it every} way of expressing something.
Indeed, everyday experience suggests that people generally make use of
an array of stock phrases when expressing themselves; there is no
obvious reason to believe that one has to have a general mechanism for
generation at this level. The unsoundness of the method is more
significant, since checking correctness is moderately expensive. One
way to attack this problem might be to attempt to infer rules using
the Explanation-Based Learning method (Hirsh 1987, Rayner 1988); the
application of reference resolution methods would be conditions on the
inferred rules.  In the rest of this chapter we describe in more
detail TRL generation as implemented in CLARE.

\section{Learning case-based description rules}
\label{Case-based-learning}

The learning method used for acquiring TRL description rules is
extremely simple, and is perhaps best introduced by an example.
Suppose that the training example is the sentence
\begin{quote}
{\it John is a dog.}
\end{quote}
uttered in a context in which ``John'' can be used to refer to the entity
{\tt john1}.
This produces the QLF
\begin{verbatim}
[dcl,
 form(verb(pres,no,no,no,y),
      A,
      B^
      [B,
       [be,
        A,
        term(proper_name(tpc),C,D^[name_of,D,John],E,F),
        G^[eq,G,term(q(H,a,sing),I,J^[dog1,J],K,L)]]],
      M)]
\end{verbatim}
(the details of the QLF representation are not interesting).  After
resolution and simplification, this produces the TRL form
\begin{quote}\begin{verbatim}
dog1(john1)
\end{verbatim}\end{quote}
The idea is to generalize the training sentence by looking for
correspondences between the QLF and the TRL representations,
and hypothesizing that they are instances of a general rule.
In our example, there are two such correspondences.
\begin{enumerate}
\item The predicate symbol {\tt dog1} occurs in both QLF and
TRL.
\item The QLF representation contains the QLF term
\begin{quote}
\begin{verbatim}
term(proper_name(tpc),C,D^[name_of,D,John],E,F)
\end{verbatim}
\end{quote}
and the TRL representation contains the TRL term {\tt john1}, and
there is a deterministically reversible resolution rule which
makes the TRL term the referent of the QLF term.
\end{enumerate}
In order to generalize the example, we hypothesize that the connection
between the QLF and TRL representations depends on the linguistic
properties associated with the predicate {\tt dog1}, and
the referential connection between the two terms. We make the first
of these precise by using the predicate {\tt genpred}, which links
a predicate symbol to the lexicon entries in which it appears.
Here, the relevant clause of {\tt genpred} is
\begin{quote}\begin{verbatim}
genpred(dog1,([dog1,_],nbar(_,_)))
\end{verbatim}\end{quote}
-- that is, \verb!dog1! is a one-place predicate and a common
noun (\verb!nbar!) sense. Our precise hypothesis with regard to
replacing \verb!dog1! is then that the information about it {\tt
genpred} constitutes a sufficiently detailed description of its
properties that any other predicate with a similar entry would be
processed in the same way.  So we are guessing, for example, that in a
context where ``Mary'' can be used to refer to the entity {\tt mary1},
the TRL
\begin{quote}\begin{verbatim}
cat1(mary1)
\end{verbatim}\end{quote}
could have been produced from the QLF
\begin{quote}\begin{verbatim}
[dcl,
 form(verb(pres,no,no,no,y),
      A,
      B^
      [B,
       [be,
        A,
        term(proper_name(tpc),C,D^[name_of,D,Mary],E,F),
        G^[eq,G,term(q(H,a,sing),I,J^[cat1,J],K,L)]]],
      M)]
\end{verbatim}\end{quote}
our justification being that {\tt cat1} has a similar {\tt
genpred} entry, i.e.
\begin{quote}\begin{verbatim}
genpred(cat1,([cat1,_],nbar(_,_)))
\end{verbatim}\end{quote}
and that there is a reversible resolution rule which given the atom
{\tt mary1} can find out that the QLF expression
\begin{quote}
\begin{verbatim}
term(proper_name(tpc),C,D^[name_of,D,Mary],E,F)
\end{verbatim}
\end{quote}
would refer to it. In the general case, our heuristic rule will be to
guess that
\begin{verbatim}
[dcl,
 form(verb(pres,no,no,no,y),
      A,
      B^
      [B,
       [be,
        A,
        <Term>,
        G^[eq,G,term(q(H,a,sing),I,J^[<PredAtom>,J],K,L)]]],
      M)]
\end{verbatim}
is a QLF from which it would be possible to derive the TRL form
\begin{quote}\begin{verbatim}
<PredAtom>(<Ref>)
\end{verbatim}\end{quote}
under the assumptions that \verb!<PredAtom>! is a atom, \verb!<Term>!
is a term, the \verb!genpred! entry for \verb!<PredAtom>! is
\begin{quote}\begin{verbatim}
genpred(<PredAtom>,([<PredAtom>,_],nbar(_,_)))
\end{verbatim}\end{quote}
and \verb!<Ref>! is the referent of \verb!<Term>!.
This information is what is encoded in the induced rule,
\begin{verbatim}
trl_gen([dcl,
         form(verb(pres,no,no,no,y),A,
              B^[B,[be,
                    A,
                    Term,
                    C^[eq,C,term(q(_,a,sing),_,
                                 D^[Pred,D],_,_)]]],
              _)],
        [Pred,Ref],
        [entity_ref_term(Ref,Term),
         trlgoal([Pred,Ref],
                 genpred(Pred,(E^[Pred,E],nbar(_,_))))])
\end{verbatim}
The three arguments to the rule are, in turn, the generalized QLF; the
associated generalized TRL representation; and the conditions of
applicability.  The rule is capable of producing QLFs for sentences of
the form
\begin{quote}
\var{NP} is a/an \var{Noun}.
\end{quote}
where \var{NP} is any referring expression for which there is a
reversible resolution method.  This includes NPs like ``John'', ``Mary
(the employee)'', ``1990 (the year)'' and ``1/1/89 (the day)''.

%

\section{Description of database entities}
\label{Talk-about}

We now show how CLARE can use the TRL description rules to produce
sentences describing database objects. This functionality is in the
implemented system linked to the semantics of English verbs like
``talk about (something)'' and ``tell (someone) about (something)''.
Thus for example processing of a sentence such as
\begin{quote}
{\it Talk about Peter Piper.}
\end{quote}
will produces a call to describe the conceptual object referred to
by the expression {\it Peter Piper}, namely
\begin{quote}\begin{verbatim}
employee1#'Peter Piper'
\end{verbatim}\end{quote}
The description module is passed the conceptual object $X$ that is to be
described, and begins by finding all conceptual goals $C_i$
(see section~\ref{DBLDT}) which have the property that at least
one argument position is filled by an occurrence of $X$. This
provides an abstract ``conceptual-level'' summary of ``what the
database knows about $X$''. The main part of the processing is then
carried out by two sub-modules,
the {\it utterance synthesizer} and the {\it utterance filter},
which are called in alternation.
The synthesizer uses the TRL description rules to attempt to find new
utterances whose truth-conditions are implies by some $C_i$.
After each such utterance is
found, but before it is displayed, the utterance filter is called
into play to determine whether or not the new utterance should be
shown to the user, and if so whether any more utterances need be searched for.
We illustrate the behaviour of the description module by sketching
the course of processing for the example given immediately above.
The first $C_i$ found is the predication {\tt (C1)}
\begin{verbatim}
employee(organization1#sri,employee1#'Peter Piper',m,y)    (C1)
\end{verbatim}
which conveys the information that the employee Peter Piper is a employee
at the organization SRI, is of sex {\tt m} (for ``Man''), and has a car
({\tt y} as opposed to {\tt n}).

The utterance synthesizer is called first. When passed a conceptual
goal $C_i$, the synthesizer begins by computing the set of goals
logically implied by the union of $C_i$ with the current linguistic
domain theory {\it excluding} the database: the result is then uses as
input to the learned TRL description rules.  The efficiency of the
process is increased by pre-compiling the set of consequences of a
generic representative of each class of tuple, and caching the results
in the form of ``implicational lemmas''.  The relevant lemmas are
of the form
\begin{quote}\begin{verbatim}
Conditions -> (<C> -> <L>)
\end{verbatim}\end{quote}
where \verb!<C>! is a conceptual goal matching $C_i$,
{\tt Conditions} is a conjunction of evaluable
predicates, and \verb!<L>! is an atomic goal whose predicate
is a linguistic predicate
(cf. section~\ref{Abductive-proof}).
Continuing with the example, synthesis from the goal {\tt (C1)}
proceeds as follows. First the TRL lemmas are used to find
the set of consequences of {\tt (C1)}.
When the concrete values for the relevant instances have been
instantiated, the result is a list which includes the goals
\begin{quote}\begin{verbatim}
car1(sk12(employee1#'Peter Piper'))
have1(sk13(employee1#'Peter Piper'),
      employee1#'Peter Piper',
      sk12(employee1#'Peter Piper')),
male1(employee1#'Peter Piper'),
man1(employee1#'Peter Piper'),
name_of1('Peter Piper',employee1#'Peter Piper'))),
sex_of1(Male,employee1#'Peter Piper'),
person1(employee1#'Peter Piper'),
at1(employee1#'Peter Piper',organization1#sri),
employee1(employee1#'Peter Piper'),
\end{verbatim}\end{quote}
Here, {\tt sk12(X)} and {\tt sk13(X)} are Skolem functions representing,
respectively, the car employee {\tt X} has (if such an object exists),
and the event of his or her having it. It is now possible
to attempt to apply learned description rules, some of which succeed.
For example, the first two ``consequence'' goals,
\begin{quote}\begin{verbatim}
car1(sk12(employee1#'Peter Piper'))
have1(sk13(employee1#'Peter Piper'),
      employee1#'Peter Piper',
      sk12(employee1#'Peter Piper')),
\end{verbatim}\end{quote}
licence an application of
the rule generalized from
\begin{quote}
{\it John loves a cat.}
\end{quote}
with {\tt employee1\#\-Peter Piper} substituting {\tt john1}, {\tt
have1} substituting {\tt love1} and {\tt car1} substituting {\tt cat1}
in the conditions. Now the noun phrase {\it Peter Piper (the
employee)} can be produced as an NP whose referent is
{\tt employee1\#Peter Piper}; the relevant resolution rule
is reversible, as there is sufficient structure in the typed variable.
Synthesis proceeds from the QLF produced, yielding the sentence
\begin{quote}
{\tt Peter Piper (the employee) has a car.}
\end{quote}
Before the sentence is displayed, the utterance filter is called.
The filter keeps track of the combined propositional
content of the sentences so far uttered when describing the current tuple,
which is stored as a global variable $P_{old}$. (Initially, the value
of $P_{old}$ is $true$).
When the filter is passed a new candidate utterance with propositional
content $P$, it first ascertains whether or not $P$ is implied by
$P_{old}$. If this is the case, it returns the information that the new
utterance is redundant and need not be displayed. If not, it displays
the new utterance, and updates the combined propositional content to
$P_{old}\wedge P$. It then checks again to find out whether $P_{old}\wedge P$
is equivalent to the current tuple. If so, no more utterances need to be
generated.

The expression representing the content is first transformed if necessary
by replacing Skolem functions with existentially quantified variables
and then translated into conceptual form. Thus
the propositional content of the first utterance is originally
\begin{verbatim}
and(car1(sk12(employee1#'Peter Piper')),
    have1(sk13(employee1#'Peter Piper'),
          employee1#'Peter Piper',
          sk12(employee1#'Peter Piper')))
\end{verbatim}
After replacement of Skolem functions by existentially quantified variables,
this becomes
\begin{verbatim}
exists([X,Y],
   and(car1(X),
       have1(Y,employee1#'Peter Piper',X)))                (C2)
\end{verbatim}
and after translation,
\begin{verbatim}
exists([X],
   employee(organization1#sri,employee1#'Peter Piper',X,y) (C3)
\end{verbatim}
Translating {\tt (C2)} into the canonical form {\tt (C3)}
makes it simple to ascertain that the
tuple has so far not been completely described, by comparing {\tt (C3)} and
the original tuple {\tt (C1)}.
 On the next cycle,
the compiled consequence goal
\begin{quote}\begin{verbatim}
man1(employee1#'Peter Piper')
\end{verbatim}\end{quote}
is used to generate the sentence
\begin{quote}
{\it Peter Piper (the employee) is a man.}
\end{quote}
The conjunction of {\tt (C3)} and the propositional content of the second
sentence now however translates to {\tt (C1)}, so the filter can return
with the information that the tuple has been completely described.

For tuples whose structure is more complex, the filter leads to a
very substantial reduction in the verbosity of the output produced
by the description module. For example, when generating from a
{\tt project} tuple the output produced is something like the following
\begin{verbatim}
  CLARE (the project)'s end date is 19/11/1992 (the day).
  CLARE (the project)'s number is 8468.
  CLARE (the project)'s start date is 20/11/1989 (the day).
\end{verbatim}
With the utterance filter disabled, the output becomes
\begin{verbatim}
  CLARE (the project)'s end date is 19/11/1992 (the day).
  CLARE (the project)'s number is 8468.
  CLARE (the project)'s start date is 20/11/1989 (the day).
  CLARE (the project)'s account is account 8468.
  CLARE (the project)'s project number is 8468.
  CLARE (the project)'s account number is 8468.
  19/11/1992 (the day) is an end date.
  8468 is a number.
  20/11/1989 (the day) is a start date.
  account 8468 is an account.
  CLARE (the project) is a project.
  8468 is a project number.
  8468 is an account number.
  CLARE (the project) began on 20/11/1989 (the day).
  CLARE (the project) ended on 19/11/1992 (the day).
  CLARE (the project) finished on 19/11/1992 (the day).
  CLARE (the project) started on 20/11/1989 (the day).
  CLARE (the project) began.
  CLARE (the project) ended.
  CLARE (the project) finished.
  CLARE (the project) started.
\end{verbatim}
As can been seen, the filter's simple uniform mechanism manages to eliminate
several different kinds of excess verbosity, some of which are not
obviously redundant and require non-trivial inference to detect.

\section{Generating questions}
\label{Generating-Questions}

It is also possible, using the methods we have just described, to
generate questions. The module in CLARE responsible for doing this is
invoked by the top-level CLARE command {\tt .ask}: if some incomplete
tuples have been created by processing assertions (see
sections~\ref{Declaration-functionality}
and~\ref{Simplify-assertions}), CLARE will respond to {\tt .ask} by
entering a loop where it poses questions to the user intended to
elucidate the information necessary to fill the unknown argument
positions.  This can involve generation of both Y-N and WH- questions.
We begin by giving an example dialogue: utterances by CLARE and the
user are prefixed by \verb!clare>! and \verb!user>! respectively, and
comments are in normal font.
\begin{verbatim}
user>  ARLA is a project.
\end{verbatim}
The user inputs a declarative statement. After processing it, CLARE has
an incomplete {\tt project} record.
\begin{verbatim}
user>  .ask
\end{verbatim}
The user invokes the question-asking module to fill the unspecified fields
in the record.
\begin{verbatim}
clare> what is ARLA (the project)'s account?
\end{verbatim}
CLARE asks a question to find out the filler of the
``account'' field in the record, and hands control back to the user to get the
answer.
\begin{verbatim}
user>  1234.
\end{verbatim}
The user answers. The utterance is processed by CLARE in the normal way,
which involves syntactic and semantic analysis, followed by ellipsis resolution
in the context of the previous question.
\begin{verbatim}
clare> account 1234 is ARLA (the project)'s account.
\end{verbatim}
CLARE prints out a paraphrase showing the result of performing
resolution, to keep the user informed. It then translates the result
in the usual way, merging it with the previous one as described in
section~\ref{Simplify-assertions}.
\begin{verbatim}
clare> what is ARLA (the project)'s start date?
\end{verbatim}
CLARE asks another question, this time to find out the project's start date.
Processing is as before.
\begin{verbatim}
user>  1/4/93.
clare> 1/4/1993 (the day) is ARLA (the project)'s start date.
clare> what is ARLA (the project)'s end date?
\end{verbatim}
Another question from CLARE; this time, the user replies with a sentence.
Processing is still the same as usual.
\begin{verbatim}
user>  ARLA will end on 1/4/96.
\end{verbatim}
When this sentence has been processed, CLARE has successfully
filled all the fields in the tuple.

The range of questions currently handled is limited to Y-N questions
whose propositional content is an existentially quantified
conjunction, and WH-questions whose propositional content is a
lambda-bound existentially quantified conjunction. For questions of
these types, we can acquire case-based rules by a slight extension of
the methods described in section~\ref{Case-based-learning}; the only
extra point is that it is necessary to store the lambda-bound variable
for a WH-question. Thus for example the rule induced from the training
sentence
\begin{quote}
{\it Who loves Mary?}
\end{quote}
is
\begin{verbatim}
trl_gen([whq,
         form(verb(pres,no,no,no,y),
              A,
              B^[B,[C,A,term(q(tpc,wh,_),_,
                             D^[personal,D],_,_),E]],_)],
        F^exists([G],
             and([personal,F],
                 [C,G,F,H]))),
        [entity_ref_term(H,E),
         personal(I),
         trlgoal([C,_,I,H],
                 genpred(C,(form(_,verb(_,_,_,_,_),
                                 J,K^[K,[C,J,_,_]],_),
                                 v(_,_))))],
        [I])
\end{verbatim}
where the fourth argument {\tt [I]} to {\tt trl\_gen} is the list of
lambda-bound variables in the conditions; note that the second argument,
the propositional content, is already a lambda-bound form. We now describe
how rules of this kind can be used.

Let us begin by supposing that we have a conceptual goal $C$, in which
some argument $X$ is so far uninstantiated. There are two obvious
strategies to use when trying to find a filler for $X$. The first is
to attempt to frame a WH-question whose propositional content is equivalent
to $\lambda X.C$; the second is to substitute some plausible value $a$ for
$X$ in $C$, and then try to find a Y-N question whose content is equivalent to
$C[X/a]$. CLARE can use either strategy, choosing according to the nature
of the argument position. If it is an argument that has to be filled by
one of a known finite set of possible code values, it tries the second
strategy; otherwise, it uses the first one.

For both strategies, the major difference compared to generating assertions is
that
what is required is an utterance whose propositional content is {\it
equivalent} to that
of the question, not {\it implied} by it. We will
refer to the initial expression, either $\lambda X.C$ or $C[X/a]$ as the case
may be, as the {\it abstract question}; we can reduce the two cases to one
by considering abstract questions as forms bound by a set of lambdas, where the
set is empty for Y-N questions and non-empty for WH-questions.



Suppose then that the abstract question is $\lambda \vec{X}.C$ where
$\vec{X}$ is a vector of variables. Generation proceeds as follows. We
try to find a description rule $R$, generalized from a question, for which
the set of lambda-bound variables is $\vec{Y}$, the conditions are
$Conds$, the propositional content is $\lambda \vec{Y}.P_R$,
and the following hold:
\begin{enumerate}
\item The cardinalities of $\vec{X}$ and $\vec{Y}$ are equal.
\item If $\vec{a}$ is a set of unique constants with the same cardinality as
$\vec{X}$ and $\vec{Y}$, then $Conds[\vec{Y}/\vec{a}]$ is implied by
$C[\vec{X}/\vec{a}]$.
\item $P_R[\vec{Y}/\vec{a}]$ implies $C[\vec{X}/\vec{a}]$
\end{enumerate}
If these conditions are met, then the propositional content of $R$ is
equivalent
to that of $\lambda \vec{X}.C$. The conditions are phased the way they are so
as to
allow use of the techniques presented in section~\ref{Talk-about}; thus the
second condition is satisfied using the compiled TRL lemmas, and the third one
with essentially the same mechanism as that used by
the ``utterance filter'' described at the end
of the last section.


\chapter{Conclusions}
\label{conclusions}

\section{Summary}

We will now pull the threads together and summarize the main arguments
of the thesis. The fundamental idea has been to provide a formal
account of how natural language utterances can be related to the
contents of databases. We have reduced this to a logical problem by
assuming that both databases and natural language utterances can be
expressed in logical terms. The databases present no problem; to deal
with the language, we have assumed the existence of a ``natural
language engine'', which can mediate between surface linguistic
expressions and their representations as ``literal logical forms''.
The problem now became one of converting between two types of logical
formulas, one relating to language and one relating to databases. In
Chapter~\ref{translation}, we showed that several different types of
interface functionality can be reduced to the same common form: we are
given a formula $F_{source}$, a background theory $\Gamma$, and
criteria which permit the making of certain assumptions. We are asked
to find a formula $F_{target}$, fulfilling certain restrictions, and
permissible assumptions $A$, such that
\begin{eqnarray}
\Gamma\cup A \Rightarrow (F_{ling} \equiv F_{db})\label{Basic-equiv-conc}
\end{eqnarray}
One important technical device used was to give the database predicates
a ``trivial''semantics, referring only to the existence of database
tuples and not to objects in the exterior world. In
Section~\ref{Closed-World}, we showed how this makes it possible to
side-step in a simple way the difficulties associated with use of
the Closed World Assumption, in effect allowing the CWA to be
``relativized'' by the domain theory $\Gamma$.

The next two chapters addressed two interlinked questions:
\begin{enumerate}
\item Can we construct suitable
theories $\Gamma$, that make it possible to solve inference problems
of type~(\ref{Basic-equiv-conc}) with reasonable efficiency?
\item If so, how are these inference problems solved?
\end{enumerate}
Answering the two questions just posed in reverse order, we first
described in Chapter~\ref{AET} an inference process, Abductive
Equivalential Translation (AET), which can be used in conjunction with
a domain theory consisting of directed conditional equivalences and
Horn-clauses. AET is sound but not complete; its strength is that it
has an efficient and easily comprehensible precedural model, which
allows equivalential theories to be constructed and debugged in much
the same way as Prolog programs. A few initial examples were
presented, showing in particular how AET is powerful enough to provide
an elegant and natural solution to the well-known ``Doctor on Board''
problem. The AET process relies on having an extended Horn-clause
inference engine, which was described in Chapter~\ref{reasoning}.

In Chapter~\ref{dmi}, we then went on to show how equivalential theories
of the type described in the previous chapter could be constructed to
solve the original problem of relating linguistic and database
predicates. We described an example theory, which relates a fairly
wide range of linguistic constructs to the database relations of a
real projects and payments database. The key idea was to structure the
theory in a modular way that allows the linguistic predicates to be
translated into forms successively ``nearer'' to the database
relations. In particular, we argued that it is useful to funnel
translation through so-called ``conceptual'' predicates; each database
predicate $D$ corresponds to one conceptual predicate, which roughly
contains the union of the information in $D$ and the lingustic
predicates that translate into it. We also show how introduction of
conceptual predicates makes it possible for translation to deal with
contingent completeness of database information, in particular the
common constraint that records are only available for a limited
period.

Nearly all the previous discussion focussed on translation of
linguistic predicates into database predicates. Chapter~\ref{genass}
considered the reverse problem of deriving language from database.  A
simple method was introduced, that extracted ``templates'' from
typical examples and could use them to generate both descriptions of
complete database records and questions asking for values of unfilled
fields in incomplete ones. The method used AET to provide a ``filter''
which limits the verbosity of descriptions of complete tuples, and
checks the validity of proposed questions.

\section{Further directions}

\subsection{Using AET in other domains}

This thesis has reported in detail how a non-trivial natural-language
interface was successfully implemented using AET. Every effort was
made to employ methods that would be as general as possible,
but the fact remains that one swallow does not make a summer: AET
needs to be used more before its merits can be properly evaluated. It
is thus pleasant to be able to say that a second AET application, a
CLARE interface to the AUTOROUTE route-finding package, has now been
constructed. This is being reported elsewhere (Lewin {\it et al}
1993). It is particularly gratifying that nearly all the work was
carried out by project personnel (primarily Ian Lewin) who had not
participated in the development of the PRM application. Two swallows
do not make a summer either, but one can at any rate say that the
swallow population seems to be increasing.

\subsection{Making the AET process symmetrical}

One unfortunate shortcoming of the work reported here is the
lack of symmetry between the roles played by equivalences in the
analysis and generation processes. Since equivalences are by
their nature symmetrical, it seems quite reasonable to expect
that the part of the generation process which is concerned
with inference could be formulated more directly in terms of AET.

In this context, I would like to make it clear that the treatment of
generation described in Chapter~\ref{genass} should not be regarded as
more than a stopgap; in this section, I will sketch how a more
satisfying solution could be achieved. Recall from
Section~\ref{LDT-and-effective-trans} the key notion of ``effective
translation'', which was used to formalize the analysis tasks. Here,
we were given a linguistic formula $F_{source}$: then the condition
for $F_{target}$ to be an {\it effective translation} of $F_{source}$
was that there existed a set of abductive assumptions $A$ such that
\begin{eqnarray}
\Gamma\cup A \Rightarrow (F_{source} \equiv F_{target})
\end{eqnarray}
and
\begin{itemize}
\item Each assumption in $A$ is acceptable in the context in which it is made.
\item There is a finite proof procedure for determining the truth of
$F_{target}$.
\end{itemize}
The symmetrical concept in the generation direction could be called a
``lingustically realizable translation''. So this time we are given an
$F_{source}$ which is a formula expressed in terms of database
predicates: then $F_{target}$ is a {\it linguistically realizable
translation} of $F_{source}$ if there is a set of abductive
assumptions $A$ such that
\begin{eqnarray}
\Gamma\cup A \Rightarrow (F_{source} \equiv F_{target})
\end{eqnarray}
and
\begin{itemize}
\item Each assumption in $A$ is acceptable in the context in which it is made.
\item There is a linguistic utterance whose ``literal'' logical
denotation is $F_{target}$.
\end{itemize}
The AET process would be used in the database to language direction to derive
possible candidates for $F_{target}$; these would be analyzed further
to discover if they could be realized as denotations of utterances.

It has been possible, using AET in the language to database direction,
to make the translation process deterministic: this is justifiable on
the grounds that database representation is a canonical form. The main
technical problem involved in implementing the proposed scheme is that
the AET process in the database to language direction would have to be
non-deterministic, since linguistic representations are not canonical.
For this reason, we did not attempt to implement reverse AET in the
CLARE project. It seems to me, however, that the obstacles posed are
by no means insuperable, and that it would be extremely interesting to
investigate the ``reverse translation'' idea more thoroughly.

\subsection{Goal-directed reasoning about acquisition of knowledge}

It is worth noting that the formalization of ``knowing wh...''
described in Section~\ref{Meta-knowledge-functionality} can easily be
extended into a formalization of ``finding out wh...''. This can be
done in several ways: the simplest is to allow the conditions on
certain equivalences to be contingent on executable assumptions,
assumptions which will be true if appropriate perceptual actions are
carried out. We illustrate with a sketch of the standard example,
``finding out someone's telephone number'' (cf~e.g.~(McCarthy 1977),
(Moore 1985)).

Simplifying as much as possible (and side-stepping some of the
temporal reasoning issues without comment), we start by letting
$\Gamma$ be the background theory as usual. We assume we have an
``open'' predicate $$phone\_no(Name,Number,T)$$ with the intended
semantics ``the phone number of the person called $Name$ is $Number$
at time $T$''.  We also have a ``closed'' predicate,
$$REC\_PHONE\_NO(Name,Number,T)$$ whose semantics are ``the phone
number of the person called $Name$ is $Number$ at time $T$, and this
is recorded in my database at time $T$''. Now suppose we want to
reason about how we might find out Mary's phone number, in other
words arrange for it to be present in the database. We formalize
this as requiring a formula $F_{db}$ expressed in terms of the database
predicate $REC\_PHONE\_NO$, a time $T_{done}$, and a set of
assumptions $A$ corresponding to actions that I can carry out,
such that
\begin{eqnarray}
\Gamma \cup \{A, finish\_time(T_{done})\} \Rightarrow \nonumber \\
((\exists N.phone\_no(mary^\prime,N,T_{done})) \equiv F_{db})
\label{Find-out-conditions}
\end{eqnarray}
where $T_{done}$ represents a time at which the information has been
added to the database. Including the goal $finish\_time(T_{done})$ on
the left-hand side is convenient for reasons that will shortly become
apparent.

The following formula expresses the fact that the phone number of the
person called $Name$ is recorded in my database after I have executed
the action of looking up $Name$ in the telephone directory
\begin{eqnarray}
&execute(directory\_look\_up(Name),T_1) \wedge before^\prime(T_1,T_2)
\rightarrow \nonumber \\
&(phone\_no(Name,N,T_2) \equiv REC\_PHONE\_NO(Name,N,T_2))
\end{eqnarray}
Lastly, we have to specify assumability conditions. We allow goals of
the form $$execute(Action,Time)$$ to be assumed if $Time$ is in the
future and $Action$ is a ground action term. We also allow goals of
the form $$before^\prime(T_1,T_2)$$ to be assumed if $T_1$ is the time
of an executed action, and $$finish\_time(T_2)$$ holds, on the grounds
that we are at liberty to choose a finish time that occurs after all
the necessary actions have been completed.

By performing normal AET, it follows (we omit the details) that a
solution to~\ref{Find-out-conditions} exists, where $F_{db}$ is
$$\exists N_1.REC\_PHONE\_NO(mary^\prime,N_1,T_{done})$$ and $A$ (the
set of assumptions) is, for some $T_{lookup}$
$$\{execute(look\_up\_no(mary^\prime),T_{lookup}),
before^\prime(T_{lookup},T_{done})\}$$ In other words, I will know
what Mary's number is at any moment after the time when I perform the
operation of looking it up. The significant thing about this example,
compared e.g.\ to the account in (Moore 1985), is that the process of
reasoning about knowledge acquisition is carried out in a
goal-directed way that is compatible with normal AI planning methods.
I think that this is an interesting idea to explore further.


\chapter*{References}
\addcontentsline{toc}
{chapter}{References}

\newenvironment{reverseindent}%
{\begin{list}{}{\setlength{\labelsep}{0in}
	        \setlength{\labelwidth}{0in}
	        \setlength{\itemindent}{-\leftmargin}}}%
{\end{list}}

\begin{reverseindent}

\item\pagebreak[3]
Alshawi, H. 1990. ``Resolving Quasi Logical Forms''.
{\it Computational Linguistics} 16:133--144.

\item\pagebreak[3]
Alshawi, H., ed. 1992. {\it The Core Language
Engine}. Cambridge, Massachusetts: The MIT Press.

\item\pagebreak[3]
Alshawi,~H., D.M.~Carter, M.~Rayner and B.~Gamb\"ack. 1991.
``Translation by Quasi Logical Form Transfer'', in {\it Proc.
29th ACL}, Berkeley, California.

\item\pagebreak[3]
Alshawi,~H. and R.~Crouch. 1992.  ``Monotonic Semantic
Interpretation'', {\it Proc. 30th ACL}, Newark, Delaware.

\item\pagebreak[3]
Alshawi, H., Rayner, M., and Smith, A.G. 1991. ``Declarative Derivation of
Database Queries from Meaning Representations'', in Society for Worldwide
Interbank Financial Telecommunications S.C. (ed.)  {\it Proceedings of the
1991 BANKAI Workshop on Intelligent Information Access}, Elsevier,
Amsterdam.

\item\pagebreak[3]
Appelt,~D. and J.R.~Hobbs. 1990.
``Making Abduction more Efficient'', in {\it Proc. DARPA
Workshop on Speech and Natural Language}, Morgan Kaufmann,
San Mateo.

\item\pagebreak[3]
Bennett,~M. 1979. {\it Questions in Montague Grammar}. University of
Indiana Linguistics Club.

\item\pagebreak[3]
Bronneberg, W.J.H.J., H.C. Bunt, S.P.J. Landsbergen, R.J.H. Scha,
W.J. Schoenmakers and E.P.C. van Utteren. 1980. ``The Question Answering
System PHLIQA1''. In L. Bolc (ed.), {\it Natural Language Question
Answering Systems}. Macmillan.

\item\pagebreak[3]
Charniak, E. and Goldman, R. 1988. ``A Logic for Semantic
Interpretation''. {\it Proceedings of the 26th Meeting of the
Association for Computational Linguistics}, 87--94.

\item\pagebreak[3]
Clark, K.L.  1978.  ``Negation as Failure''.  In H.  Gallaire and J.
Minker (editors), {\it Logic and Data Bases}, Plenum, New York.

\item\pagebreak[3]
Dalrymple, M., S.~M.~Shieber, and F.~C.~N.~Pereira. 1991.
``Ellipsis and Higher-Order Unification''. {\it Linguistics and Philosophy},
14:399--452.

\item\pagebreak[3]
Davidson, D. 1967.
``The Logical Form of Action Sentences''.
Reprinted in Davidson, 1980, {\it Essays on Actions and Events},
Oxford University Press.

\item\pagebreak[3]
Grosz, B.~J., D.~E.~Appelt, P.~Martin, and F.~Pereira. 1987. ``TEAM:  An
Experiment in the Design of Transportable Natural-Language Interfaces''. {\it
Artificial Intelligence} 32: 173--243.

\item\pagebreak[3]
Hobbs,~J.R., M.~Stickel, P.~Martin and D.~Edwards. 1988.
``Interpretation as Abduction'', in {\it Proceedings of the 26th ACL},
Buffalo, New York.

\item\pagebreak[3]
Hobbs, J.~R., and S.~M.~Shieber. 1987. ``An Algorithm for Generating
Quantifier Scopings''. {\it Computational Linguistics} 13:47--63.

\item\pagebreak[3]
Kartunnen,~L. 1977. ``Syntax and Semantics of Questions''.
{\it Linguistics and Philosophy} 1:3--44.

\item\pagebreak[3]
Konolige, K. 1981. {\it The Database as Model: A Metatheoretic
Approach}, SRI technical note 255.

\item\pagebreak[3]
Korf, R. 1986. ``Search: A Survey of Recent Results'' in
{\it Exploring Artificial Intelligence: Survey Talks
from the National Conferences on Artificial Intelligence}, Morgan
Kaufmann, San Mateo.

\item\pagebreak[3]
Lewin,~I., M.~Russell, D.M.~Carter, S.~Browning, K.~Ponting and
S.G.~Pulman. 1993. ``A speech-based route enquiry system built from
general-purpose compoments''. {\it Proceedings of EUROSPEECH '93}.

\item\pagebreak[3]
McCarthy,~J. 1977. ``Epistemological Problems of Artificial
Intelligence''. {\it Proceedings of the 4th IJCAI}, Cambridge, MA.

\item\pagebreak[3]
McCord,~M.C. 1987. ``Natural Language Processing in Prolog'', in
A. Walker (ed.) {\it Knowledge Systems and Prolog}. Addison-Wesley,
Reading, MA.

\item\pagebreak[3]
Moore,~R.C. 1985. ``A Logic of Knowledge and Action''. In Hobbs,~J.R.
and Moore,~R.C., {\it Formal Theories of the Common-Sense World},
Ablex.

\item\pagebreak[3]
Moran, D.~B. 1988. ``Quantifier Scoping in the SRI Core Language Engine''.
Proceedings of the 26th Annual Meeting of the Association for Computational
Linguistics, State University of New York at Buffalo, Buffalo,
New York, 33--40.

\item\pagebreak[3]
Pereira, F.~C.~N. 1990.
``Categorial Semantics and Scoping'', {\it Computational Linguistics}
16:1 1--10.

\item\pagebreak[3]
Pereira,~F.C.N. and M.E.~Pollack. 1991. ``Incremental Interpretation'',
{\it Artificial Intelligence} {\bf 50} 37-82.

\item\pagebreak[3]
Pereira,~F.C.N. and S.M.~Shieber. 1985. {\it Prolog and Natural-Language
Understanding}, CLSI Lecture Notes.

\item\pagebreak[3]
Perrault, C. Raymond and Barbara J. Grosz. 1988. ``Natural Language
Interfaces'' in {\it Exploring Artificial Intelligence: Survey Talks
from the National Conferences on Artificial Intelligence}, Morgan
Kaufmann, San Mateo.

\item\pagebreak[3]
Rayner,~M. and S.~Janson. 1989. ``Finding Out = Achieving
Decidability'', in {\it Working notes: IJCAI workshop on
Knowledge, Planning and Perception}, Detroit, Michigan.
Also available as Research
Report R89015, Swedish Institute of Computer Science, Kista, Sweden.

\item\pagebreak[3]
Scha,~R.J.H. 1983. {\it Logical Foundations for Question Answering},
Ph.D. Thesis, University of Groningen, the Netherlands.

\item\pagebreak[3]
Shieber, S.~M., G.~van~Noord, F.~C.~N.~Pereira, and R.~C.~Moore. 1990.
``Semantic-Head-Driven Generation''.
{\it Computational Linguistics} 16:30--43.

\item\pagebreak[3]
Stallard,~D.G. 1986. ``A Terminological Simplification Transformation
for Natural Language Question-Answering Systems''. {\it Proceedings of the
24th Annual Meeting of the Association for Computational Linguistics}, ACL,
241--246.

\item\pagebreak[3]
Stickel, M. E. 1986. ``A Prolog technology theorem prover: implementation by an
extended Prolog compiler.'', {\it Journal of Automated Reasoning}, 4,
353-380.

\item\pagebreak[3]
Sterling,~L. and S.~Shapiro. 1985. {\it The Art of Prolog}, Addison-Wesley,
Reading, MA.

\item\pagebreak[3]
Williams,~B.C.. 1991. ``A Theory of Interactions: Unifying Qualitative
and Quantitative Algebraic Reasoning''. {\it Artificial Intelligence}
51: 39--94.

\end{reverseindent}

\appendix

\chapter{The semantics of questions}
\label{qsem}

This appendix summarizes the theory of question semantics used
throughout the thesis. There are two main points. First, we
distinguish between {\it questions} and {\it question utterances};
this is not an idea that most linguists will find controversial,
though non-linguists will perhaps feel it needs some expanation.
Briefly, then, {\it questions} are a type of linguistic construction
that can occur either alone or {\it embedded} in other constructions.
The only question-embedding construction that is considered in any
detail in the main body of the thesis is ``know {\it Q}'', with
{\it Q} a question (see Section~\ref{Meta-knowledge-functionality});
typical examples, with the embedded questions italicized, might be
\begin{description}
\item[(Q1)] Do you know {\it which employees work on each project}?
\item[(Q2)] Do you know {\it whether any payments were made after
1/1/92?}
\end{description}
The embedded question in {\bf (Q1)} is considered to be {\it Which
employees work on each project?}, and in {\bf (Q2)} {\it Was any
payment made after 1/1/92?}.

When questions stand on their own, we call them {\it question
utterances}. A question utterance is considered to be a command to
answer the question in some appropriate way. Now when assigning
semantics to questions, we will require that embedded questions and
question utterances be treated uniformly. A given question, $Q$, will
have some semantic value, $Q^\prime$. We thus require that the
semantic value of $Q$ used as a question-utterance be a function of
$Q^\prime$, and also that the semantic value of a construction
embedding $Q$ be a function of $Q^\prime$. For our present purposes,
we can be even more specific: we demand that
\begin{enumerate}
\item The semantics of constructions of type ``X knows $Q$''
should be schematically $$know^\prime(x^\prime,Q^\prime)$$
In practice, we will only be interested in what the system knows,
and we will abbreviate $$know^\prime(system^\prime,Q^\prime)$$ to
$$kw(Q^\prime)$$
\item The semantics of $Q$ used as a question utterance should
be schematically $$question\_answered(Q^\prime)$$
This is interpreted as a goal that will be true if an action is
carried out appropriate for answering $Q$.
\end{enumerate}
We in discuss appropriate semantics for the operators
$question\_answered$ and $kw$ in
Sections~\ref{YNQ-functionality},~\ref{Command-functionality}
and~\ref{Meta-knowledge-functionality}.

The remaining point to consider is how to assign appropriate semantics
to the questions themselves. Here, we will adopt a minimal proposal,
which as far as we know was originally suggested by Bennett (1979); we
try to abstract away as much as possible from the details of any
specific semantic theory. The proposal is as follows:
\begin{enumerate}
\item If $Q$ is a Y-N question
corresponding to a declarative sentence $D$ (i.e. $Q$ is the question
that asks whether or not $D$ is true), then we take its semantic value
to be that same as that of $D$.
\item If $Q$ is a WH-question, which can be read as asking
which objects stand in a given relationship $R$, then the semantic
value of $Q$ is $R$.
\end{enumerate}
The Y-N question case should be clear, but the WH-question case needs
some further explanation. Usually, there will be only one WH-element,
and the question can be read as asking for objects with a certain
property. Thus for example consider the questions {\bf (Q3)}--{\bf (Q5)}:
\begin{description}
\item[(Q3)] Who worked on the CLARE project?
\item[(Q4)] When did the CLARE project finish?
\item[(Q5)] Which projects started during 1990?
\end{description}
Here, the semantic value of {\bf (Q3)} is a property that holds of
people who worked on the CLARE project; the semantic value of
{\bf (Q4)} is a property that holds of dates when the CLARE
project finished; and the semantic value of {\bf (Q5)} is a
property holding of projects which started during 1990. The
semantic value of a property can most simply be represented
as a $\lambda$-abstraction: if we allow this, the semantic values
of {\bf (Q3)}--{\bf (Q5)} are roughly those shown
in~(\ref{q3-sem})--(\ref{q5-sem}):
\begin{eqnarray}
&\lambda X.person^\prime(X)\wedge work\_on^\prime(X,clare)
\label{q3-sem} \\
\nonumber \\
&\lambda D.\exists E.date(D)\wedge finish^\prime(E,clare)\wedge date\_of(E,D)
\label{q4-sem} \\
\nonumber \\
&\lambda P.\exists E.project(P)\wedge start^\prime(E,P)\wedge
during^\prime(E,1990)
\label{q5-sem}
\end{eqnarray}
If there are (either explicitly or implicitly) several question-elements,
then the semantic value of the question becomes a relation, as
in {\bf (Q6)} and {\bf (Q7}):
\begin{description}
\item[(Q6)] Who works on which project?
\item[(Q7)] When did each project start?
\end{description}
which receive the semantic values shown in~(\ref{q6-sem}) and~(\ref{q7-sem}):
\begin{eqnarray}
&\lambda X\lambda Y.person^\prime(X)\wedge project^\prime(Y)\wedge
work\_on^\prime(X,Y)
\label{q6-sem} \\
\nonumber \\
&\lambda P\lambda D.\exists E.project(P)\wedge start^\prime(E,P)\wedge
date\_of(E,D)
\label{q7-sem}
\end{eqnarray}
It should be noted that the theory we have outlined here is not the
most popular one in theoretical linguistic circles. Currently, the
``standard'' approach is that described in (Kartunnen 1977), which
makes the denotation of a question the set of true answers.  We feel
that there are good reasons to query the appropriateness of this idea,
but a discussion of the issues lies too far outside the scope of the
thesis; the interested reader is referred to (Rayner and Janson 1987)
for our objections to Kartunnen's approach.

\chapter{Syntax and semantics of TRL}\label{trlsem}

This appendix briefly defines the syntax and semantics of the TRL
language. TRL is a conservatively extended first-order logic, which is
used as the level of representation in CLARE that supports inference
operations. When discussing AET at an abstract level, we consequently
often find it convenient to use normal logical notation. We indicate
for each construct the ``normal'' logical equivalent.

There are three basic types of construct: terms, abstracts and forms.
We list each one seperately.

{\bf NB:} The version of TRL described here is that used in the
body of the thesis, and represents a slightly rationalized form
of the implemented version. The most important differences concern
the concrete syntax for typed terms (here, \verb!Type#Id!), and
higher-order constructs for counting, summing and ordering.

\section{Terms}

A TRL term can be any of the following:
\begin{enumerate}
\item A variable, represented as a Prolog variable.
\item A constant, represented as a Prolog atom.
\item The result of applying a function symbol to a list of terms,
represented as a complex Prolog term.
\end{enumerate}
When using standard logical notation, we do not enforce any syntactic
convention to distinguish between variables and constants, relying
on this being clear from context.

We will distinguish terms of the \verb!Type#Id!, where {\tt Type} is a
TRL constant and {\tt Id} is a TRL term. These will have the
conventional meaning ``the object of type {\tt Type}, whose database
identifier is {\tt Id}''. In normal logical syntax we will write these
as $Type:Id$.

\section{Abstracts}

If {\tt P} is a TRL form or a TRL abstract, and {\tt X1}... {\tt Xn}
are TRL variables, then
\begin{quote}
\begin{verbatim}
X1^X2...Xn^P
\end{verbatim}
\end{quote}
is a TRL abstract. It denotes a {\tt n}-ary relation $R$ between
objects, such that $R$ holds between the objects denoted by {\tt
A1}... {\tt An} iff the formula resulting from substituting {\tt Ai}
for {\tt Xi} in {\tt P} holds.  In normal logical notation, we write
$\lambda X_1...\lambda X_n.P$.

\section{Forms}

A TRL form can be any of the following:
\begin{description}
\item[Conjunction] If {\tt P} and {\tt Q} are TRL forms, then
\begin{quote}
\begin{verbatim}
and(P,Q)
\end{verbatim}
\end{quote}
is a TRL form, which holds iff both {\tt P} and {\tt Q} hold.
Conjunctions are as usual written $P\wedge Q$ in normal logical
notation.
\item[Disjunction] If {\tt P} and {\tt Q} are TRL forms, then
\begin{quote}
\begin{verbatim}
or(P,Q)
\end{verbatim}
\end{quote}
is a TRL form, which holds iff either {\tt P} or {\tt Q} holds.
Disjunctions are as usual written $P\vee Q$ in normal logical
notation.
\item[Negation] If {\tt P} is a TRL form, then
\begin{quote}
\begin{verbatim}
not(P)
\end{verbatim}
\end{quote}
is a TRL form, which holds iff {\tt P} does not hold.
Negations are written $\neg P$ in normal logical
notation.
\item[Implication] If {\tt P} and {\tt Q} are TRL forms, then
\begin{quote}
\begin{verbatim}
impl(P,Q)

Q <- P
\end{verbatim}
\end{quote}
are alternate ways of writing the same TRL form, which holds iff
either {\tt Q} holds or {\tt P} does not hold.  In normal notation, we will
permit both $P \rightarrow Q$ and $Q \leftarrow P$.
\item[Equivalence] If {\tt P} and {\tt Q} are TRL forms, then
\begin{quote}
\begin{verbatim}
P <-> Q
\end{verbatim}
\end{quote}
is a TRL form, which holds iff {\tt P} and {\tt Q} either both hold or
both fail to hold.  In normal notation, we write $P \equiv Q$.
\item[Existential quantification] If {\tt P} is a TRL form, and
{\tt [X1, X2,...]} is a list of TRL variables, then
\begin{quote}
\begin{verbatim}
exists([X1, X2,...],P)
\end{verbatim}
\end{quote}
is a TRL form, which holds iff {\tt P} hold for some instantiation
of the variables {\tt [X1, X2,...]}. In normal notation, we write
$\exists X_1, X_2,... .P$.
\item[Universal quantification] If {\tt P} is a TRL form, and
{\tt [X1, X2,...]} is a list of TRL variables, then
\begin{quote}
\begin{verbatim}
forall([X1, X2,...],P)
\end{verbatim}
\end{quote}
is a TRL form, which holds iff {\tt P} hold for all instantiations
of the variables {\tt [X1, X2,...]}. In normal notation, we write
$\forall X_1, X_2,... .P$.
\item[Counting] If \verb!X^P! is a TRL abstract, and {\tt N} is a TRL term,
then
\begin{quote}
\begin{verbatim}
cardinality(N,X^P)
\end{verbatim}
\end{quote}
is a TRL form, which holds iff there are precisely {\tt N}
instantiations of {\tt X} such that {\tt P} holds. In normal notation,
we write $count(N,\lambda X.P)$.
\item[Summing] If \verb!X^P! is a TRL abstract, and {\tt N} is a TRL term, then
\begin{quote}
\begin{verbatim}
sum(Sum,X^P)
\end{verbatim}
\end{quote}
is a TRL form, which holds iff
\begin{enumerate}
\item all objects {\tt A} such {\tt P} holds
when {\tt A} is substituted for {\tt X} are summable quantities
\item {\tt Sum} is their sum.
\end{enumerate}
In normal notation, we write $sum(Sum,\lambda X.P)$.
\item[Ordering] If \verb!X^D^P! is a TRL abstract, and {\tt N} and {\tt
Ordering} are TRL terms, then
\begin{quote}
\begin{verbatim}
order(Selected,X^D^P,Ordering)
\end{verbatim}
\end{quote}
is a TRL form, which holds iff
\begin{enumerate}
\item {\tt Ordering} is a term representing an ordering relation
\item {\tt DMax} is the maximal {\tt D1} under the relation represented
by {\tt Ordering} such that P when {\tt D1} is substituted for {\tt D}
and some {\tt X1} is substituted for {\tt X}, and
\item {\tt Selected} is such an {\tt X1}
\end{enumerate}
In normal notation, we write
$order(Selected,\lambda X\lambda D.P(X,D),Ordering)$.
\item[Meta-knowledge] If {\tt P} is a TRL form, then
\begin{quote}
\begin{verbatim}
kw(P)
\end{verbatim}
\end{quote}
(read ``knows-whether'' {\tt P}) is a TRL form, which holds iff there
is an effective translation of {\tt P} (cf.
Section~\ref{Effective-translation}). In normal notation we
write $kw(P)$.
\item[Necessity] If {\tt P} is a TRL form, then
\begin{quote}
\begin{verbatim}
def(P)
\end{verbatim}
\end{quote}
(read ``{\tt P} is true by definition'') is a TRL form, which holds iff
{\tt P} is implied by the background theory {\it without} the database.
\end{description}

\chapter{Translating to SQL}\label{SQL-convert}

This section briefly describes the module responsible for synthesis of
actual SQL queries. The conversion module takes as input a TRL form
$P$, assumed to represent a question; it outputs a form $P^\prime$,
such that $P$ and $P^\prime$ are equivalent and as much as possible of
$P$ has been replaced by calls to the SQL interface. The interface is
mediated through the predicate
\begin{quote}\begin{verbatim}
sql_select(Vars,SQLQuery)
\end{verbatim}\end{quote}
where {\tt SQLQuery} is a term representing an SQL {\tt SELECT} statement, and
{\tt Vars} is a list whose length is equal to the number of selected
variables in {\tt SQLQuery}. The intended semantics are that
\begin{quote}\begin{verbatim}
sql_select(Vars,SQLQuery)
\end{verbatim}\end{quote}
holds iff {\tt Vars} are a row selected by {\tt SQLQuery}.  {\tt
SQLQuery} is a logical term representing the abstract SQL syntax;
conversion to concrete SQL syntax is handled by the low-level SQL
interface, and is straightforward.  We now describe how the conversion
process is carried out; we will illustrate by continuing the extended
example from section~\ref{Examples}.  We use the same abbreviated
notation, shortening {\tt exists} to {\tt E} and {\tt forall} to {\tt
A}.  Recall that the original query was
\begin{description}
\item[(S1)] Show all payments made to BT during 1990.
\end{description}
This receives the final translated TRL representation (slightly
simplified in the interests of readability),
\begin{verbatim}
A([TransId,Date,DBDate,AmtNum],
  impl(and(TRANS(TransId,DBDate,bt,AmtNum)
           and(sql_date_convert(Date,DBDate),
               and(t_precedes(date([1990,1,1]),Date),
                   t_precedes(Date,date([1990,12,31])))))
       execute_in_future(display([TransId,DBDate,bt,AmtNum]))))
\end{verbatim}
which can be glossed as
\begin{quote}
``Find all {\tt TRANS} tuples with {\tt trn\_id} field {\tt TransId},
{\tt cheque\_date} field {\tt DBDate}, {\tt payee} field {\tt bt} and
{\tt amount} field {\tt AmtNum}, such that {\tt DBDate} represents a
date after 1/1/90 and before 31/12/90; and for each one display the
list {\tt [TransId, DBDate, bt, AmtNum]}.''
\end{quote}
Taking the conclusion first, the translation to SQL form produced is
\begin{verbatim}
A([TransId,DBDate,AmtNum],
  impl(sql_select([TransId,DBDate,AmtNum],
                  SELECT([t_1.trn_id,t_1.amount,t_1.cheque_date],
                         FROM([alias(TRANS,t_1)]),
                         WHERE([t_1.payee=bt,
                                sql_date_=<(1-JAN-90,
                                            t_1.cheque_date)),
                                sql_date_=<(t_1.cheque_date,
                                            31-DEC-90))])),
       execute_in_future(display([TransId,DBDate,bt,AmtNum])))
\end{verbatim}
Here, the abstract SQL syntax is a hopefully transparent
representation of the SQL query whose concrete syntax will be
\begin{verbatim}
"SELECT DISTINCT t_1.trn_id , t_1.cheque_date , t_1.amount
 FROM TRANS t_1
 WHERE t_1.payee = 'bt'
 AND '1-JAN-90' <= t_1.cheque_date
 AND t_1.cheque_date <= '31-DEC-90"
\end{verbatim}

We now explain in more detail how this result can be produced.
Processing traces the following series of top-level steps:
\begin{enumerate}
\item Translate representation-independent evaluable predicates such
as {\tt t\_\-pre\-cedes} into database access language dependent primitives
using AET.
\item Insert database column names into database predicates.
\item Make column names unique by associating a ``relation alias'' with each
      occurrence of a database predicate.
\item Group together conjunctions of goals suitable for turning into
      {\tt SELECT} statements. Goals suitable for inclusion are goals
      representing database lookups and goals that map onto conditions
      in {\tt WHEN} clauses. For each such conjunction, do the following:
      \begin{enumerate}
      \item Extract the variables that are to be selected.
      \item Replace variables by column descriptions.
      \item Extract the variables that are constrained to have fixed values,
            and add these as suitable equalities to the {\tt WHEN} clause.
      \item If more than one relation instance is referenced, add equalities
            to the {\tt WHEN} clause to represent the join.
      \end{enumerate}
\item Simplify the result.
\end{enumerate}
We use the example to illustrate. The first step is to use AET to translate
the occurrences of {\tt t\_precedes}. The relevant equivalence is
\begin{verbatim}
t_precedes(Date1,Date2) <->
exists([DBDate1,DBDate2],
  and(sql_date_convert(Date1,DBDate1),
      and(sql_date_convert(Date2,DBDate2),
          sql_date_=<(DBDate1,DBDate2))))
\end{verbatim}
applied twice; after some simplification (exploiting the fact that
{\tt db\_\-date\_\-convert} is a function from each of its arguments
to the other one, see section~\ref{Functional-relations}), the result is
\begin{verbatim}
A([TransId,Date,DBDate,AmtNum],
  impl(and(TRANS(TransId,DBDate,bt,AmtNum)
        and(sql_date_convert(Date,DBDate),
         and(sql_date_=<(1-JAN-90,DBDate)
             sql_date_=<(DBDate,31-DEC-90))))),
       execute_in_future(display([TransId,DBDate,bt,AmtNum])))
\end{verbatim}
Note that {\tt 1-JAN-90} and {\tt 31-DEC-90} are atomic SQL date
representations.
The next two steps are fairly trivial in nature, and involve substituting
SQL column names in the {\tt TRANS} relation and creating a unique relation
alias
{\tt t\_1} for it. (Since there is only one relation here, this step is not
actually necessary, but we show it for completeness). The result is
\begin{verbatim}
A([TransId,Date,DBDate,AmtNum],
  impl(and(tuple(t_1,TRANS([trn_id=TransId,
                            cheque_date=DBDate,
                            payee=bt,
                            amount=AmtNum])),
         and(sql_date_convert(Date,DBDate),
          and(sql_date_=<(1-JAN-90,DBDate)
              sql_date_=<(DBDate,31-DEC-90))))),
       execute_in_future(display([TransId,DBDate,bt,AmtNum])))
\end{verbatim}
Conjunctions are now when possible turned into {\tt sql\_select} goals;
the only suitable conjunction is
\begin{verbatim}
and(tuple(t_1,TRANS([trn_id=TransId,
                     cheque_date=DBDate,
                     payee=bt,
                     amount=AmtNum])),
     and(sql_date_=<(1-JAN-90,DBDate)
         sql_date_=<(DBDate,31-DEC-90)))
\end{verbatim}
Here, the database look-up goals are the singleton
\begin{verbatim}
{tuple(t_1,TRANS([trn_id=TransId,
                  cheque_date=DBDate,
                  payee=bt,
                  amount=AmtNum]))}
\end{verbatim}
and the initial {\tt WHEN} clause goals are
\begin{verbatim}
{sql_date_=<(1-JAN-90,DBDate)
 sql_date_=<(DBDate,31-DEC-90))}
\end{verbatim}
Consulting the database look-up goals, the variables to be selected
are {\tt TransId}, {\tt DBDate} and
{\tt AmtNum}, so they end up being the list of variables in the
{\tt sql\_select} goal's first argument; then replacing the variables
inside the {\tt SELECT} with column descriptions, we replace
{\tt TransId} with {\tt t\_1.trn\_id}, {\tt DBDate} with
{\tt t\_1.cheque\_date} and {\tt AmtNum} with {\tt t\_1.amount}.
The {\tt WHEN} clause goals now become
\begin{verbatim}
{sql_date_=<(1-JAN-90,t_1.cheque_date)
 sql_date_=<(t_1.cheque_date,31-DEC-90))}
\end{verbatim}
There is a single column, {\tt payee}, constrained to have a fixed value, {\tt
bt},
so it adds another goal
\begin{verbatim}
t_1.payee=bt
\end{verbatim}
to the {\tt WHEN} clause. The final result is the {\tt sql\_select} goal
\begin{verbatim}
sql_select([TransId,DBDate,AmtNum],
           SELECT([t_1.trn_id,t_1.amount,t_1.cheque_date],
                  FROM([alias(TRANS,t_1)]),
                  WHERE([t_1.payee=bt,
                         sql_date_=<(1-JAN-90,
                                     t_1.cheque_date)),
                         sql_date_=<(t_1.cheque_date,
                                     31-DEC-90))])))))
\end{verbatim}
Replacing the conjunction with the {\tt sql\_select} goal, we reach the final
form,
\begin{verbatim}
A([TransId,DBDate,AmtNum],
  impl(sql_select([TransId,DBDate,AmtNum],
                  SELECT([t_1.trn_id,t_1.amount,t_1.cheque_date],
                          FROM([alias(TRANS,t_1)]),
                          WHERE([t_1.payee=bt,
                                 sql_date_=<(1-JAN-90,
                                             t_1.cheque_date)),
                                 sql_date_=<(t_1.cheque_date,
                                             31-DEC-90))]))),
       execute_in_future(display([TransId,DBDate,bt,AmtNum])))
\end{verbatim}

\end{document}